\documentclass{article}
\usepackage[accepted]{template/icml2026}

\usepackage{template/packages}

\def\title{Vision Transformer Finetuning Benefits from Non-Smooth Components}

\icmltitlerunning{Vision Transformer Benefits from Non-Smooth Components}

\begin{document}
\addtocontents{toc}{\protect\setcounter{tocdepth}{0}}

\twocolumn[
  \icmltitle{\title}
  \icmlsetsymbol{equal}{*}

  \begin{icmlauthorlist}
    \icmlauthor{Ambroise Odonnat}{a,b,c} 
    \icmlauthor{Laetitia Chapel}{b,d}
    \icmlauthor{Romain Tavenard}{b,c}
    \icmlauthor{Ievgen Redko}{a}
  \end{icmlauthorlist}

  \icmlaffiliation{a}{Noah's Ark Lab}
  \icmlaffiliation{b}{IRISA}
  \icmlaffiliation{c}{Univ. Rennes 2, Inria}
  \icmlaffiliation{d}{L'Institut Agro Rennes-Angers}

  \icmlcorrespondingauthor{Ambroise Odonnat}{\href{mailto:ambroise.odonnat@gmail.com}{ambroise.odonnat@gmail.com}}

  \icmlkeywords{Machine Learning, ICML}

  \vskip 0.3in
  
]

\printAffiliationsAndNotice{}

\begin{abstract}
The smoothness of the transformer architecture has been extensively studied in the context of generalization, training stability, and adversarial robustness. However, its role in transfer learning remains poorly understood. In this paper, we analyze the ability of vision transformer components to adapt their outputs to changes in inputs, or, in other words, their \emph{plasticity}. Defined as an average rate of change, it captures the sensitivity to input perturbation; in particular, a high plasticity implies a low smoothness. Our theoretical analysis and extensive experiments -- over $1,000$ finetuning runs on large-scale vision transformers -- showcase that this perspective provides principled guidance in choosing the components to prioritize during adaptation. A key takeaway for practitioners is that the high plasticity of the attention modules and feedforward layers consistently leads to better finetuning performance. Our findings depart from the prevailing assumption that smoothness is desirable, offering a novel perspective on transformers' functional properties.
\begin{center}
\faGithub \quad \href{https://github.com/ambroiseodt/vit-plasticity}{\texttt{vit-plasticity}}
\end{center}
\vspace{-1.2em}
\end{abstract}

\begin{figure}[!t]
    \centering
    \includegraphics[width=\linewidth]{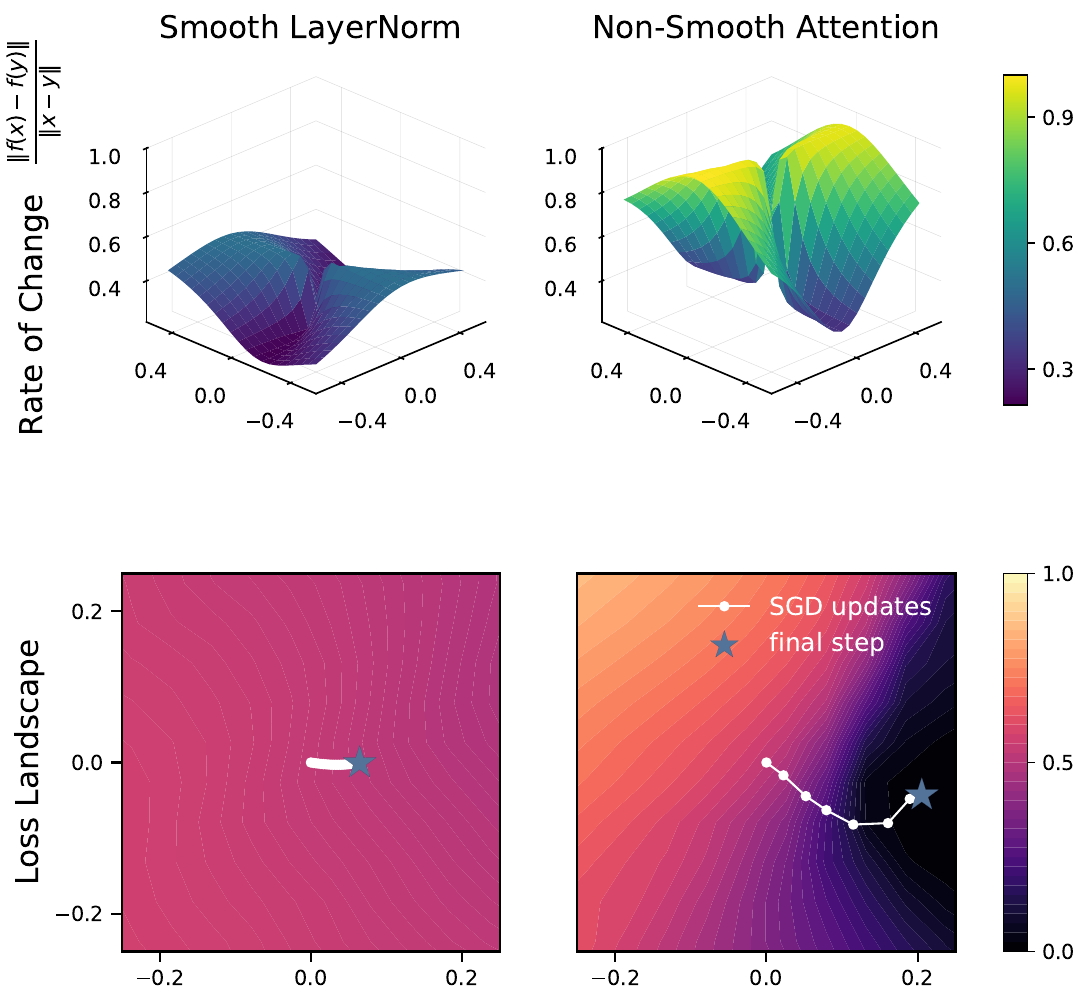}
    \caption{\textbf{Non-smooth components facilitate finetuning.} We illustrate the benefits of a high plasticity during the finetuning of ViT-Base on Cifar10 (values normalized to $[0,1]$). Smooth modules like LayerNorm (\textbf{top left}) have low and steady rates of change, resulting in low plasticity (see~\cref{def:measure}). This constrains the gradient norms during the optimization, leading to a slow descent on the loss landscape (\textbf{bottom left}). In contrast, the rates of change of non-smooth components, such as multi-head attention (\textbf{top right}), are large and vary a lot, resulting in high plasticity and gradients of high magnitude. This allows the exploration of the loss landscape and a faster descent towards (local) minima (\textbf{bottom right}).}
    \label{fig:intro}
    \vspace{-1.2em}
\end{figure}

\begin{figure*}[!h]
\centering
\begin{minipage}[c]{.25\textwidth}
    \centering
    \begin{adjustbox}{width=.95\textwidth}
    \begin{tikzpicture}[scale=1,baseline={([yshift=-.5ex]current bounding box.center)}]
    \begin{scope}[xshift=-3cm]
\definecolor{ln1_color}{HTML}{daa4ac} 
\definecolor{mha_color}{HTML}{37abb5}
\definecolor{ln2_color}{HTML}{b153a1}
\definecolor{ffn_color}{RGB}{194,232,247}
\definecolor{fc1_color}{HTML}{a291e1}
\definecolor{fc2_color}{HTML}{858ec2}
\definecolor{gray_bbox_color}{RGB}{247,247,247}
\def\background{gray_bbox_color}

\draw[fill=\background, line width=0.046875cm, rounded corners=0.300000cm] (-0.55000, -0.45) -- (2.8000, -0.45) -- (2.8000, 6.12) -- (-0.55000, 6.12) -- cycle;

\node[text width=2.500000cm, anchor=north, align=center] at (1.250000,-1.0000) {$(x_1, \ldots, x_n)$};
\draw[line width=0.046875cm, -latex] (1.250000, -0.950000) -- (1.250000, 0.300000);

\draw[line width=0.046875cm, fill=ln1_color!80, rounded corners=0.100000cm] (0.450000, 0.800000) -- (2.0000, 0.800000) -- (2.0000, 0.300000) -- (0.450000, 0.300000) -- cycle;
\node[text width=2.500000cm, align=center] at (1.250000,0.550000) {LN1};
\draw[line width=0.046875cm, -latex] (1.250000, 0.800000) -- (1.250000, 1.60000);

\draw[-latex, line width=0.046875cm, rounded corners=0.200000cm] (1.250000, -0.20000) -- (-0.25000, -0.2000) -- (-0.250000, 2.720000) -- (1.05000, 2.720000);
\draw[-latex, line width=0.046875cm, rounded corners=0.200000cm] (1.250000, 1.230000) -- (0.7000, 1.230000) -- (0.7000, 1.63000);
\draw[-latex, line width=0.046875cm, rounded corners=0.200000cm] (1.250000, 1.230000) -- (1.80, 1.230000) -- (1.80, 1.63000);

\draw[line width=0.046875cm, fill=mha_color!90, rounded corners=0.100000cm] (0.450000, 2.130000) -- (2.0000, 2.130000) -- (2.0000, 1.630000) -- (0.450000, 1.630000) -- cycle;
\node[text width=2.500000cm, align=center] at (1.250000,1.880000) {MHA};
\draw[line width=0.046875cm] (1.250000, 2.130000) -- (1.250000, 2.5250000);
\draw[line width=0.046875cm, -latex] (1.250000, 2.940000) -- (1.250000, 3.430000);

\node[circle, draw, minimum size=0.25em, inner sep=0pt, line width=0.046875cm, align=center, fill=white] at (1.250000,2.720000) {$\bm{\mathbf{+}}$};

\draw[line width=0.046875cm, fill=ln2_color!70, rounded corners=0.100000cm] (0.450000, 3.930000) -- (2.00000, 3.930000) -- (2.00000, 3.430000) -- (0.450000, 3.430000) -- cycle;
\node[text width=2.500000cm, align=center] at (1.250000,3.68000) {LN2};
\draw[line width=0.046875cm, -latex] (1.250000, 3.93000) -- (1.250000, 4.68000);

\draw[-latex, line width=0.046875cm, rounded corners=0.200000cm] (1.250000, 3.050000) -- (-0.250000, 3.050000) -- (-0.250000,5.770000) -- (1.05000, 5.770000);

\draw[line width=0.046875cm, fill=white, rounded corners=0.100000cm] (0.450000, 5.18000) -- (2.00000, 5.18000) -- (2.00000, 4.68000) -- (0.450000, 4.68000) -- cycle;
\node[text width=2.500000cm, align=center] at (1.250000,4.930000) {FFN};
\draw[line width=0.046875cm] (1.250000, 5.180000) -- (1.250000, 5.5750000);
\draw[line width=0.046875cm, -latex] (1.250000, 5.990000) -- (1.250000, 6.480000);

\node[circle, draw, minimum size=0.25em, inner sep=0pt, line width=0.046875cm, align=center, fill=white] at (1.250000,5.770000) {$\bm{\mathbf{+}}$};

\draw[thick][dashed](2.0, 4.68) -- (3.05, 3.87);
\draw[thick][dashed] (2.0, 5.18) -- (3.05, 6.1);

\pgfmathsetmacro{\yposOutput}{6.48000} 
\node[text width=2.500000cm, anchor=south, align=center] at (1.250000,\yposOutput) {$(z_1, \ldots, z_n)$};

\draw[line width=0.046875cm, fill=fc1_color!80, rounded corners=0.100000cm] (3.050000, 4.34) -- (4.6000, 4.34) -- (4.6000, 3.84) -- (3.050000, 3.84) -- cycle;
\node[text width=2.500000cm, align=center] at (3.850000,4.090000) {FC1};
\draw[line width=0.046875cm] (3.850000, 4.34) -- (3.850000, 4.685000);
\draw[line width=0.046875cm, -latex] (3.850000, 5.19000) -- (3.850000, 5.64000);

\node[circle, draw, minimum size=1.2em, inner sep=0pt, line width=0.046875cm, align=center] at (3.850000,4.930000) {$\bm{\sigma}$};

\draw[line width=0.046875cm, fill=fc2_color!70, rounded corners=0.100000cm] (3.050000, 6.1400) -- (4.60000, 6.14000) -- (4.60000, 5.64000) -- (3.050000, 5.64000) -- cycle;
\node[text width=2.500000cm, align=center] at (3.850000,5.890000) {FC2};
\end{scope}
    \end{tikzpicture}
    \end{adjustbox}
    \label{fig:transformer_encoder}
\end{minipage}%
\begin{minipage}[c]{.75\textwidth}
  \centering
  \includegraphics[width=\linewidth, valign=m]{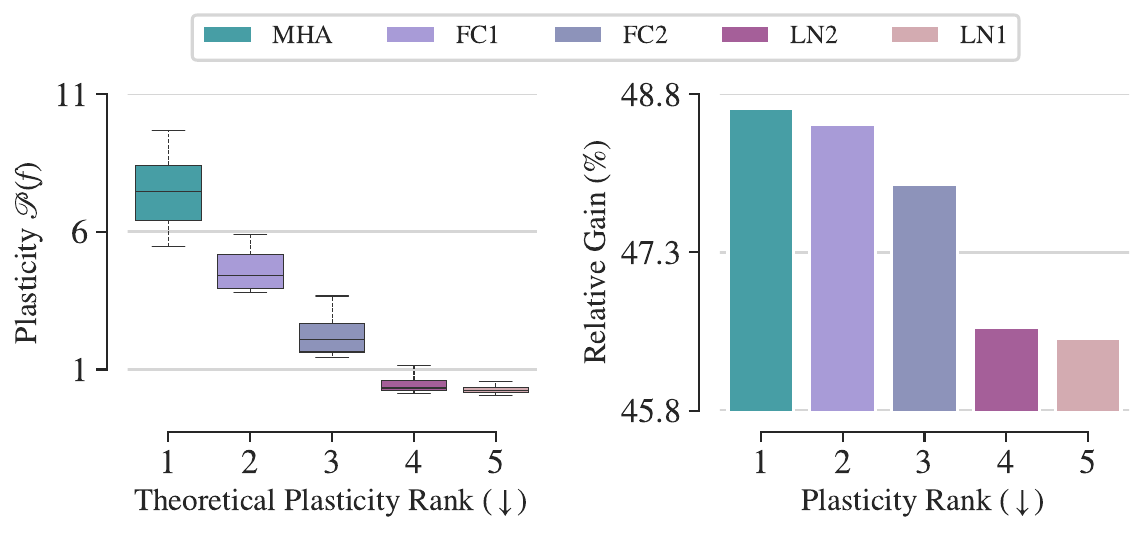}
\end{minipage}
\caption{\textbf{Overview of our contributions.} We conduct a comprehensive analysis of vision transformer components (\textbf{left}) through the perspective of their plasticity (\cref{def:measure}). Our theoretical analysis allows us to rank modules in terms of their plasticity (\cref{sec:theory}). Experiments on large-scale ViTs support our theoretical insights (\cref{sec:analysis}), as shown by the distribution of plasticity over all benchmarks (\textbf{middle}). Through large-scale finetuning runs on an $86$M-parameter ViT (\cref{sec:finetuning}), we demonstrate the real-world benefits of plasticity. As showcased by the average relative gain, i.e,  improvement over the linear probing accuracy, on a diverse set of $11$ classification benchmarks (\textbf{right}), a higher plasticity yields greater finetuning benefits.}
\label{fig:overview}
\vspace{-1.2em}
\end{figure*}
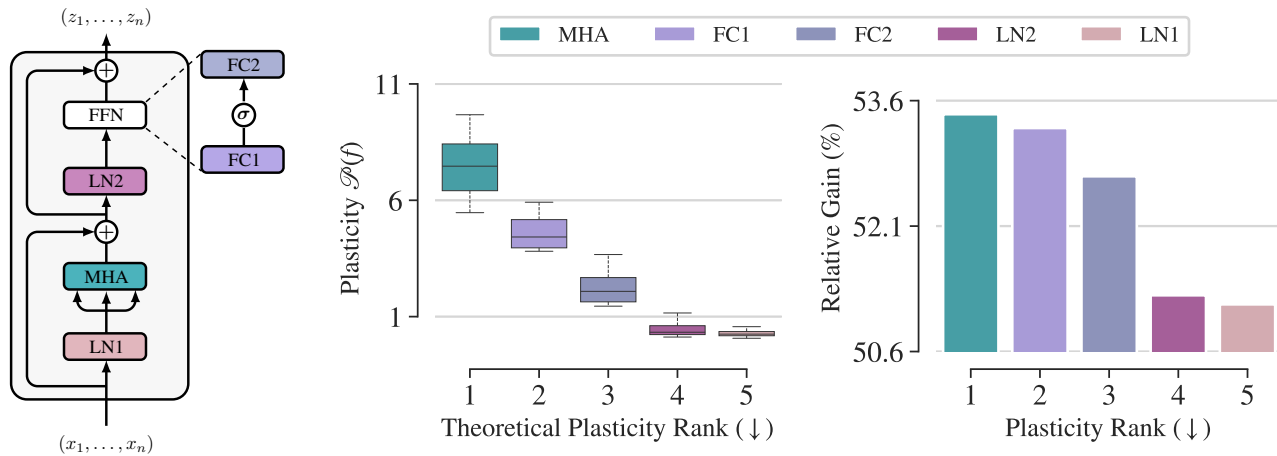

\section{Introduction}
\label{sec:intro}
Transformers~\citep{vaswani2017attention} have become the default backbone of state-of-the-art models in a wide range of domains, including natural language processing~\citep{touvron2023llama,brown2020gpt3}, computer vision~\citep{caron2021emerging, dosovitskiy2021vit}, time series forecasting~\citep{nie2023patchtst, ilbert2024samformer}, and mathematical reasoning~\citep{comanici2025gemini25pushingfrontier, deepseekai2025deepseekr1}. These foundation models are typically pretrained on large amounts of diverse data and then adapted to more specific domains~\citep{shukor2025scalinglawsoptimaldata}. In practice, the discrepancy between the training and downstream data can hurt performance~\citep{quionero2009shift} and requires updating the model weights to adapt to the distribution shift. 

\paragraph{Finetuning foundation models.} The cost of adaptation has drastically increased, with models growing larger and larger as a byproduct of the scaling hypothesis~\citep{hoffman2022computeoptimal, kaplan2020scalinglawsneurallanguage}. This has led to considerable research effort toward parameter-efficient finetuning methods~\citep[PEFT, ][]{han2024parameterefficient,liu2022fewshot,houlsby2019peft}. It allows finetuning foundation models at a fraction of the cost required for full adaptation and quickly became a standard practice in research and industry~\citep{peft}. We focus on the popular family of selective approaches, where only a subset of parameters is updated during finetuning~\citep{lee2023surgical,lee2019elsadofreezinglayers,guo2019spottune}. Recent works \emph{empirically} studied the benefits of adapting one type of transformer component across the whole network: the normalization layers~\citep{zhao2024tuning}, the attention module~\citep{touvron2022three}, or the feedforward layers~\citep{ye2023partial}. However, little is known from a theoretical perspective about the adaptability of those modules\footnote{In what follows, we use the terms module and component interchangeably, always referring to normalization layers, multi-head attention modules, and feedforward layers.}. This motivates us to ask: 
\begin{quote}
\centering
\textit{Which transformer components should be prioritized during finetuning and why?}
\end{quote}
\paragraph{Our approach.} We focus on vision transformers~\citep[ViT,][]{dosovitskiy2021vit} and aim to reconcile the intrinsic functional properties of the individual components with the empirical performance observed when adapting them. To avoid confounders and since considering all possible combinations of the modules is computationally prohibitive, we conduct a systematic component-wise study where each type of module is finetuned in isolation. We build upon the intuition that promoting the smoothness of a neural network, e.g., by regularizing its Lipschitz constant~\citep{newhouse2025lipschitz}, reduces its sensitivity to input perturbations~\citep{rosca2020smooth}. While this is desirable for generalization~\citep{krogh1991wd,rosca2020smooth,neyshabur2017generalization}, training stability~\citep{zhai2023sigmareparam}, or adversarial robustness~\citep{miyato2018spectral},
it limits the degree of freedom a given component has to adapt its outputs to changes in the inputs, and thus its \emph{plasticity}. As a result, it hinders the adaptation to downstream data during finetuning. This motivates us to quantify the plasticity of transformer components as an average rate of change, where high values would indicate low smoothness (the formal definition shall come in ~\cref{sec:approach}).

\paragraph{Our contributions.}  We provide a theoretical ranking of vision transformer components in terms of plasticity, supported by empirical evidence. We demonstrate through comprehensive experiments that high plasticity consistently leads to better finetuning performance. Our main contributions, illustrated in~\cref{fig:overview}, are:
\begin{enumerate}[leftmargin=*]
    \item \textbf{Intuitive measure:} We formalize the plasticity of a module as its average rate of change, which captures how it responds to variations in its input (\cref{sec:approach}). 
    \item \textbf{Theoretical analysis:} We establish a theoretical ranking among transformer components by deriving upper bounds on their plasticity (\cref{sec:theory}). 
    \item \textbf{Plasticity ranking:} We validate our theoretical insights on pretrained ViTs, showing that the attention module consistently has the highest plasticity, followed by the first and second feedforward layers, the LayerNorm preceding the feedforward, and finally the LayerNorm preceding the attention module. This ranking is not limited to ViTs, and holds for DINOv3 and GPT2 (\cref{sec:analysis}). 
    \item \textbf{Finetuning benefits:} We conduct exhaustive finetuning runs of large-scale ViTs on a diverse set of classification benchmarks with both SGD and Adam optimizers. Our findings showcase that adapting modules with high plasticity, namely attention modules and the feedforward layers, results in higher and more stable performance across initialization and learning rates (\cref{sec:finetuning}).
\end{enumerate}
Our work, supported both theoretically and empirically, provides a novel perspective on the role of smoothness in finetuning. We believe that the highlighted link between plasticity and gradient norms, illustrated in~\cref{fig:intro}, will help guide the design of more efficient adaptation methods. 

\section{Background}
\label{sec:background}
Throughout the paper, we use the notation $[n]$ to represent the set $\{1, \ldots, n\}$. The Euclidean norm of $\RR^n$ is denoted by $\|\cdot \|$ and its $\ell_\infty$-norm is denoted by $\|\cdot\|_\infty$. A sequence of tokens $x = (x_1, \ldots, x_n) \in (\RR^d)^n$ can be seen as a matrix\footnote{In the PyTorch implementation, all matrices are transposed because the input data is viewed as matrices in $\RR^{n \times d}$ instead of the common $\RR^{d \times n}$ we use.} in $\RR^{d \times n}$ with Frobenius norm $\| x \|_\mathrm{F} = (\sum_{i}^n \|x_{i}\|^2)^{1/2}$ and spectral norm $\|x\|_2 = \sigma_\mathrm{max}(x)$, with $\sigma_\mathrm{max}(x)$ the largest singular value of $x$. We denote by $B_r \subset \RR^d$ the closed ball centered at $0$ with radius $r>0$. 

\textbf{Neural network smoothness.} Formally, the smoothness of a function is related to the number of continuous derivatives it has on its domain. In deep learning, it can refer to several related concepts, such as differentiability, Lipschitz continuity, or robustness to input perturbations~\citep{rosca2020smooth}. A common way to quantify smoothness is through the notion of Lipschitz continuity. A function $f \colon (\RR^d)^n \to (\RR^d)^n$ is said to be Lipschitz continuous if there exists a constant $K \leq 0$ such that for any pair of inputs $x, y \in (\RR^d)^n$, we have 
$\|f(x) - f(y) \|_\mathrm{F} \leq K \|x-y\|_\mathrm{F}$. The smallest constant $K$ is called the Lipschitz constant of $f$, denoted by $\mathrm{Lip}(f)$, and writes $\mathrm{Lip}(f) = \sup_{x \neq y \in (\RR^d)^n} \frac{\|f(x) - f(y)\|_\mathrm{F}}{\|x-y\|_\mathrm{F}}$.

\textbf{Vision transformers.} A ViT takes as input 2D images, embedded into sequences of tokens by splitting them into patches of size $P$, which are then flattened and linearly projected in $\RR^d$. The architecture consists of a succession of transformer encoders. Akin to BERT~\citep{devlin2019bert}, a classification token \texttt{CLS} is prepended to the sequence of tokens to perform classification. A transformer encoder is illustrated in~\cref{fig:overview} (left), where the LayerNorms are denoted by LN1 and LN2, the attention module is denoted by MHA, and the feedforward linear layers are denoted by FC1 and FC2. After the last layer, the embedding of the \texttt{CLS} token is pooled to perform classification. Implementation details are given in~\cref{app:vit}.

\textbf{Transformer components.} We recall below how each module operates on a sequence of tokens $x \in (\RR^d)^n$.
\begin{itemize}[leftmargin=*]
\item \textbf{LayerNorms}: A LayerNorm with weights $\gamma, \beta \in \RR^d$ acts on each input token individually with the formula
\begin{equation*}
    f(x) = \mleft(\gamma \odot \frac{x_j - \mu(x_j)}{\sigma(x_j)} + \beta\mright)_{1 \leq j \leq n} \in (\RR^d)^n,
\end{equation*}
where $\odot$ is the element-wise product and $\mu(x_j), \sigma(x_j)$ are the mean and standard deviation of the token $x_j$.
\item \textbf{Multi-head self-attention}: Let $H \in \NN$ such that $k = \frac{H}{d}$ is an integer. Let $Q^h, K^h, V^h$ be matrices in $\RR^{k \times d}$ and $O^h \in \RR^{d \times k}$. A multi-head self-attention module with weights $(O^h, Q^h, K^h, V^h)_{1 \leq h \leq H}$ outputs 
\[f(x) = \sum_{h=1}^H O^h f^h_\mathrm{att}(x) \in (\RR^d)^n,
\]
where the single-head self-attention $f_\mathrm{att}^h$ has weights $Q^h, K^h, V^h$ and writes 
\[
    f^h_\mathrm{att}(x) = (V^hx) \cdot \mathrm{softmax}\mleft(\frac{\mleft(Q^hx\mright)^\top K^hx}{\sqrt{k}}\mright) \in (\RR^k)^n,
\]
with the softmax applied row-wise.
\item \textbf{Feedforward linear layers}: A feedforward module with weights $W_1 \in \RR^{d \times 4d}, W_2 \in \RR^{4d \times d}$ combines two linear layers $x \mapsto W_1x$ and $x \mapsto W_2x$ with a GeLU to output
\[
f(x) = W_2\mathrm{GeLU}(W_1x)\in (\RR^d)^n.
\]
\end{itemize}

\section{Vision transformer plasticity}
\label{sec:approach}
Regularizing the Lipschitz constant of a model is a common approach to encourage smoothness~\citep{miyato2016distributionalsmoothingvirtualadversarial,newhouse2025lipschitz}. While it serves as a useful inductive bias in generalization~\citep{sokolic2017robust,bartlett2017spectral}, training stability~\citep{zhai2023sigmareparam,miyato2018spectral}, and adversarial robustness~\citep{jia2024lipschitz,tsuzuku2018lipschitz}, too much smoothness can constrain the model's capacity and its adaptability to new tasks, as shown in~\citet{rosca2020smooth}. This motivates us to identify the components whose Lipschitz constants might be too small~\citep[see][Section 5]{rosca2020smooth}, which could impact the adaptation during finetuning. This can be done by analyzing the rates of change of the components $f$ since they lower bound the Lipschitz constant via $\frac{\|f(x) - f(y)\|_\mathrm{F}}{\|x-y\|_\mathrm{F}} \leq \mathrm{Lip}(f)$.

\paragraph{Plasticity measure.} Building upon this intuition, we formalize below the \emph{plasticity}\footnote{Akin to the neuroplasticity of the brain defined as its ability to change its activity ``in response to intrinsic or extrinsic stimuli"~\citep{puderbaugh2023neuroplasticity}. The loss of plasticity at the network level has been studied in deep reinforcement learning~\citep{lyle2023plasticity} or continual learning~\citep{dohare2024plasticity}.} of a module, i.e., its ability to adapt its output in response to changes in the inputs:
\begin{boxdef}[Plasticity]
\label{def:measure}
    Let $\nu$ be the uniform distribution over the set of distinct pairs of sequences of tokens in $(\RR^d)^n$. We define the plasticity of a transformer component $f$ as 
    \begin{equation}
        \label{eq:measure}
        \mathscr{P}(f) = \EE_{(x, y) \sim \nu} \mleft[\frac{\| f(x) - f(y)\|_\mathrm{F}}{\|x-y\|_\mathrm{F}}\mright].
    \end{equation}
\end{boxdef}
\cref{def:measure} ensures that, for any component $f$, we have $\mathscr{P}(f) \leq \mathrm{Lip}(f)$. The Lipschitz constant is a worst-case estimation that ensures control over each rate of change. To better capture the overall behavior of transformer components over the distribution of input sequences, we compute the average rate of change. This is reminiscent of the notion of average smoothness, defined for functions on a metric probabilistic space in~\citet{ashlagi2021smoothness}. Two regimes of plasticity can be distinguished. If $\mathscr{P}(f) < 1$, the module $f$ contracts the input discrepancy on average. If $\mathscr{P}(f) > 1$, then $f$ amplifies the change in the input on average and pushes the value of $\mathrm{Lip}(f)$ from below. In the rest of this work, we will say that components in the first regime have low plasticity and are smooth, and that the components in the second regime have high plasticity and low smoothness (or are non-smooth, by abuse of language).

\paragraph{Connection to finetuning.} The plasticity measure introduced in~\cref{def:measure} captures the sensitivity of transformer components to input changes. A high plasticity implies a high Lipschitz constant and thus low smoothness. For a given module $f$ with weights $\theta$, we have from~\citet{federer1969geometric} that $\|\nabla_x f\|_\mathrm{F} \leq \mathrm{Lip}(f)$. Let $\mathcal{L}$ be the finetuning loss. During gradient descent, the weights are updated following $\theta \leftarrow \theta - \eta \nabla_\theta \mathcal{L}$, which involves the gradient of $f$ with respect to the parameters via
\[
\nabla_\theta \mathcal{L} = (\nabla_f\mathcal{L})\frac{\partial f}{\partial \theta},
\]
using the Vector-Jacobian product notation~\citep{bethune2024dpsgd,dagreou2024how}. Since our goal is to identify the components that adapt best to downstream data during finetuning, a natural question is: \emph{what is the connection between the gradient with respect to inputs and the gradient with respect to the parameters?} On the theoretical side, \citet{bethune2024dpsgd} showed that these notions are two sides of the same coin. More precisely, the authors proved that regularizing the Lipschitz constant with respect to the inputs amounts to bounding the norm of the gradients with respect to the parameters~\citep[see][Theorem 1]{bethune2024dpsgd}. As such, too much regularization on the smoothness can impact optimization; conversely, having looser Lipschitz constraints, e.g., thanks to high plasticity, might facilitate the learning process. This has been empirically observed in \citet[Section 4.4]{newhouse2025lipschitz}, where the authors show that reducing the Lipschitz constant negatively impacts the performance of a $145$M Lipschitz-constrained transformer on FineWeb \citep{penedo2024the}. In particular, matching the NanoGPT baseline~\citep{modded_nanogpt_2024} requires a Lipschitz constant of up to $10^{264}$.

\paragraph{Expected benefits of plasticity.} The connection between input-output and weight-output smoothness hints at the role of plasticity in the learning process. We expect the components with high plasticity, i.e., the non-smooth ones, to allow large gradient norms during finetuning, thus leading to a faster and better adaptation (note that we do not expect a linear relationship with downstream performance, either, as is the case with unsupervised accuracy estimation methods~\citep{xie2025leveraging,xie2024mano,deng2023nuclear,garrido2023rankme}). This process is illustrated in~\cref{fig:intro}. This can be understood intuitively, with plastic components carrying more information about the downstream data than the smooth ones. Provided our insights are confirmed through experiments (\cref{fig:overview} offers a sneak peek for impatient readers), our perspective would depart from the conventional wisdom that promoting smoothness is beneficial to learning~\citep{rosca2020smooth,neyshabur2017generalization,miyato2018spectral,zhai2023sigmareparam}.

\section{Theoretical analysis} 
\label{sec:theory}
In this section, we derive upper bounds on the plasticity $\mathscr{P}(f)$. It allows us to compare transformer components in terms of plasticity. The proofs are given in~\cref{app:proofs}. We start with the LayerNorms whose upper bound is stated in the proposition below.
\begin{boxprop}[LayerNorm]
\label{prop:ln_bounds}
Let $f$ be a LayerNorm with weights $\gamma, \beta \in \RR^d$. Assume that all tokens in position $i \in [n]$ have the same mean $\mu_i$ and standard deviation $\sigma_i$ on $\RR^d$ and let $\sigma > 0$ be the minimal standard deviation. Then, we have $\mathscr{P}(f) \leq \frac{1}{\sigma} \|\gamma\|_\infty$.
\end{boxprop}
The requirement on tokens comes from the fact that images are normalized with ImageNet1k statistics during preprocessing~\citep{kolesnikov2020bit} and embedded into sequences of tokens with the same layer. This implies that the $\mu_i, \sigma_i$ depend only on the embedding layer. Having $\sigma_i = 0$ for some $i \in [n]$ would force all the tokens in position $i$ to be equal, independently of the embedded images. Since this is not the case, neither at initialization nor after pretraining, we must have $\sigma = \min_{i\in [n]} \sigma_i > 0$. We now proceed to bound the plasticity of the feedforward linear layers. This is reminiscent of the well-known upper bound on the Lipschitz constant of linear operators~\citep{virmaux2018lipschitz,federer1969geometric}.
\begin{boxprop}[Feedforward layer]
\label{prop:fc_bounds}
Let $f$ be a feedforward linear layer with weights $W \in \RR^{d \times 4d}$ (resp. $W \in \RR^{4d \times d}$). Then, we have $\mathscr{P}(f) \leq \|W\|_2$.
\end{boxprop}
We now proceed to the upper bound for the multi-head self-attention module. Since self-attention is not globally Lipschitz continuous~\citep{kim2021lipschitz}, we need to restrict ourselves to sequences $(x_1, \ldots, x_n)$ in $B_r^n$, where $B_r \subset \RR^d$ is the closed ball centered in $0$ with a radius of $r$. 
\begin{boxprop}[Multi-head self-attention]
\label{prop:mha_bounds}
Let $f$ be a multi-head self-attention module with weights $(O^h, Q^h, K^h, V^h)_{1 \leq h \leq H}$. Let $A^h = (Q^h)^\top K^h / \sqrt{k}$ and $r>0$. Assume that sequences of tokens are in $B_r^n$. Then, we have
\begin{equation*}
    \mathscr{P}(f) \leq \sum_{h=1}^H \|O^h\|_2 \|V^h\|_2\sqrt{3n + (12n+3)r^4\|A^h\|_2^2}.
\end{equation*}
\end{boxprop}
The setting of bounded tokens has been studied in~\citet{castin2024smooth} and holds in practice~\citep[see][Fig. 4]{darcet2024vision}. This can be understood by the fact that images are normalized during preprocessing before being projected in $\RR^d$ using a layer with bounded weights. As shown in~\citet[Proposition 3.4]{castin2024smooth}, the bound in~\cref{prop:mha_bounds} is tight in terms of sequence length $n$. In a ViT, the average token norm is $20$ (see~\cref{sec:analysis}) and the sequence length is around $200$. Hence, $r$ and $\sqrt{n}$ have a similar order of magnitude. It leads to an effective growth rate of $r^2\sqrt{n}$ in~\cref{prop:mha_bounds}, since $\|A^h\|_2 \geq 1$ in practice~\citep[see][Fig. 3]{zhai2023sigmareparam}. Recalling that the total energy of a digital image is defined as the sum of its squared pixel intensities, the next corollary allows us to obtain a tighter bound with a growth rate in $\sqrt{n}$.
\begin{boxprop}[Tighter upper bound]
\label{prop:mha_tighter_bounds}
Let $f$ be a multi-head self-attention module with weights $(O^h, Q^h, K^h, V^h)_{1 \leq h \leq H}$. Let $A^h = (Q^h)^\top K^h / \sqrt{k}$ and let $\alpha$ be the spectral norm of the embedding layer. Assume that sequences of tokens are obtained from images with a total energy bounded by $\mathcal{E} > 0$. Then, we have
\begin{equation*}
    \mathscr{P}(f) \leq \sum_{h=1}^H \|O^h\|_2 \|V^h\|_2\mleft(\sqrt{n} + \alpha^2 \mathcal{E}  \|A^h\|_2\mright).
\end{equation*}
\end{boxprop}
The assumption on input images, discussed in detail in~\cref{app:mha_tighter_bounds}, holds in a standard signal processing setting~\citep[see, e.g.,][]{goodman2005introduction,mallat2008wavelet}; It allows us to bound the Frobenius norm of sequences of tokens. This is key to obtaining the growth rate $\sqrt{n}$ in~\cref{prop:mha_tighter_bounds}, further improving~\cref{prop:mha_bounds}. Note that the mean-field limit with $n \to + \infty$~\citep{sander2022sinkformer,geshkovski2023the,castin2024smooth} is interesting from a mathematical perspective. In particular, it leads to upper bounds independent of the sequence length~\citep{castin2024smooth, geshkovski2023the}. However, this setting is not suitable for vision transformers where $n$ is usually below $10^3$~\citep{dosovitskiy2021vit,kolesnikov2020bit,dehghani2023vit22b}.

\paragraph{Theoretical ranking.} To compare the modules, we focus on the relative order of their upper bounds. ~\cref{prop:ln_bounds,prop:fc_bounds} imply that the bound over $\mathscr{P}(f)$ is tighter for the normalization than for the linear layers. Indeed, for a vector $\gamma \in \RR^d$ and a matrix $W \in \RR^{d \times m}$ with entries in a similar range, $\|\gamma\|_\infty$ is comparable to $\|W\|_\infty$, which is smaller than the spectral norm of $W$ since
\[
\forall i, j, |W_{ij}| \leq \sqrt{\sum_{i=1}^k |W_{ij}|^2} = \|We_j\| \leq \|W\|_2,
\]
with $e_j \in \RR^m$ has zero entry everywhere except in $j$-th position, where we used the fact that the spectral norm is the operator norm induced by the Euclidean norm. A similar analysis can be done for the multi-head self-attention module: since spectral norms are above $1$ in practice~\citep[see][Fig. 3]{zhai2023sigmareparam}, the sum over the heads of products of spectral norms and the dependency in the sequence length $n$ of~\cref{prop:mha_bounds,prop:mha_tighter_bounds} imply a looser control over the plasticity of the multi-head self-attention module compared to the LayerNorms and the feedforward. We validate our insight by numerically computing the upper bounds on an $86$M pretrained ViT with sequence length $n=197$ and $12$ attention heads; see~\cref{app:plasticity_theory} for details. In~\cref{fig:theory}, we can see the ranking between modules: the multi-head self-attention module has the highest upper bound, followed by the feedforward linear layers, and then the LayerNorms. The conclusion of the theoretical analysis is the following:
\begin{tcolorbox}[colback=\boxcol,
    colframe=black,
    arc=4pt,
    boxsep=0pt,
    boxrule=\boxwidth pt,
]%
\textbf{Takeaway 1.} Our analysis suggests the following plasticity ranking:
$\text{MHA} \rightarrow \text{FC1} \approx \text{FC2} \rightarrow \text{LN2} \approx \text{LN1}$.
\end{tcolorbox}
\paragraph{Extension to large language models.}
The results presented in~\cref{prop:fc_bounds,prop:ln_bounds,prop:mha_bounds} also hold for decoder-only models such as large language models~\citep[LLMs]{grattafiori2024llama3herdmodels,gemmateam2025gemma3technicalreport,radford2019gpt2}. Indeed, decoder blocks~\citep{vaswani2017attention} have the same global structure as encoder blocks and differ only at the attention module level, which becomes causal. Fortunately, ~\cref{prop:mha_bounds} is still verified for masked self-attention thanks to Theorem 4.3 in~\citet{castin2024smooth}. We will show in~\cref{sec:analysis} that ViTs and LLMs have similar plasticity patterns.

\begin{figure*}[!t]
    \centering
    \includegraphics[width=\linewidth]{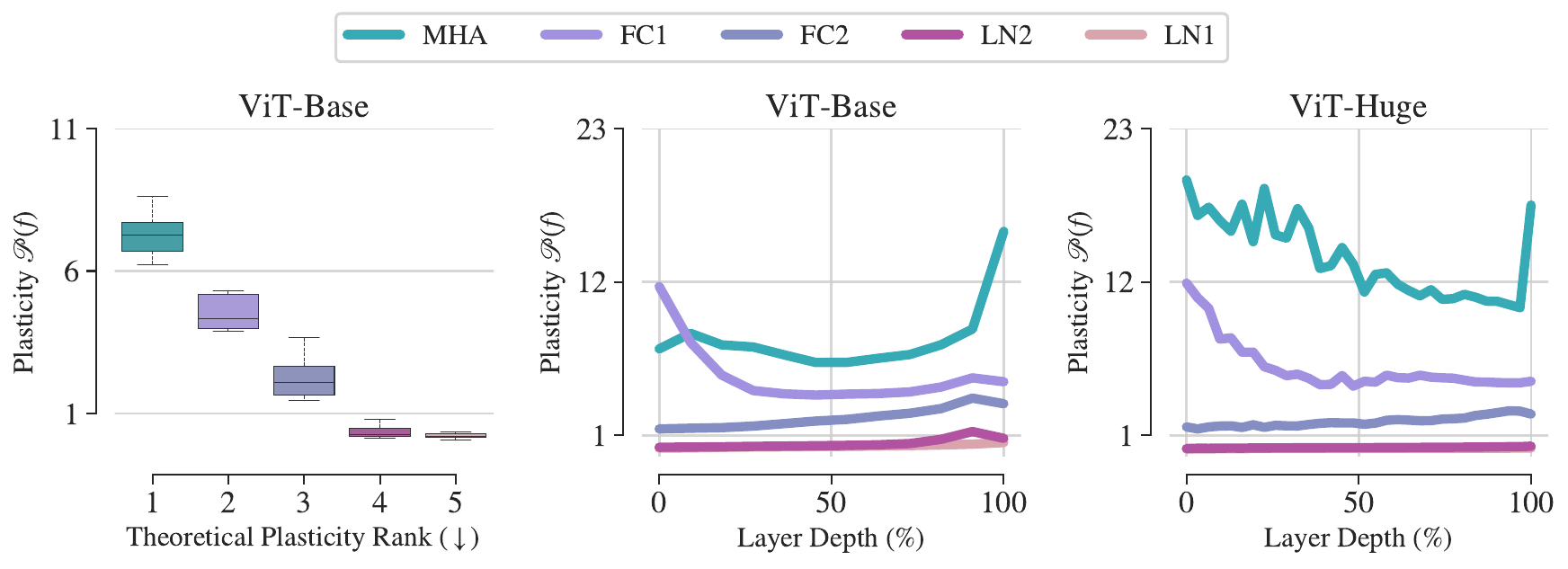}
    \caption{\textbf{Plasticity analysis on Sketch.} The distribution of rates of change $\|f(x)-f(y)\|_\mathrm{F}/\|x-y\|_\mathrm{F}$ on ViT-Base (\textbf{left}) follows the theoretical ranking of~\cref{sec:theory}. We observe along transformer blocks of ViT-Base (\textbf{middle}) that the attention module has the highest plasticity $\mathscr{P}(f)$, followed by the first and second linear layers of the feedforward. The LayerNorms are the most rigid, with a plasticity below $1$. The same pattern is obtained on ViT-Huge (\textbf{right}), where the higher attention plasticity further validates our theory (see \cref{prop:mha_bounds}) since the sequence length $n$ is larger than with ViT-Base.}
    \label{fig:analysis_sketch}
    \vspace{-1.2em}
\end{figure*}
\section{Experiments}
\label{sec:exps}
In this section, we experimentally show that (a) the plasticity of vision transformer components follows the ranking predicted by our theory (\cref{sec:analysis}) and (b) components with high plasticity lead to better and more stable finetuning accuracy across initializations and learning rates (\cref{sec:finetuning}). Our code is available at \href{https://github.com/ambroiseodt/vit-plasticity}{\texttt{github.com/ambroiseodt/vit-plasticity}}, and reproducibility details are given in~\cref{app:reproducibility}.

\paragraph{Experimental setup.} Unless otherwise specified, we conduct our experiments on large-scale ViTs of varying sizes ($86$M, $307$M, and $632$M parameters), pretrained on ImageNet22k~\citep{deng2009imagenet} (see~\cref{app:vit} for details). We perform both the plasticity and finetuning studies on a diverse set of $11$ commonly used classification benchmarks: Cifar10, Cifar100~\citep{krizhevsky2009learning}; $5$ variants from Cifar10-C~\citep{hendrycks2019robustness}: Contrast, Gaussian Noise, Motion Blur, Snow, Speckle Noise; $2$ domains from DomainNet~\citep{peng2019moment}: Clipart, Sketch; Flowers102~\citep{nilsback2008flowers102} and Pets~\citep{parkhi2012pets}. Images are resized to $224\times 224$ resolution and preprocessed following the protocol of~\citet{dosovitskiy2021vit}, see~\cref{app:data} for details. 
\subsection{Plasticity analysis}
\label{sec:analysis}
In this section, we compute the plasticity measure introduced in~\cref{def:measure} on pretrained vision transformers. To allow for diverse discrepancies $\|x-y\|_\mathrm{F}$, the sequences $x$ are obtained by embedding $12800$ pretraining images from ImageNet1k~\citep{deng2009imagenet}, and the sequences $y$ are obtained similarly on various downstream data. The full experimental details are provided in~\cref{app:plasticity_implem}. Results on Sketch are displayed in~\cref{fig:analysis_sketch}, and additional results on all benchmarks are given in~\cref {app:plasticity_exp}. 

\paragraph{Empirical ranking.} In~\cref{fig:analysis_sketch} (left), we display, for each module $f$ of ViT-Base, the distribution of rates of change $\|f(x)-f(y)\|_\mathrm{F}/\|x-y\|_\mathrm{F}$ on Sketch. We can see that the ranking established in~\cref{sec:theory} correctly predicts the practical behavior of transformer components. In addition, we observe that the first feedforward layer has larger rates of change than the second. Despite being closer, the LayerNorms also exhibit distinct plasticity, with the LayerNorm preceding the feedforward layer being less rigid than the one preceding the attention module. Our findings are consistent across all benchmarks, as can be seen in the overall distribution of plasticity displayed in~\cref{fig:overview} (middle) and in~\cref{fig:analysis_cifar10,fig:analysis_cifar100,fig:analysis_contrast,fig:analysis_gaussian_noise,fig:analysis_motion_blur,fig:analysis_snow,fig:analysis_speckle_noise,fig:analysis_clipart,fig:analysis_flowers102,fig:analysis_pet}. It allows us to refine the ordering established in~\cref{sec:theory} into:  
\[
\text{MHA} \rightarrow \text{FC1} \rightarrow \text{FC2} \rightarrow \text{LN2} \rightarrow \text{LN1}.
\]
In the rest of this work, this ordering will define the \emph{plasticity rank} of each component. In~\cref{fig:analysis_sketch} (middle), the evolution of the plasticity $\mathscr{P}(f)$ over the layers of ViT-Base is displayed. The $x$-axis represents the layer depth, denoted as a percentage of the overall depth. The ordering previously mentioned is respected. We can also see the two regimes of plasticity mentioned in~\cref{sec:approach}: the attention module and the feedforward linear layers have a high plasticity with values $\mathscr{P}(f) > 1$. In contrast, the LayerNorms have a low plasticity $\mathscr{P}(f) < 1$. Following our terminology, this implies that the attention modules and feedforward layers are non-smooth, contrary to the LayerNorms. 
\begin{rmk}[Smooth normalization layers]
The smoothness and low plasticity of normalization layers can be explained by the fact that, by design, they limit the propagation of perturbations in the input by rescaling the features. This has notably been leveraged in prior works to mitigate the non-stationarity in time series~\citep{kim2022revin}. As we will see in~\cref{sec:finetuning}, this is not a desirable property to adapt to downstream data.
\end{rmk} 
\paragraph{Impact of the sequence length.} We further confirm the theoretical insights of~\cref{sec:theory} by conducting a similar plasticity analysis on ViT-Huge, which has a longer sequence length $n=257$. The results are displayed in~\cref{fig:analysis_sketch} (right). We observe a similar evolution along the depth, with a larger plasticity for the attention module than with ViT-Base. This can be explained by the dependency on the sequence length $n$ in the attention upper bound of~\cref{prop:mha_bounds}. Our findings are consistent across all benchmarks, as displayed in~\cref{fig:analysis_cifar10,fig:analysis_cifar100,fig:analysis_clipart,fig:analysis_contrast,fig:analysis_flowers102,fig:analysis_gaussian_noise,fig:analysis_motion_blur,fig:analysis_pet,fig:analysis_snow,fig:analysis_speckle_noise}. This showcases, as hinted by the upper bounds of~\cref{sec:theory}, that plasticity is an intrinsic property of the components and their weights. The conclusion of the plasticity analysis is:
\begin{tcolorbox}[colback=\boxcol,
    colframe=black,
    arc=4pt,
    boxsep=0pt,
    boxrule=\boxwidth pt,
]%
\textbf{Takeaway 2.} The empirical \emph{plasticity ranking} of vision transformer modules supports our theoretical insights with:
$\text{MHA} \rightarrow \text{FC1} \rightarrow \text{FC2} \rightarrow \text{LN2} \rightarrow \text{LN1}.$
\end{tcolorbox}
\begin{figure}[!b]
    \centering
    \includegraphics[width=\linewidth]{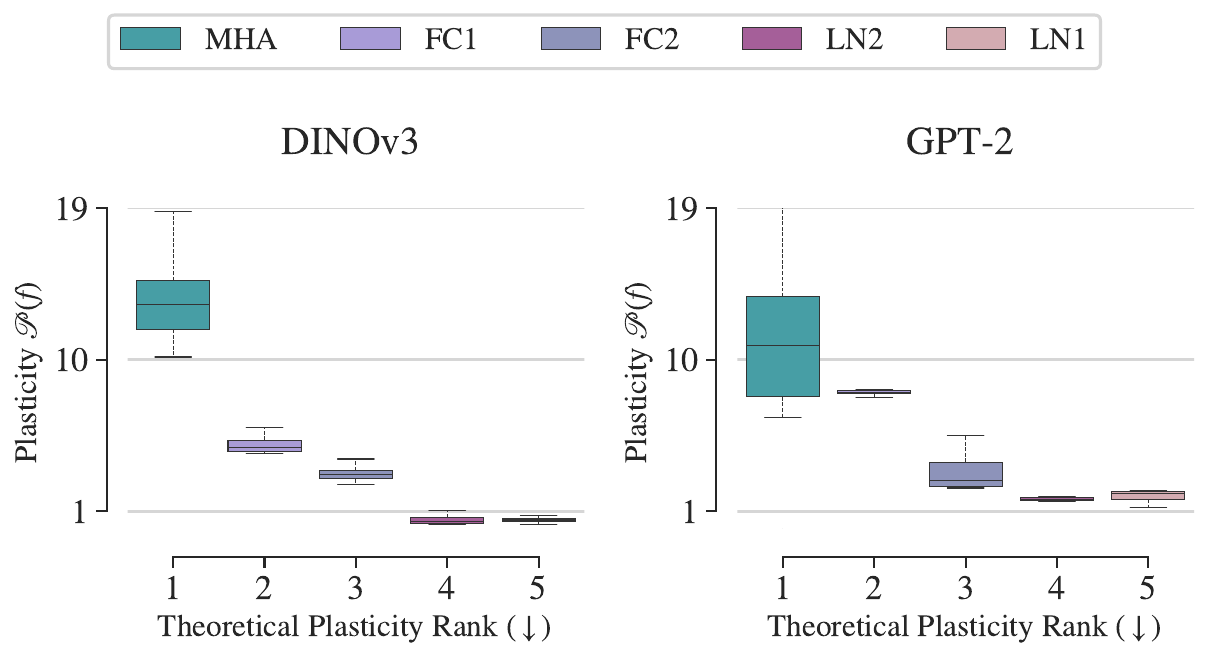}
    \caption{\textbf{Plasticity analysis on DINOv3 and GPT2.} The distribution of rates of change $\|f(x)-f(y)\|_\mathrm{F}/\|x-y\|_\mathrm{F}$ on DINOv3 (\textbf{left}) and GPT2 (\textbf{right}) follows the theoretical ranking of~\cref{sec:theory}: the attention module has the highest plasticity $\mathscr{P}(f)$, followed by the first and second linear layers of the feedforward. The LayerNorms are the most rigid, with a plasticity below or near $1$. This behavior is consistent with pretrained ViTs shown in~\cref{fig:analysis_sketch}.}
    \label{fig:plasticity_ablation}
    \vspace{-1.2em}
\end{figure}
\paragraph{Beyond supervised vision transformers.} We extend our analysis to other pretraining paradigms and transformer architectures. We consider DINOv3~\citep{simeoni2025dinov3}, a $7$B-parameter vision transformer trained in a self-supervised fashion, and GPT2~\citep{radford2019gpt2}, a $124$M-parameter decoder-only language model. We follow a similar setup as before, with pretraining and downstream data taken as ImageNet-22k and Cifar10 for DINOv3, and AGNews~\citep{zhang2015agnews} and WikiText-103~\citep{merity2017pointer} for GPT2; details on the choice of datasets are given in~\cref{app:plasticity_dinov3_gpt2}). We display the distribution plasticity on DINOv3 and GPT2 in~\cref{fig:plasticity_ablation} (see~\cref{fig:plasticity_ablation_all} for the evolution across layers). The observed patterns are consistent with those of supervised ViTs shown in~\cref{fig:analysis_sketch}, showcasing the generality of our findings across types of pretraining and transformer architectures. 
\begin{rmk}[Impact of distillation and alignment]
Since the plasticity of transformer components heavily depends on the model's weights (see~\cref{sec:theory}), different values may be observed for distilled or instruct-version models.
\end{rmk}
\subsection{Benefits of plasticity for finetuning}
\label{sec:finetuning}
In this section, each transformer component is finetuned in isolation along the depth of ViT-Base, leading to the $5$ configurations in~\cref{tab:model_configurations}. The optimization is done with SGD following the protocol of~\citet{dosovitskiy2021vit} summarized in~\cref{tab:training_details}. We conduct a sweep over $4$ well-spaced learning rates to ensure a fair comparison of modules with different numbers of trainable parameters. Each experiment is done over $3$ seeds, leading to a total of $\sim 1000$ finetuning runs. The experimental details are given in~\cref{app:finetuning}.
\begin{figure*}[!t]
    \centering
    \includegraphics[width=\linewidth]{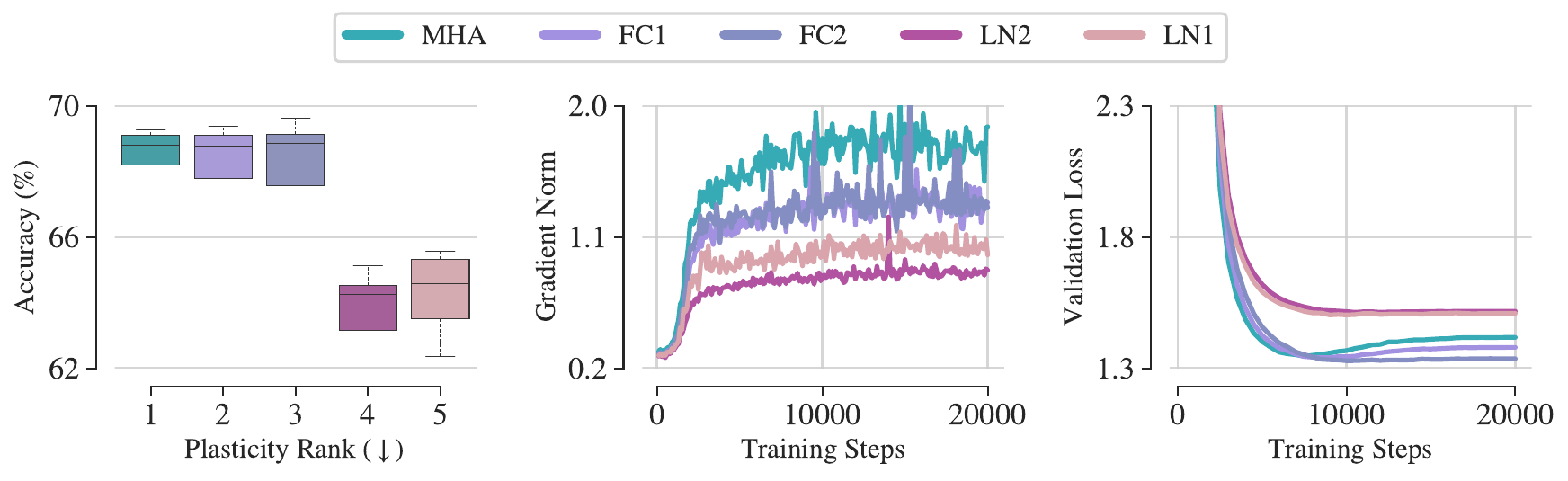}
    \caption{\textbf{Benefits of plasticity with SGD on Sketch.} Transformer components are ordered in terms of decreasing plasticity. We can see that the performance across learning rates and seeds (\textbf{left}) is better and more stable for plastic components. This can be understood by looking at the evolution of the gradient norms (\textbf{middle}) and the validation loss (\textbf{right}) throughout training: we can see that the higher plasticity, the larger gradient norms, and the better the generalization.}
    \label{fig:robustness_training}
\vspace{-1.2em}
\end{figure*}
\paragraph{Better performance.} The overall performance on all the $11$ benchmarks is displayed in~\cref{tab:statistical_test} (full results in~\cref{tab:results}, deferred to~\cref{app:finetuning_exp} due to space constraints). We observe that the attention modules and feedforward layers with high plasticity lead to enhanced finetuning performances, surpassing the LayerNorms on most benchmarks. The benefit of plasticity is even more salient on challenging datasets such as Cifar100, Clipart, and Sketch, where MHA, FC1, and FC2 surpass LN1 and LN2 by a large margin. We can see that the performance ordering is aligned with the \emph{plasticity ranking} from~\cref{sec:analysis}: the attention modules and feedforward layers result in higher accuracy than the LayerNorms. These results are consistent with~\cref{fig:overview} (right), where we display the relative gain, i.e., the percentage improvement of the finetuning accuracy over the linear probing accuracy. We can see that the ranking is also respected among components of the same size, namely the attention modules and the feedforward layers on the one hand, and the LayerNorms on the other hand. This conclusion holds at a larger scale, as shown on Clipart in~\cref{tab:base_vs_large} with ViT-Base ($86$M parameters) and ViT-Large ($307$M parameters). 
\begin{table}[!h]
\centering
    \caption{\textbf{Better finetuning performance (11 benchmarks)}. We report the average top-$1$ accuracy (\%) on the test set over 11 diverse image classification benchmarks ($\uparrow$). Transformer components are ordered in decreasing order of plasticity. The best performance among components is in \textbf{bold}, non-smooth components are highlighted in \colorbox{tablegray}{gray}, and \underline{underlined} entries indicate a statistically significant difference with MHA according to a Wilcoxon signed-rank test at a confidence level of $5\%$. Full results are in~\cref{tab:results}.}
    \label{tab:statistical_test}
    \begin{NiceTabular}{lccccc}
    \CodeBefore
    \columncolor{tablegray}{2-4} 
    \rowcolor{white}{1}
    \Body
    configuration & MHA & FC1 & FC2 & LN2 & LN1 \\
    \toprule[\thick pt]
    \emph{accuracy} &  \textbf{90.8} & 90.7 & \underline{90.3} & \underline{89.9} & \underline{89.8} \\
    \end{NiceTabular}
\vspace{-1.2em}
\end{table}
\paragraph{A strong baseline.} We can also see in~\cref{tab:statistical_test} that, except for FC1, the improvement of finetuning the MHA is statistically significant compared to the other modules. The adaptability of the attention module to downstream data is reminiscent of~\citet[Section 4]{touvron2022three}, where the authors found that tuning the attention module alone can be beneficial for ViTs models of varying sizes, ranging from $6$M to $340$M parameters; it notably surpasses the full finetuning baseline on small datasets. This showcases the relevance of our plasticity analysis in understanding the behavior of vision transformers. Another interesting insight of our work is that single-component finetuning is competitive with parameter-efficient methods such as LoRA~\citep{hu2022lora}. For instance, finetuning only the LayerNorms with Adam outperforms LoRA \emph{with $15\times$ fewer trainable parameters} on Cifar100: $89.4\%$ with $28$K parameters (see~\cref{tab:adam_performance}) compared to $88.1\%$ with $400$K parameters (see~\citep[Table 1]{bafghi2024loravit}). As a byproduct, it also shows that parameter count alone does not determine the performance.

\paragraph{Robust adaptation.} The absolute best performance is not the only factor to take into account: being robust to the choices of hyperparameters is also of great importance for practitioners. The main source of variability during finetuning comes from the initialization and the learning rate. In~\cref{fig:robustness_training} (left), we report the distribution of top-$1$ accuracy over the grid of learning rates (see~\cref{tab:training_details}) and $3$ seeds. We can see that the finetuning performance of the components with high plasticity is steadier. In particular, the attention module has the smallest variability. This pattern remains consistent overall, notably for the multi-head self-attention, as can be seen in~\cref{fig:robustness_all}. Our findings hint at the benefits of plasticity during the optimization process discussed in the next paragraph.

\paragraph{Interplay between plasticity and optimization.} In~\cref{sec:intro}, we argued that, given the interplay between input-output smoothness and weight-output smoothness, a high plasticity should lead to large gradient norms. In~\cref{fig:robustness_training}, we display the evolution of the gradient norms (middle) and the validation loss (right) for the finetuning run on Sketch that achieves the highest accuracy (this corresponds to learning rate $\eta=1\mathrm{e-}2$). This confirms our intuition: the ordering of gradient norms is aligned with the \emph{plasticity ranking} established in~\cref{sec:analysis}. In tandem, the loss descent is steeper for components with high plasticity, such as the attention modules and the feedforward layers. These patterns are consistent across benchmarks, learning rates, and seeds (see~\cref{fig:training_evolution_cifar10_seed_0,fig:training_evolution_cifar10_seed_42,fig:training_evolution_cifar10_seed_3407,fig:training_evolution_cifar100_seed_0,fig:training_evolution_cifar100_seed_42,fig:training_evolution_cifar100_seed_3407,fig:training_evolution_cifar10_c_contrast_seed_0,fig:training_evolution_cifar10_c_contrast_seed_42,fig:training_evolution_cifar10_c_contrast_seed_3407,fig:training_evolution_cifar10_c_gaussian_noise_seed_0,fig:training_evolution_cifar10_c_gaussian_noise_seed_42,fig:training_evolution_cifar10_c_gaussian_noise_seed_3407,fig:training_evolution_cifar10_c_motion_blur_seed_0,fig:training_evolution_cifar10_c_motion_blur_seed_42,fig:training_evolution_cifar10_c_motion_blur_seed_3407,fig:training_evolution_cifar10_c_snow_seed_0,fig:training_evolution_cifar10_c_snow_seed_42,fig:training_evolution_cifar10_c_snow_seed_3407,fig:training_evolution_cifar10_c_speckle_noise_seed_0,fig:training_evolution_cifar10_c_speckle_noise_seed_42,fig:training_evolution_cifar10_c_speckle_noise_seed_3407,fig:training_evolution_domainnet_clipart_seed_0,fig:training_evolution_domainnet_clipart_seed_42,fig:training_evolution_domainnet_clipart_seed_3407,fig:training_evolution_domainnet_sketch_seed_0,fig:training_evolution_domainnet_sketch_seed_42,fig:training_evolution_domainnet_sketch_seed_3407,fig:training_evolution_flowers102_seed_0,fig:training_evolution_flowers102_seed_42,fig:training_evolution_flowers102_seed_3407,fig:training_evolution_pet_seed_0,fig:training_evolution_pet_seed_42,fig:training_evolution_pet_seed_3407}), which confirms the role of plasticity during the optimization process illustrated in~\cref{fig:intro}. Our findings are reminiscent of the intuition that ResNet layers with larger gradient magnitudes carry more information about the target data~\citep[see][Section 4]{lee2023surgical}. The benefits of plasticity on the learning process are in accordance with the empirical evidence from~\cref{fig:overview} (right). The conclusion of the finetuning analysis can be summarized as:
\begin{tcolorbox}[colback=\boxcol,
    colframe=black,
    arc=4pt,
    boxsep=0pt,
    boxrule=\boxwidth pt,
]%
\textbf{Takeaway 3.} A higher plasticity facilitates the optimization and leads to better and more stable finetuning performance. Our findings indicate that the components to prioritize during finetuning should be the attention module and the first feedforward linear layer.
\end{tcolorbox}
\vspace{-.1em}
\paragraph{Extension to Adam.} We extend the finetuning analysis with Adam~\citep{kingma2014adam} (more precisely with its decoupled weight-decay version from~\citet{loshchilov2018decoupled}), which became the de facto choice to train foundation models~\citep{simeoni2025dinov3}, notably LLMs~\citep{orvieto2026adam,grattafiori2024llama3herdmodels}. We follow the same setup as with SGD and rescale learning rates by $10^{-2}$ as is standard in the literature on adaptive methods~\citep{kumar2023finetunevisionmodelssgd, dosovitskiy2021vit,touvron2021deit}. Full implementation details are given in~\cref{app:adam_exp}. We observe similar benefits than with SGD: finetuning non-smooth components consistently yields higher accuracy and superior stability across learning rates compared to smoother modules. The robust adaptation and the interplay between plasticity and optimization are similar between Adam and SGD, as illustrated in~\cref{fig:robustness_training_adam} on Sketch, where the behavior is consistent with~\cref{fig:robustness_training}.
\begin{table}[!t]
\centering
    \caption{\textbf{Consistent benefits with Adam}. We report the average top-$1$ accuracy (\%) on the test set over the learning rate grid of each dataset ($\uparrow$). Transformer components are ordered in decreasing order of plasticity. The best performance among components is in \textbf{bold} and non-smooth components are highlighted in \colorbox{tablegray}{gray}.}
    \label{tab:adam_performance}
    \scalebox{.75}{
    \begin{NiceTabular}{lccccc}
    \CodeBefore
    \columncolor{tablegray}{2-4} 
    \rowcolor{white}{1}
    \Body
    configuration & MHA & FC1 & FC2 & LN2 & LN1 \\
    \toprule[\thick pt]
    Cifar100 & 91.0 $_{\tiny \pm \text{0.2}}$ & \textbf{91.3} $ _{\tiny \pm \text{0.6}}$ & 90.6 $ _{\tiny \pm \text{1.4}}$ & 89.4 $ _{\tiny \pm \text{2.7}}$ & 88.4 $ _{\tiny \pm \text{3.2}}$ \\
    Motion Blur & \textbf{92.4} $_{\tiny \pm \text{0.2}}$ & 91.8 $ _{\tiny \pm \text{0.4}}$ & 90.7 $ _{\tiny \pm \text{0.9}}$ & 91.7 $ _{\tiny \pm \text{2.4}}$ & 90.7 $ _{\tiny \pm \text{2.4}}$ \\
    Clipart & \textbf{76.4} $_{\tiny \pm \text{0.5}}$ & 74.8 $ _{\tiny \pm \text{0.8}}$ & 75.8 $ _{\tiny \pm \text{0.3}}$ & 73.3 $ _{\tiny \pm \text{1.9}}$ & 73.5 $ _{\tiny \pm \text{1.5}}$ \\
    Sketch & 69.0 $_{\tiny \pm \text{0.5}}$ & 67.9 $ _{\tiny \pm \text{0.6}}$ & \textbf{69.2} $ _{\tiny \pm \text{0.4}}$ & 63.1 $ _{\tiny \pm \text{2.9}}$ & 64.0 $ _{\tiny \pm \text{2.6}}$ \\
    \midrule[\midthick pt]
    Average & \textbf{82.2} & 81.4 & 81.7 & 79.4 & 79.2 \\
    \end{NiceTabular}
    }
\end{table}
\paragraph{Extension to low-data regimes.} We extend the finetuning experiments to low-data regimes by following the standard Visual Task Adaptation Benchmark protocol~\citep[VTAB]{zhai2020vtab}, where training is restricted to $1,000$ samples per dataset. Following the original ViT paper setup~\citep{dosovitskiy2021vit}, we finetuned models for $2,500$ steps on $1,000$ training samples. Results are displayed in~\cref{tab:low_data_regime}. We can see that prioritizing high-plasticity components, especially the attention module MHA, yields overall superior performance compared to low-plasticity components (LN1, LN2). This remains consistent with the full-dataset setting.

\section{Related work}
\label{sec:related_work}
\paragraph{Smoothness.} Smoothness has been studied extensively in deep learning, e.g., in generalization~\citep{rosca2020smooth,bartlett2017spectral,jukic2025robustness}, training stability~\citep{zhai2023sigmareparam}, generative modeling~\citep{miyato2016distributionalsmoothingvirtualadversarial,szegedy2014intriguing}, adversarial robustness~\citep{tsuzuku2018lipschitz,weng2018evaluating,hein2017robustness,jia2024lipschitz}, and differential privacy~\citep{bethune2024dpsgd}. Common practices in deep learning, such as weight decay~\citep{hanson1988wd}, dropout~\citep{srivastava2014dropout}, and early stopping~\citep{hardt2016earlystopping}, encourage smoothness. We extend this discussion in~\cref{app:related_work}. In our work, we identify the components with low smoothness and showcase the benefits of ``non-smooth" components for finetuning. However, we do not promote smoothness during the learning process in any way.

\paragraph{Lipschitz constant estimation.} Estimating the Lipschitz constant of neural networks is a hard problem~\citep{virmaux2018lipschitz}. Theoretical bounds are often loose, except for simple blocks such as linear maps and activation~\citep{bethune2024dpsgd}. For transformers, the non-linear nature of self-attention makes the estimation more involved. Notably,~\citet{kim2021lipschitz} showed that vanilla attention is not globally Lipschitz. Tight upper bounds have been obtained when restricted to sequences of bounded tokens~\citet{castin2024smooth}. Imposing Lipschitz constraints is a common way to promote smoothness~\citep{rosca2020smooth,newhouse2025lipschitz}. Note that the proof techniques to derive the upper bounds in~\cref{sec:theory} are akin to those used to bound the Lipschitz constant of the modules.
\begin{table}[!t]
\centering
    \caption{\textbf{Consistent benefits in low-data regimes.} Study akin to~\cref{tab:adam_performance} but finetuning is done on only $1,000$ samples per dataset (VTAB protocol). Transformer components are ordered in decreasing order of plasticity. The best performance among components is in \textbf{bold} and non-smooth components are highlighted in \colorbox{tablegray}{gray}.}
    \label{tab:low_data_regime}
    \scalebox{.75}{
    \begin{NiceTabular}{lccccc}
    \CodeBefore
    \columncolor{tablegray}{2-4} 
    \rowcolor{white}{1}
    \Body
    configuration & MHA & FC1 & FC2 & LN2 & LN1 \\
    \toprule[\thick pt]
    Contrast & \textbf{96.7} $_{\tiny \pm \text{0.7}}$ & 96.1 $_{\tiny \pm \text{1.1}}$ & 95.4 $_{\tiny \pm \text{1.1}}$ & 95.8 $_{\tiny \pm \text{1.1}}$ & 95.6 $_{\tiny \pm \text{1.3}}$ \\
    Motion Blur & \textbf{93.6} $_{\tiny \pm \text{1.2}}$ & 92.7 $_{\tiny \pm \text{12.3}}$ & 92.0 $_{\tiny \pm \text{2.4}}$ & 91.8 $_{\tiny \pm \text{2.5}}$ & 90.7 $_{\tiny \pm \text{2.7}}$ \\
    Pets & \textbf{92.5} $_{\tiny \pm \text{0.3}}$ & 92.4 $_{\tiny \pm \text{0.7}}$ & 92.2 $_{\tiny \pm \text{0.8}}$ & 92.5 $_{\tiny \pm \text{0.9}}$ & 92.1 $_{\tiny \pm \text{0.9}}$ \\
    Snow & \textbf{94.9} $_{\tiny \pm \text{0.6}}$ & 94.1 $_{\tiny \pm \text{1.9}}$ & 93.4 $_{\tiny \pm \text{2.1}}$ & 93.7 $_{\tiny \pm \text{1.8}}$ & 93.4 $_{\tiny \pm \text{1.9}}$ \\
    \midrule[\midthick pt]
    Average & \textbf{94.4} & 93.8 & 93.3 & 93.4 & 93.0\\
    \end{NiceTabular}
    }
\end{table}
\paragraph{Parameter-efficient finetuning.} There exists a plethora of PEFT methods~\citep{zhang2025survey}. Our work is in line with the selective approaches, common in vision models, where only a subset of the parameters is finetuned~\citep{guo2019spottune,lee2023surgical,xu2021raisechildlargelanguage,liu2021autofreezeautomaticallyfreezingmodel,lee2019elsadofreezinglayers,wang2021tent}. Another widely used category consists of additive methods, where small adapters, such as normalization layers, are inserted in the model~\citep{houlsby2019peft,pfeiffer2021adapter,lian2022shifting}. The well-known LoRa~\citep{hu2022lora} method belongs to the reparameterization methods, where the weights are decomposed and reparameterized to adapt to fewer parameters. While those approaches are performance-oriented and often tune several types of modules together, we conduct a component-wise analysis with the aim of theoretically understanding the adaptability of each transformer module. 

\section{Discussion} 
This paper investigates the plasticity of vision transformer components by analyzing their average smoothness. Our theoretical and empirical analysis demonstrate the benefits of this approach to identify the components to prioritize during finetuning. In particular, finetuning non-smooth components (with high plasticity), namely the attention modules and the feedforward layers, consistently results in better and more stable performance. Our findings offer a novel perspective on the role of smoothness in finetuning transformers. We hope our work can help the design of more efficient adaptation methods and contribute to the effort towards better understanding the transformer architecture~\citep[see, e.g.,][]{zekri2025largelanguagemodelsmarkov,raghu2021do,vonoswald2023icl,jelassi2022vision}.
 
\paragraph{Limitations and future work.} To extend our analysis, a promising approach would be to study the effect of a tailored optimization (e.g., adaptive learning rates and scheduler) on the performance of each module. Moreover, since our methodology and theoretical insights naturally extend to decoder-only models, our paper could serve as groundwork to explore the adaptability and plasticity of LLMs.

\section*{Acknowledgements}
The authors would like to thank Zehao Xiao, Abdelhakim Benechehab, Vasilii Feofanov, and Albert Thomas for insightful comments about early versions of this work, as well as Th\'eo Moutakanni and Guillaume Carlier for fruitful discussions that led to this project. The authors would also like to thank the anonymous reviewers and the area chair for their time and constructive feedback that helped us improve our work.

\section*{Impact statement}
This paper presents work whose goal is to advance the field of Machine Learning. There are many potential societal consequences of our work, none of which we feel must be specifically highlighted here.

\bibliography{references}
\bibliographystyle{template/icml2026}

\clearpage
\appendix
\onecolumn
\textbf{\LARGE Appendix}
\paragraph{Roadmap.} In this appendix, we discuss additional related work in~\cref{app:related_work}. We detail the proofs of our theoretical results in~\cref{app:proofs}. The open-source code and carbon footprint of the project are given in~\cref{app:reproducibility}. We provide the full implementation details in~\cref{app:implem} and, finally, extensive additional experiments are provided in~\cref{app:exp}. We display the corresponding table of contents below.

\addtocontents{toc}{\protect\setcounter{tocdepth}{3}}

\renewcommand*\contentsname{\Large Table of Contents}
\tableofcontents
\clearpage

\section{Extended related work}
\label[secinapp]{app:related_work}
In this section, we extend the discussion of prior works related to our paper.
\paragraph{Smoothness in neural networks.} Neural networks smoothness, typically quantified via Lipschitz constants and spectral norms, has been studied in the context of in-domain generalization~\citep{neyshabur2017generalization,rosca2020smooth,sokolic2017robust,bartlett2017spectral,luxburg2004lipschitz,jukic2025robustness,novak2018sensitivity}, training stability~\citep{zhai2023sigmareparam,miyato2018spectral}, generative modeling~\citep{miyato2016distributionalsmoothingvirtualadversarial,szegedy2014intriguing}, adversarial robustness~\citep{rosca2020smooth,tsuzuku2018lipschitz,weng2018evaluating,anil2019lip,hein2017robustness,jia2024lipschitz} and differential privacy~\citep{bethune2024dpsgd}. \citet{neyshabur2017generalization} discusses the interplay between complexity measures based on norms, margin control, Lipschitz constants, and sharpness. These works discuss the benefits of promoting smoothness by regularizing Lipschitz constants~\citep{rosca2020smooth} or spectral norms~\citep{zhai2023sigmareparam} during training. Common practices in deep learning such as weight decay~\citep{krogh1991wd,hanson1988wd,bartlett1997wd}, dropout~\citep{srivastava2014dropout} and early stopping~\citep{hardt2016earlystopping} encourage neural networks smoothness. 

\paragraph{Smoothness at scale.} Recently, smoothness has been studied in the context of stabilizing large models, such as LLMs~\citep{zhai2023sigmareparam}. During training at such a large scale, many instabilities appear and notable loss spikes~\citep{chowdhery2023palm,marin,alej2025apertus}. They can cause the model to diverge and have been the subject of many studies. Inspired by mechanistic interpretability~\citep{elhage2021mathematical}, companion work proposed to study the gradient descent of the transformer~\citep{odonnat2025clusteringhead,odonnat2025easing} on the sparse modular addition problem~\citep{nanda2023progress}. It provides a simple yet sufficient testbed to observe involved optimization dynamics at a small scale. In more realistic settings, methods to control the gradient norms have been proposed, such as QK-Norm~\citep{dehghani2023vit22b,wortsman2024smallscale}, or constraining the representation space on the hypersphere~\citep{loshchilov2025ngpt}, and are used to train industry-level LLMs. These approaches are reminiscent of generalization bounds based on margins and spectral norms~\citep{bartlett2017spectral,neyshabur2017generalization}. Recently, the Muon optimizer~\citep{jordan2024muon}, that normalize the spectral norms of the weights, has shown tremendous benefits in solving training instabilities. In addition,~\citet{newhouse2025lipschitz} that Muon allows for optimizing Lipschitz-constrained neural networks at scale. The combination of these approaches can also be beneficial: the Kimi team managed to train on over $1$T tokens without any loss spikes thanks to MuonClip~\citep{team2025kimi} that combines Muon with QK-Norm.

\paragraph{Lipschitz constant estimation.} A lot of effort has been put into estimating the Lipschitz constants of neural networks. While linear and activation layers have a known tight Lipschitz constant~\citep{virmaux2018lipschitz,castin2024smooth,bethune2024dpsgd}, estimating the Lipschitz constant of feedforward networks is NP-hard~\citep{virmaux2018lipschitz} with loose theoretical upper bounds~\citep{virmaux2018lipschitz}. The non-linear nature of the self-attention module makes the estimation of its Lipschitz constant more involved. \citet{kim2021lipschitz} showed that the vanilla attention is not globally Lipschitz, while the dot-product attention is. Tight upper bounds on the attention module restricted to sequences of bounded tokens have been provided in~\citet{castin2024smooth}. \citet{dasoulas2021lip} showed the benefits of enforcing Lipschitz continuity in self-attention for graph neural networks. Imposing Lipschitz constraints on neural networks has also been done in generative modeling~\citep{arjovsky2017wgan} or to ensure robustness and explainability, e.g., with $1$-Lipschitz neural networks~\citep{serrurier2023on, serrurieret2021robustness}. 

\paragraph{Parameter-efficient finetuning.} PEFT methods can be categorized into $5$ main families~\citep{zhang2025survey}. Selective methods, common in vision models, aim to finetune only a subset of the parameters~\citep{guo2019spottune,lee2023surgical,xu2021raisechildlargelanguage,liu2021autofreezeautomaticallyfreezingmodel,lee2019elsadofreezinglayers,wang2021tent}. Additive methods insert small adapter networks between the model's layers to be trained during finetuning~\citep{houlsby2019peft,pfeiffer2021adapter,lian2022shifting}. Prompt methods, commonly used for large language models~\citep{gu2022ppt,liu2023gptunderstands,li2021prefix}, involve learning soft commands to guide the model. Reparameterization methods, such as LoRa~\citep{hu2022lora} and its companions~\citep{gu2023mixofshow,zi2023deltalorafinetuninghighrankparameters}, decompose and reparameterize the model's weights to adapt fewer parameters. These methods can be combined, leading to the last family of hybrid approaches~\citep{mao2022pelt}. We note that the recent shift of large language models from static predictors towards dynamic, context-aware agents redefines finetuning methods. The benefits of learning to use tools instead of incorporating the knowledge in the model weights have been demonstrated in~\citep{houliston2025provable}. This explain the superiority and scalability of approaches such as Toolformer~\citep{schick2023toolformer}, Retrieval-Augmented Generation~\citep[RAG,][]{lewis2020rag} and a plethora of other approaches~\citep{qu2025tool}. 

\section{Proofs}
\label[secinapp]{app:proofs}
In this section, we detail the proofs of our theoretical results, which involve simple manipulations of matrix norms. 

\paragraph{Notations.} Throughout the paper, we use the notation $[n]$ to represent the set $\{1, \ldots, n\}$. The Euclidean norm of $\RR^n$ is denoted by $\|\cdot \|$ and its $\ell_\infty$ norm is denoted by $\|\cdot\|_\infty$. Entries of a matrix $A \in \RR^{n \times m}$ write $A_{ij}$, its rows write $A_i$ and its columns write $A_{\cdot, j}$. The Frobenius norm of a matrix $A \in \RR^{n \times m}$ writes $\|A\|_\mathrm{F} = \mleft(\sum_{i=1}^n \sum_{j=1}^m A_{ij}^2\mright)^{1/2}$ and its spectral norm writes $\|A\|_2 = \sigma_\mathrm{max}(A)$, defined as the largest singular value of $A$. We denote by $B_r \subset \RR^d$ the closed ball centered at $0$ with radius $r>0$. 

\paragraph{Useful properties.} The next lemma recalls some well-known properties of the Frobenius norm and its connection with the spectral norms~\citep[see][p.364, Section 5.6.P20]{horn2012matrix}, which will be used in our proofs.
\begin{boxlem}
\label{lem:frob_spectral}
For any matrices $A \in \RR^{n \times m}$ and $B \in \RR^{m \times p}$, we have
\begin{equation*}
    \begin{cases}   
        &\|AB\|_\mathrm{F} \leq \|A\|_\mathrm{F} \|B\|_\mathrm{F} \\
        &\|AB\|_\mathrm{F} \leq \|A\|_2 \|B\|_\mathrm{F} \\
        &\|AB\|_\mathrm{F} \leq \|A\|_\mathrm{F} \|B\|_2,
    \end{cases}
\end{equation*}
where the first property is referred to as the submultiplicativity of the Frobenius norm.
\end{boxlem}
\begin{proof}
Let $C = AB$. The entries of $C$ writes $C_{ij} = \sum_{k=1}^m A_{ik}B_{kj} = A_i ^\top B_{\cdot, j}$. Applying Cauchy-Schwartz leads to:
\[
\|C_{ij}\|^2 = \|A_i ^\top B_{\cdot, j}\|^2 \leq \|A_i\|^2 \|B_{\cdot, j}\|^2.
\]
Hence, the Frobenius norm of $AB$ verifies
\begin{align*}
\|AB \|_\mathrm{F}^2 = \|C\|_\mathrm{F}^2 &= \sum_{i=1}^n \sum_{j=1}^p \|C_{ij}\|^2\\
&\leq \sum_{i=1}^n \sum_{j=1}^p \|A_i\|^2 \|B_{\cdot, j}\|^2 \\
&= \sum_{i=1}^n \|A_i\|^2 \sum_{j=1}^p \|B_{\cdot, j}\|^2 \\
&= \sum_{i=1}^n\sum_{k=1}^m A_{ik}^2 \sum_{j=1}^p \sum_{l=1}^m B_{lj}^2 \\
&= \sum_{i=1}^n\sum_{k=1}^m A_{ik}^2 \sum_{l=1}^m\sum_{j=1}^p B_{lj}^2 \\
&= \|A\|_\mathrm{F}^2 \|B\|_\mathrm{F}^2.
\end{align*}
Taking the square root concludes the proof by monotonicity. For the second result, using the same notations, we recall that the columns of $C$ write
\[
C_{\cdot, j} = A(B_{\cdot, j}).
\]
Recalling that the spectral norm $\|\cdot\|_2$ is the operator norm induced by $\|\cdots\|$ on $\RR^n$, it leads to
\begin{align*}
\|AB\|_\mathrm{F} = \|C\|_\mathrm{F} &= \sum_{i=1}^n\sum_{j=1}^p C_{ij}^2 \\
&= \sum_{j=1}^p \|C_{\cdot, j}\|^2 \\
&= \sum_{j=1}^p \|A(B_\cdot, j)\|^2 \\
&\leq \sum_{j=1}^p \|A\|_2^2 \|(B_\cdot, j)\|^2 \tag{operator norm property of $\|\cdot\|_2$} \\
&= \|A\|_2^2 \sum_{j=1}^p \|B_{\cdot, j}\|^2 \\
&= \|A\|_2^2 \sum_{j=1}^p \sum_{l=1}^m B_{lj}^2 \\
&= \|A\|_2^2 \sum_{l=1}^m \sum_{j=1}^p B_{lj}^2 \\
&= \|A\|_2^2 \|B\|_\mathrm{F}^2.
\end{align*}
Taking the square root concludes the proof by monotonicity. For the last result, we use the previous one by using the fact that the Frobenius norm is the sum of singular values, which are invariant by transposition, and that the spectral norm is the maximal singular value. This leads to the Frobenius and the spectral norm to remain invariant under transposition. As such, we have
\[
\|AB\|_\mathrm{F} = \|(AB)^\top\|_\mathrm{F} = \|B^\top A^\top\|_\mathrm{F} \leq \|B^\top\|_2 \|A^\top\|_\mathrm{F} = \|A\|_\mathrm{F} \|B\|_2,
\]
which concludes the proof.
\end{proof}

\subsection{Proof of~\cref{prop:ln_bounds}}
\label[secinapp]{app:ln_bounds}
\begin{proof}
We start by upper-bounding the plasticity of LayerNorms. Let $f$ be a LayerNorm with weights $\gamma, \beta \in \RR^d$ and let $\nu$ be the uniform distribution over the set of distinct pairs of sequences of tokens in $(\RR^d)^n$. Let $(x, y)$ be a pair of two distinct sequences of tokens sampled according to $\nu$. By assumption, we have for any $i \in [n]$ that
\[
\forall 1 \leq i \leq n, \quad \mu(x_i) = \mu(y_i) = \mu_i \text{ and } \sigma(x_i) = \sigma(y_i) = \sigma_i > 0,
\]
with $\mu(x_i), \sigma(x_i)$ (respectively $\mu(y_i), \sigma(y_i)$) the mean and standard deviation of the token $x_i \in \RR^d$ (respectively of $y_i)$. From the definition of LayerNorm (see~\cref{sec:background}), it leads to
\begin{align*}
f(x) - f(y) &= \mleft( \gamma \odot \frac{x_1 - \mu(x_1)}{\sigma(x_1)} + \beta , \ldots, \gamma \odot \frac{x_n - \mu(x_n)}{\sigma(x_n)} + \beta\mright) \\
& \qquad - \mleft( \gamma \odot \frac{y_1 - \mu(y_1)}{\sigma(y_1)} + \beta , \ldots, \gamma \odot \frac{y_n - \mu(y_n)}{\sigma(y_n)} + \beta\mright) \\
&= \mleft( \gamma \odot \frac{x_1 - \mu_1}{\sigma_1} + \beta , \ldots, \gamma \odot \frac{x_n - \mu_n}{\sigma_n} + \beta\mright) \\
& \qquad - \mleft( \gamma \odot \frac{y_1 - \mu_1}{\sigma_1} + \beta , \ldots, \gamma \odot \frac{y_n - \mu_n}{\sigma_n} + \beta\mright) \\
&= \mleft( \gamma \odot \frac{x_1-y_1}{\sigma_1}, \ldots, \gamma \odot \frac{x_n-y_n}{\sigma_n}\mright).
\end{align*}
Denoting $\tilde{x}$, respectively $\tilde{y}$, the sequence with entries $\mleft(\frac{x_i}{\sigma_i}\mright)_{i=1}^n$, respectively $\mleft(\frac{y_i}{\sigma_i}\mright)_{i=1}^n$, we have
\begin{align*}
    \|f(x) - f(y) \|_\mathrm{F} &= \| \gamma \odot (\tilde{x} - \tilde{y}) \|_\mathrm{F}\\
    &= \| \Gamma (\tilde{x} - \tilde{y}) \|_\mathrm{F} \\
    &\leq \|\Gamma\|_2 \|\tilde{x} - \tilde{y}\|_\mathbf{F},
\end{align*}
where the second line comes from defining $\Gamma \in \RR^{d \times d}$ as a diagonal matrix with values the entries of $\gamma \in \RR^d$ and replacing the element-wise product by a matrix product, and the last line comes from using~\cref{lem:frob_spectral}. We recall that 
\[
\|\tilde{x} - \tilde{y}\|_\mathbf{F}^2 = \sum_{i=1}^n \| (x_i - y_i) / \sigma_i \|_2 \leq \mleft(\frac{1}{\min_{i=1}^n \sigma_i}\mright)^2 \sum_{i=1}^n \|(x_i - y_i) \|_2 = \frac{1}{\sigma^2} \|x - y\|_\mathbf{F}^2 ,
\]
where $\sigma = \min_{i=1}^n \sigma_i > 0$, and that $\|\Gamma\|_2 = \max_{i=1}^d |\gamma_i| = \|\gamma\|_\infty $ by the definition of the spectral norm of a diagonal matrix. We then obtain 
\[
\|f(x) - f(y) \|_\mathrm{F} \leq \frac{1}{\sigma}\|\gamma\|_\infty \|x - y\|_\mathbf{F}.
\]
Since the result holds for randomly sampled sequences $x, y$, and by assumption, the $\sigma_i$ depend on the embedding layer and not on the input sequences, we can upper bound the rate of change and take the expectation over distinct sequences of tokens $x, y$. We have
\[
\mathcal{P}(f) = \EE_{(x, y) \sim \mu} \mleft[\frac{\| f(x) - f(y)\|_\mathrm{F}}{\|x-y\|_\mathrm{F}}\mright] \leq \frac{1}{\sigma}\|\gamma\|_\infty,
\]
which concludes the proof for the LayerNorms. 
\end{proof}

\subsection{Proof of~\cref{prop:fc_bounds}}
\label[secinapp]{app:fc_bounds}
\begin{proof}
The proof derivation is simple for the case of linear layers. We detail it below for consistency. Let $f$ be a linear layer with weights $W_1 \in \RR^{d \times 4d}$ and let $\nu$ be the uniform distribution over the set of distinct pairs of sequences of tokens in $(\RR^d)^n$. Let $(x, y)$ be a pair of two distinct sequences of tokens sampled according to $\nu$. Let $x, y \in \RR^{n \times d}$ be two distinct sequences of tokens sampled according to $\mu$.. By definition of the linear layer, a simple application of~\cref{lem:frob_spectral} leads to
\[
\| f(x) - f(y)\|_\mathbf{F} = \| W_1 (x-y)\|_\mathrm{F} \leq \|W_1\|_2 \|x-y\|_\mathbf{F}.
\]
We obtain a similar result for the second linear layer with weights in $\RR^{4d \times d}$. As in~\cref{app:ln_bounds}, since the upper bound holds for randomly sampled sequences $x, y$, we can bound the rate of change and take the expectation to conclude the proof.
\end{proof}

\subsection{Proof of~\cref{prop:mha_bounds}}
\label[secinapp]{app:mha_bounds}
\begin{proof}
Let $f$ be a multihead self-attention module with weights $(O^h, Q^h, K^h, V^h)_{1 \leq h \leq H}$, with $A^h = (Q^h)^\top K^h / \sqrt{k}$, and let $\nu$ be the uniform distribution over the set of distinct pairs of sequences of tokens in $(\RR^d)^n$. Let $(x, y)$ be a pair of two distinct sequences of tokens sampled according to $\nu$. Let $x, y \in \RR^{n \times d}$ be two distinct sequences of tokens sampled according to $\mu$. By the definition of multihead self-attention (see~\cref{sec:background}), we have
\begin{align*}
\|f(x) - f(y)\|_\mathrm{F} &= \| \sum_{h=1}^H O^h \mleft(f^h_\mathrm{att}(x) - f^h_\mathrm{att}(x) \mright)\|_\mathrm{F} \\
& \leq \sum_{h=1}^H \|O^h \mleft(f^h_\mathrm{att}(x) - f^h_\mathrm{att}(x) \mright)\|_\mathrm{F} \tag{triangular inequality} \\
&\leq \sum_{h=1}^H \|O^h\|_2 \| f^h_\mathrm{att}(x) - f^h_\mathrm{att}(x)\|_\mathrm{F} \tag{\cref{lem:frob_spectral}}.
\end{align*}
We recall that following~\citet{castin2024smooth}, the Lipschitz constant of $f$ on $\mathcal{X} \subset \mleft( \RR^d\mright)^n$ writes
\begin{equation}
\label{eq:lip}
\mathrm{Lip}\mleft(f_{|\mathcal{X}}\mright) = \sup_{\substack{x, y \in \mathcal{X} \\ x \neq y}} \frac{\| f(x) - f(y)\|_\mathrm{F}}{\|x-y \|_\mathrm{F}}.
\end{equation}
We recall the following result on the Lipschitz constant of self-attention.
\begin{boxthm}{\citep[Theorem 3.3]{castin2024smooth}}
\label{thm:castin}
Let $Q, K, V \in \RR^{k \times d}$ and $A = Q^\top K / \sqrt{k}$. Let $r>0$ and $n \in \NN$. A self-attention module $f_\mathrm{att}$ with weights $Q, K, V$ is Lipschitz continuous on $B_r^n$, with 
\[
\mathrm{Lip}\mleft(f_\mathrm{att}{|B_r^n}\mright) \leq \sqrt{3} \|V\|_2\sqrt{\|A\|_2^2r^4(4n+1) + n}.
\]
\end{boxthm}  
By assumption, the sequences of tokens are restricted to $B_r^n$. We can thus apply~\cref{thm:castin} on each self-attention module $f_\mathrm{att}^h$ with weights $Q^h,K^h,V^h$. Using the fact that the Lipschitz constant in~\cref{eq:lip} is a supremum over individual rates of change, we have
\begin{align*}
\|f(x) - f(y)\|_\mathrm{F} &\leq \sum_{h=1}^H \|O^h\|_2 \cdot \mathrm{Lip}\mleft(f_\mathrm{att}^h{|B_r^n}\mright) \|x-y\|_\mathrm{F} \\
&\leq \mleft(\sum_{h=1}^H \|O^h\|_2  \sqrt{3} \|V^h\|_2\sqrt{\|A^h\|_2^2r^4(4n+1) + n} \mright) \|x-y\|_\mathrm{F}.
\end{align*}
As in~\cref{app:ln_bounds}, since the upper bound holds for randomly sampled sequences $x, y$, we can bound the rate of change and take the expectation to conclude the proof.
\end{proof}

\subsection{Proof of~\cref{prop:mha_tighter_bounds}}
\label[secinapp]{app:mha_tighter_bounds}
\begin{proof}
Let $f$ be a multihead self-attention module with weights $(O^h, Q^h, K^h, V^h)_{1 \leq h \leq H}$, with $A^h = (Q^h)^\top K^h / \sqrt{k}$, and let $\nu$ be the uniform distribution over the set of distinct pairs of sequences of tokens in $(\RR^d)^n$. Let $(x, y)$ be a pair of two distinct sequences of tokens sampled according to $\nu$. We first show that the assumption on the energy of images leads to sequences of tokens with bounded Frobenius norm. The digital image embedded to obtain the sequence of tokens $x$ can be seen as a discretization of its continuous intensity, denoted by $I \colon \Omega \subset \RR^2 \to \RR_+$ (for convenience, we consider a grayscale image, but similar derivations are straightforward for an RGB image). The total energy of the image is defined as the sum of the squared intensities over pixels~\citep[see][Chapter 1, page 2]{mallat2008wavelet}. In line with the signal processing literature~\citep{goodman2005introduction,mallat2008wavelet}, the image has a finite energy $\mathcal{E}_x \geq 0$, which is bounded by $\mathcal{E}$ by assumption. Summing over pixels, we have
\[
\mathcal{E}_x = \sum_{u, v \in \Omega} |I(u,v)|^2  \leq \mathcal{E} < \infty.
\]
In vision transformers, images are split into $n$ square patches of size $P$. The $i$-th patch, denoted by $p_i \in \RR^{P \times P}$, covers an area $\Omega_i$ and has an energy $\mathcal{E}_i \geq 0$. We have:
\[
\mathcal{E} = \sum_{u, v \in \Omega} |I(u,v)|^2 = \sum_{i=1}^n \sum_{u, v \in \Omega_i} |I(u,v)|^2 = \sum_{i=1}^n \mathcal{E}_i.
\]
Denoting by $x = (x_1, \ldots, x_n) \in (\RR^d)^n$ the sequence of tokens obtained after embedding the image, where the $i$-th token $x_i$ is obtained by flattening the $i$-th patch $p_i$ and linearly projecting it in $\RR^d$. Since the input images have dimensions $H \times W \times C$~\citep[see][Section 3.1]{dosovitskiy2021vit}, the flattened patches have a dimension of $m = P^2 \times C$. We denote by $E \in \RR^{d \times m}$ the weights of the embedding layer. Using the property of the spectral norm $\|\cdot \|_2$ (which is the operator norm induced by the Euclidean norm $\|\cdot\|$), we have 
\[
\|x_i\| = \|E\mathrm{vec}(p_i)\| \leq \|E\|_2 \|\mathrm{vec}(p_i)\|,
\]
where $\mathrm{vec}(\cdot)$ denotes the operator that transforms a matrix into a column vector. By definition, we have 
\[
\|\mathrm{vec}(p_i)\|^2 = \|p_i\|_\mathrm{F}^2 = \sum_{u, v \in \Omega_i} |I(u,v)|^2 dudv = \mathcal{E}_i.
\]
As such, the Frobenius norm of the sequence of tokens $x$ verifies
\begin{equation}
\label{eq:bound_frobenius}
\|x\|_\mathrm{F} = \sqrt{\sum_{i=1}^n \|x_i\|^2} \leq \sqrt{\sum_{i=1}^n \|E\|_2 \|^2\mathrm{vec}(p_i)\|^2} = \|E\|_2 \sqrt{\sum_{i=1}^n \|\mathrm{vec}(p_i)\|^2} = \|E\|_2 \sqrt{\sum_{i=1}^n \mathcal{E}_i} = \|E\|_2 \cdot \sqrt{\mathcal{E}},
\end{equation}
as intended. In the rest of the proof, we denote $R = \alpha \cdot \sqrt{\mathcal{E}}$, with $\alpha$ the spectral norm of $E$. This implies that the sequences of tokens are in $B_R^n$, which corresponds to the setting of~\cref{prop:mha_bounds}. Indeed, each token verifies $\|x_i\| \leq R$; otherwise, the Frobenius norm would be greater than $R$. We now proceed to bound $\|f(x) - f(y)\|_\mathrm{F}$. For any $h \in [H]$, we introduce for convenience the function $S^h \colon \RR^{d \times n} \to \RR^{n \times n}$ as
\[
S^h(x) = \mathrm{softmax}\mleft(\frac{\mleft(Qx\mright)^\top Kx}{\sqrt{k}}\mright) = \mathrm{softmax}\mleft(x^\top A^h x\mright).
\]
Similarly to the proof of~\cref{prop:mha_bounds}, we have used the triangular inequality and~\cref{lem:frob_spectral} that
\begin{equation}
\label{eq:part_1}
\begin{split}
\|f(x) - f(y)\|_\mathrm{F} &= \| \sum_{h=1}^H O^h \mleft(f^h_\mathrm{att}(x) - f^h_\mathrm{att}(x) \mright)\|_\mathrm{F} \\
& \leq \sum_{h=1}^H \|O^h \mleft(f^h_\mathrm{att}(x) - f^h_\mathrm{att}(x) \mright)\|_\mathrm{F} \\
&\leq \sum_{h=1}^H \|O^h\|_2 \| f^h_\mathrm{att}(x) - f^h_\mathrm{att}(x)\|_\mathrm{F}.
\end{split}
\end{equation}
Moreover, by the definition of the self-attention layer, we have 
\begin{equation}
\label{eq:part_2}
\|f^h_\mathrm{att}(x) - f^h_\mathrm{att}(y)\|_\mathrm{F} = \|\mleft(V^hx\mright) S^h(x) - \mleft(V^hy\mright) S^h(y)\|_\mathrm{F} \leq \|V^h\|_2 \|x S^h(x) - yS^h(y)\|_\mathrm{F},
\end{equation}
where we used~\cref{lem:frob_spectral} for the inequality. Moreover, we have 
\begin{align*}
\|x S^h(x) - yS^h(y)\|_\mathrm{F} &= \| x\mleft(S^h(x) - S^h(y) \mright) + (x-y)S^h(y)\|_\mathrm{F} \\
&\leq \| x\mleft(S^h(x) - S^h(y) \mright)\|_\mathrm{F} + \| (x-y)S^h(y)\|_\mathrm{F} \tag{triangular inequality} \\
&\leq \|x\|_\mathrm{F} \| S^h(x) - S^h(y)\|_\mathrm{F} + \|x-y\|_\mathrm{F} \|S^h(y)\|_\mathrm{F},
\end{align*}
where we used~\cref{lem:frob_spectral} for the last inequality. We recall that we have $ \|x\|_\mathrm{F} \leq R$ from~\cref{eq:bound_frobenius}. Recalling the fact that $S$ is row-stochastic leads to $\|S^h(y)\|_\mathrm{F} \leq \sqrt{n}$ via the simple derivation
\[\|S^h(y)\|_\mathrm{F}^2 = \|S\|_\mathrm{F}^2 = \sum_{i=1}^n\sum_{j=1}^n S_{ij}^2 \leq \sum_{i=1}^n \underbrace{\sum_{j=1}^n S_{ij}}_{=1} \leq n,
\]
where we used $S = S^h(y)$ to alleviate the notations, this leads to
\begin{equation}
\label{eq:part_3}
\|x S^h(x) - yS^h(y)\|_\mathrm{F} \leq R \| S^h(x) - S^h(y)\|_\mathrm{F} + \sqrt{n} \|x-y\|_\mathrm{F}.
\end{equation}
We now proceed to bound the term $\| S^h(x) - S^h(y)\|_\mathrm{F}$. Since the softmax operator is applied row-wise, the $i$-th row of $S^h(x)$ writes $g(x)_i = \mathrm{softmax}\mleft( (x^\top A^h x)_i\mright) \in \RR^{1 \times n}$. We define $g(y)_i$ similarly. Then, we have
\[
\| S^h(x) - S^h(y)\|_\mathrm{F}^2 = \sum_{i=1}^n \sum_{j=1}^n (S^h(x) - S^h(y))_{ij}^2 =  \sum_{i=1}^n \|g(x)_i - g(y)_i\|^2.
\]
We recall the following result on the Lipschitz constant of the softmax operator with respect to the Euclidean norm, i.e., the $\ell_2$ norm (we note that \citet{nair2026softmax} states the result for any $\ell_p$ norm).
\begin{boxthm}{\citep[Theorem 1]{nair2026softmax}}
\label{thm:softmax}
Let $n \in \NN$ and $u, v\in \RR^n$. Then, 
$\| \mathrm{softmax}(u) - \mathrm{softmax}(v)\| \leq \frac{1}{2} \|u-v\|.$
\end{boxthm}  
Applying~\cref{thm:softmax} leads for any $i \in [n]$ to
\[
\|g(x)_i - g(y)_i\| = \|\mathrm{softmax}\mleft( (x^\top A^h x)_i\mright) - \mathrm{softmax}\mleft( (y^\top A^h y)_i\mright)\| \leq \frac{1}{2} \| (x^\top A^h x)_i - (y^\top A^h y)_i\|.
\]
Hence, we obtain using the fact that $\sqrt{\cdot}$ is monotonically increasing that
\begin{align*}
\| S^h(x) - S^h(y)\|_\mathrm{F} &\leq \mleft(\frac{1}{4} \sum_{i=1}^n \| (x^\top A^h x)_i - (y^\top A^h y)_i\| ^2\mright)^{1/2} \\
&= \frac{1}{2} \mleft(\sum_{i=1}^n \| (x^\top A^h x - y^\top A^h y)_i\|^2\mright)^{1/2} \\
&= \frac{1}{2} \| x^\top A^h x - y^\top A^h y\|_\mathrm{F}\\
&= \frac{1}{2} \| (x-y)^\top A^h x - y^\top A^h (x-y)\|_\mathrm{F} \\
&\leq \frac{1}{2} \mleft(\| (x-y)^\top A^h x \|_\mathrm{F} + \| y^\top A^h (x-y)\|_\mathrm{F}\mright) \\
&\leq \frac{1}{2} \mleft(\| (x-y)^\top \|_\mathrm{F} \|A^h x \|_\mathrm{F} + \| y^\top A^h \|_\mathrm{F} \|x-y\|_\mathrm{F}\mright) \tag{\cref{lem:frob_spectral}} \\
&\leq \frac{1}{2} \mleft(\| (x-y)^\top \|_\mathrm{F} \|A^h\|_2\|x \|_\mathrm{F} + \| y^\top \|_\mathrm{F} \|A^h \|_2 \|x-y\|_\mathrm{F}\mright) \tag{\cref{lem:frob_spectral}}.
\end{align*}
Since singular values are invariant to transposition and the Frobenius norm can be expressed as the sum of the singular values, we know that $\| (x-y)^\top \|_\mathrm{F} = \|x-y\|_\mathrm{F}$ and that $\| y^\top \|_\mathrm{F}  = \| y \|_\mathrm{F}$. Recalling that by assumption, the Frobenius norm of sequences of tokens is bounded by $R$ from~\cref{eq:bound_frobenius}, we have
\begin{align*}
\| S^h(x) - S^h(y)\|_\mathrm{F} &\leq \frac{1}{2} \mleft(\| x-y \|_\mathrm{F} \|A^h\|_2\|x \|_\mathrm{F} + \| y \|_\mathrm{F} \|A^h \|_\mathrm{F} \|x-y\|_\mathrm{F}\mright) \\
&\leq R \|A^h\|_2\|x-y\|_\mathrm{F}.
\end{align*}
From~\cref{eq:part_3}, we obtain 
\[
\|x S^h(x) - yS^h(y)\|_\mathrm{F} \leq \mleft(R^2 \|A^h\|_2 + \sqrt{n} \mright) \|x-y\|_\mathrm{F}, 
\]
and from~\cref{eq:part_2} we obtain
\[
\|f^h_\mathrm{att}(x) - f^h_\mathrm{att}(y)\|_\mathrm{F} \leq \|V^h\|_2\mleft(R^2  \|A^h\|_2 + \sqrt{n}\mright)\|x-y\|_\mathrm{F},
\]
This leads using~\cref{eq:part_1} to
\[
\|f(x) - f(y)\|_\mathrm{F} \leq \sum_{h=1}^H \|O^h\|_2 \|V^h\|_2\mleft(R^2  \|A^h\|_2 + \sqrt{n}\mright)\|x-y\|_\mathrm{F},
\]
with and $R = \alpha \sqrt{\mathcal{E}}$. As in~\cref{app:ln_bounds}, since the upper bound holds for randomly sampled sequences $x, y$, we can bound the rate of change and take the expectation to conclude the proof.
\end{proof}

\clearpage
\section{Reproducibility}
\label[secinapp]{app:reproducibility}
In this section, we provide details to reproduce our work.
\subsection{Open-source code}
To facilitate knowledge transfer and minimize redundant training runs within the community, our code and findings have been made publicly available at \href{https://github.com/ambroiseodt/vit-plasticity}{\texttt{github.com/ambroiseodt/vit-plasticity}}. Our library is built such that researchers can adapt all or part of the code for their specific use cases. We notably hope it can be used to further extend the study to large language models. 

\subsection{Carbon footprint}
\label[secinapp]{app:footprint}
This project required comprehensive, large-scale experiments with around $1000$ finetuning runs for an equivalent of $3700$ GPU hours. With public cloud providers such as Azure or Amazon Web Services, this can cost up to $\$40,000$. With a carbon efficiency of $0.1$ kgCO$_2$eq/kWh in the France region, the total emissions are estimated to be roughly 259 kgCO$_2$eq. This is equivalent to a round-trip flight from Paris to Madrid with a Boeing 737. We note that this number is low because France's electricity grid relies heavily on nuclear and renewable energy. For similar GPU-hours in regions using coal and gas, such as Germany, carbon emissions would be much higher. Estimations were conducted using the \href{https://mlco2.github.io/impact#compute}{ML Impact calculator} presented in~\citet{lacoste2019quantifying}. 

\section{Implementation details}
\label[secinapp]{app:implem}
In this section, we provide full implementation details.
\subsection{Vision transformers}
\label[secinapp]{app:vit}
This section is focused on the vision transformer implementation.
\paragraph{Architecture.} In vision transformers~\citep[ViT,][]{dosovitskiy2021vit}, inputs are $2$D images that are split into square patches of size $P$, which are flattened and linearly embedded in dimension $d$. A classification token \texttt{CLS} is prepended to the sequence of tokens before adding positional embeddings. The obtained sequence of tokens $x = (x_1, \ldots, x_n) \in (\RR^d)^n$ is fed to a succession of transformer encoders~\citep{vaswani2017attention}. Each block consists of a multihead self-attention module followed by a feedforward network implemented as a two-layer MLP with GeLU activation~\citep{hendrycks2016gelu} and a hidden dimension taken as $4$ times the embedding dimension~\citep{vaswani2017attention, dosovitskiy2021vit}. A LayerNorm~\citep{ba2016layernormalization} is applied before each block, and a residual connection is applied after each block. It leads to the $5$ modules displayed in~\cref{fig:overview} (left).

\paragraph{Implementation.} In our experiments, we use ViT models of size $86$M, $307$M, and $632$M  with patch sizes $16$, $16$, and $14$, respectively. Models are pretrained on ImageNet22k. Their characteristics are given in~\cref{tab:vit_variants}. In our code, we follow the original ViT implementation from~\citet{dosovitskiy2021vit} and use a convolutional layer to embed images~\citep[see][\S ``Hybrid Architecture"]{dosovitskiy2021vit}. This is also the standard in the implementation from~\citet{transformers}. In~\cref{fig:vit_implementation}, we display the implementation of the ViT-Base model with a classification head for $10$ classes obtained using our modular library \href{https://github.com/ambroiseodt/vit-plasticity}{\texttt{vitef}}.

\begin{figure}[!h]
\begin{center}
\begin{minipage}{0.95\linewidth}
\begin{lstlisting}[language=Python, frame=single]
# Python snippet to print the ViT architecture
from vitef.models import build_model

model = build_model(implementation="vit", model_name="base", n_classes=10)
print(model)

# Corresponding output
Transformer(
  (embedding): Embedding(
    (patching): PatchImages(
      (patching): Sequential(
        (0): Conv2d(3, 768, kernel_size=(16, 16), stride=(16, 16))
        (1): Flatten(start_dim=2, end_dim=-1)
      )
    )
  )
  (blocks): ModuleList(
    (0-11): 12 x TransformerBlock(
      (attn_norm): LayerNorm((768,), eps=1e-12, elementwise_affine=True)
      (attn): SelfAttention(
        (qkv_mat): Linear(in_features=768, out_features=2304, bias=True)
        (output): Linear(in_features=768, out_features=768, bias=True)
      )
      (ffn_norm): LayerNorm((768,), eps=1e-12, elementwise_affine=True)
      (ffn): FeedForward(
        (fc1): Linear(in_features=768, out_features=3072, bias=True)
        (fc2): Linear(in_features=3072, out_features=768, bias=True)
      )
    )
  )
  (output): Output(
    (output_layer): ClassificationLayer(
      (output_norm): LayerNorm((768,), eps=1e-12, elementwise_affine=True)
      (output): Linear(in_features=768, out_features=10, bias=True)
    )
  )
)
\end{lstlisting}
\end{minipage}
\end{center}
\caption{ViT-Base Implementation.}
\label{fig:vit_implementation}
\end{figure}

\begin{table}[!h]
    \centering
        \caption{Details of ViT variants~\citep{dosovitskiy2021vit} with the patch size, the number of layers, the number of attention heads, the embedding dimension, the number of parameters, and the link to the pretrained weights.}
        \begin{tabular}{lccccccc}
        model & patch size $P$ & seq. length $n$ & layers & heads $H$ & embedding $d$ & FFN hidden dimension &  parameters\\
        \toprule[\thick pt]
        ViT-Base & 16 & 197& 12 & 12 & 768 & 3072 & 86M \\
        ViT-Large & 16 & 197 & 24 & 16 & 1024 & 4096 & 307M\\
        ViT-Huge & 14 & 257 & 32 & 16 & 1280 & 5120& 632M\\
        \end{tabular}
    \label{tab:vit_variants}
\end{table}

\subsection{Data preprocessing}
\label[secinapp]{app:data}
All our experiments are conducted on a varied collection of $11$ classification benchmarks: Cifar10, Cifar100~\citep{krizhevsky2009learning}; variants from Cifar10-C~\citep{hendrycks2019robustness} with severity $5$: Contrast, Gaussian Noise, Motion Blur, Snow, Speckle Noise; $2$ domains from DomainNet~\citep{peng2019moment}, a challenging benchmark typically used for domain generalization: Clipart, Sketch; Flowers102~\citep{nilsback2008flowers102} and Pets~\citep{parkhi2012pets}.

The preprocessing follows~\citet{dosovitskiy2021vit} and~\citet{kolesnikov2020bit}: for training data, we apply random cropping, a 224$\times$224 image resizing, and random horizontal flip for training images. For validation and test data, the 224$\times$224 image resizing is applied before center cropping images. All images are normalized using the ImageNet1k~\citep{deng2009imagenet} statistics. It ensures images with mean $[0.485, 0.456, 0.406]$ and standard deviation $[0.229, 0.224, 0.225]$. For datasets that do not have predefined training and test sets (i.e., datasets from Cifar10-C and DomainNet), we manually create \emph{deterministic} training and test sets following a $80\%-20\%$ split. The deterministic part is crucial to ensure no data contamination.

\subsection{Plasticity setup}
\label[secinapp]{app:plasticity_implem}
This section is focused on the plasticity analysis.
\paragraph{Realistic setting.} In real-world applications, the discrepancy between the pretraining and downstream data is not known a priori. This motivates us to compute the plasticity images coming from the pretraining distribution and various downstream distributions, without any additional assumption. This differs from prior work, where the distribution shift can be categorized, e.g., into natural, subpopulation, or synthetic shift~\citep{lee2023surgical, xie2024mano, xie2025leveraging, deng2023nuclear}. 

\paragraph{Practical implementation.} The sequences of tokens $x, y$ are obtained by embedding preprocessed images with the pretrained model studied. We loop over $N$ batches of size $b$ with forward passes on the GPU and store high-dimensional outputs on the CPU. This ensures a fast computation and avoids out-of-memory issues. The total number of samples used to compute the plasticity is equal to $N \times b$. We note that all the transformer components take as input sequences of tokens in $\RR^d$, except for the second layer of the feedforward $f_\mathrm{fc2}$, where the tokens must be in $\RR^{4d}$. Akin to how a vector in the plane can be mapped to a 3D vector $(u_1, u_2, 0)$, we lift each token $x_i$ into $\RR^{4d}$ by padding the remaining entries with zeros.

\subsection{Finetuning setup}
\label[secinapp]{app:finetuning}
The finetuning experiments of~\cref{sec:finetuning} follow the protocol from~\citet{dosovitskiy2021vit} with a resolution of $224 \times 224$.

\paragraph{Configurations.} We consider pretrained models like ViT-Base and ViT-Large and finetune each of their trainable components in isolation: we freeze all the weights of each model, except the group studied, which is optimized across the depth: the attention norm (LN1), the attention module (MHA), the feedforward norm (LN2), the first feedforward layer (FC1), and the second feedforward layer (FC2). The classification head is randomly initialized following~\citet{dosovitskiy2021vit}. This leads to the $5$ configurations described in~\cref{tab:model_configurations} along with their corresponding number of trainable parameters.  We add as a baseline the full-finetuning (All), where all the model's parameters are trainable. 

\begin{table}[!h]
    \centering
        \caption{\textbf{Finetuning configurations}. Configurations are denoted by the name of the trainable transformer component and ordered in terms of plasticity ranking (see~\cref{sec:analysis}). We report the number of trainable parameters on ViT-Base.}
        \begin{tabular}{lccccc}
        configuration & MHA & FC1 & FC2 & LN2 & LN1 \\
        \toprule[\thick pt]
        parameters & 28M & 28M & 28M & 18K & 18K \\
        $\%$ of total & 33 & 33 & 33 & 0.02 & 0.02 \\
        \end{tabular}
    \label{tab:model_configurations}
\end{table}

\paragraph{Memory load.} The finetuning configurations have the same inference cost since they share the same ViT architecture. However, the number of trainable parameters differs. The GPU usage of training a model consists of the memory load to store the model parameters, the optimizer states, the gradients, and the activations~\citep{thor}. In our setting, the memory load is the same between configurations except for the optimizer and the gradient computation. For a model with $P$ parameters and a precision of $b$ bytes, the memory required to store the gradients is $Pb$ because backpropagation computes a gradient per parameter. The same memory is needed for the optimizer states with SGD (and the double for Adam~\citep{kingma2014adam, loshchilov2018decoupled}, which also computes the variance). In~\cref{tab:memory}, the memory usage for one training step on Cifar10 for each configuration with a default FP32 precision. 
\begin{table}[!h]
\centering
    \caption{\textbf{Memory load comparison.} Memory usage of the optimizer and gradients for one training step (in MB).}
    \label{tab:memory}
    \begin{tabular}{lccccc}
    configuration & MHA & FC1 & FC2 & LN2 & LN1 \\
    \toprule[\thick pt]
    \emph{memory load} & 220 & 220 & 220 & 0.14 & 0.14 \\
    \end{tabular}
\end{table}

\paragraph{Optimization.} We optimize models with the Stochastic Gradient Descent (SGD), a momentum of 0.9, no weight decay, a cosine learning rate decay, a warmup of $2000$ steps, a batch size of $512$, and gradient clipping at norm $1$. The finetuning resolution is of $224$. For each pair of dataset - configuration, we perform a sweep over $4$ learning rates, as summarized in~\cref{tab:training_details}, and conduct $3$ runs with different seeds relative to network initialization and dataloaders.

\begin{table}[!h]
    \centering
       \caption{\textbf{Finetuning hyperparameters}. We report the choice of optimizer, batch size, training steps, and learning rates.}
       \scalebox{1}{
        \begin{tabular}{lccccc}
        dataset & optimizer & batch size & training steps & learning rates $\eta$\\
        \toprule[\thick pt]
        Cifar10 & SGD & 512 & $10000$ & \{$1\mathrm{e}{-3}$, $3\mathrm{e}{-3}$, $1\mathrm{e}{-2}$, $3\mathrm{e}{-2}$\} \\
        Cifar100 & SGD & 512&  $10000$& \{$1\mathrm{e}{-3}$, $3\mathrm{e}{-3}$, $1\mathrm{e}{-2}$, $3\mathrm{e}{-2}$\} \\
        Contrast & SGD & 512&  $10000$ &\{$1\mathrm{e}{-3}$, $3\mathrm{e}{-3}$, $1\mathrm{e}{-2}$, $3\mathrm{e}{-2}$\} \\
        Gaussian Noise & SGD & 512&  $10000$ &\{$1\mathrm{e}{-3}$, $3\mathrm{e}{-3}$, $1\mathrm{e}{-2}$, $3\mathrm{e}{-2}$\} \\
        Motion Blur & SGD & 512&  $10000$ &\{$1\mathrm{e}{-3}$, $3\mathrm{e}{-3}$, $1\mathrm{e}{-2}$, $3\mathrm{e}{-2}$\} \\
        Snow & SGD & 512&  $10000$ &\{$1\mathrm{e}{-3}$, $3\mathrm{e}{-3}$, $1\mathrm{e}{-2}$, $3\mathrm{e}{-2}$\} \\
        Speckle Noise & SGD & 512&  $10000$ &\{$1\mathrm{e}{-3}$, $3\mathrm{e}{-3}$, $1\mathrm{e}{-2}$, $3\mathrm{e}{-2}$\} \\
        Clipart & SGD & 512&  $20000$ & \{$3\mathrm{e}{-3}$, $1\mathrm{e}{-2}$, $3\mathrm{e}{-2}$, $6\mathrm{e}{-2}$\}\\
        Sketch& SGD & 512&  $20000$ & \{$3\mathrm{e}{-3}$, $1\mathrm{e}{-2}$, $3\mathrm{e}{-2}$, $6\mathrm{e}{-2}$\} \\
        Flowers102 & SGD & 512&  $5000$ & \{$1\mathrm{e}{-3}$, $3\mathrm{e}{-3}$, $1\mathrm{e}{-2}$, $3\mathrm{e}{-2}$\} \\
        Pets & SGD & 512&  $4000$ & \{$1\mathrm{e}{-3}$, $3\mathrm{e}{-3}$, $1\mathrm{e}{-2}$, $3\mathrm{e}{-2}$\} \\
        \end{tabular}
        }
    \label{tab:training_details}
\end{table}

\paragraph{Performance.} For each run, we monitor the training using a validation set ($20\%$ of the training set). The final performance is the test accuracy of the checkpoint that achieves the best validation accuracy. 

\section{Additional experiments}
\label[secinapp]{app:exp}
In this section, we report the detailed results corresponding to the figures presented in the paper, along with additional experiments not shown in the main due to space constraints. For reproducibility, we provide the carbon footprint of our project.

\subsection{Plasticity analysis}
\label[secinapp]{app:analysis}
In this section, we present the additional figures and experiments related to~\cref{sec:theory,sec:analysis}.

\subsubsection{Theoretical plasticity ranking} 
\begin{figure}[!h]
    \centering
    \includegraphics[width=0.8\linewidth]{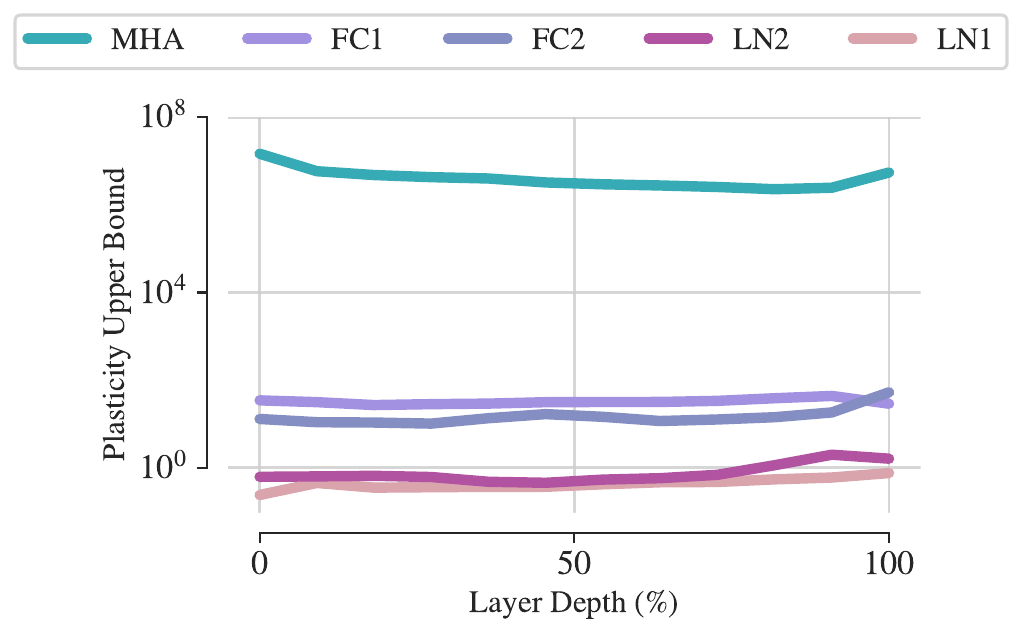}
    \caption{\textbf{Plasticity upper bounds on ViT-Base.} The sequence length is $n=197$, the number of heads is $H=12$ and the average radius is computed over input sequences $x = (x_1, \ldots, x_n)$ as $r = \sqrt{\frac{1}{n}\sum_{i=1}^n \|x_i\|^2}$. We obtain a value of $r=19.4$. We can see that the attention module has the highest plasticity, followed by the first and second feedforward layers, the LayerNorm preceding the feedforward, and finally the LayerNorm preceding the attention module. This aligns with our theoretical insights.}
    \label{fig:theory}
\end{figure}

\label[secinapp]{app:plasticity_theory}
We numerically compute the plasticity upper bounds of~\cref{sec:theory} on ViT-Base. The sequence length is $n=197$, and the number of attention heads is $H=12$. Following~\citet[Section 5]{castin2024smooth}, the average radius is computed over input sequences $x = (x_1, \ldots, x_n)$ as $r = \sqrt{\frac{1}{n}\sum_{i=1}^n \|x_i\|^2}$. The value $r=19.4$ obtained on Cifar10~\citep{krizhevsky2009learning} is used as the reference for the computation of the bounds in~\cref{prop:ln_bounds,prop:fc_bounds,prop:mha_bounds}. We display the upper bounds in~\cref{fig:theory}. The upper bounds ranking follows our theoretical insight,s with the attention module having the largest upper bound, followed by the first and second feedforward layer, the LayerNorm preceding the feedforward network, and finally, the LayerNorm preceding the attention module. We note that the upper bound of the attention module is several orders of magnitude larger than the other components. Even with the dependency in $n^{1/4}$ empirically observed in~\citet[Fig. 1]{castin2024smooth}, the order of the bound remains $10^6$. We attribute this scale to the dependency of the bound on the number of heads, the radius $r$, and the sequence length $n$. As explained in~\cref{sec:theory}, the bound is tight in terms of dependency in $n$, the numerical values of $r$ and $n$ being close leads to a large bound in practice. We notice in~\cref{sec:analysis} that the plasticity scales are more similar between modules than the upper bounds. This further confirms that the difference in scale between the upper bounds is due to the difficulty of bounding the self-attention Lipschitz constant. In particular, we observe in~\cref{sec:analysis} that the plasticity computed as an average rate of change follows the same ranking but with lower magnitude, notably for the attention module. This is reminiscent of~\citet{ashlagi2021smoothness} where the authors showed that the gap between the Lipschitz constant and the average rate of change can be considerable. 

\subsubsection{Plasticity experiments of all benchmarks} 
\begin{figure}[!h]
    \centering
    \includegraphics[width=\linewidth]{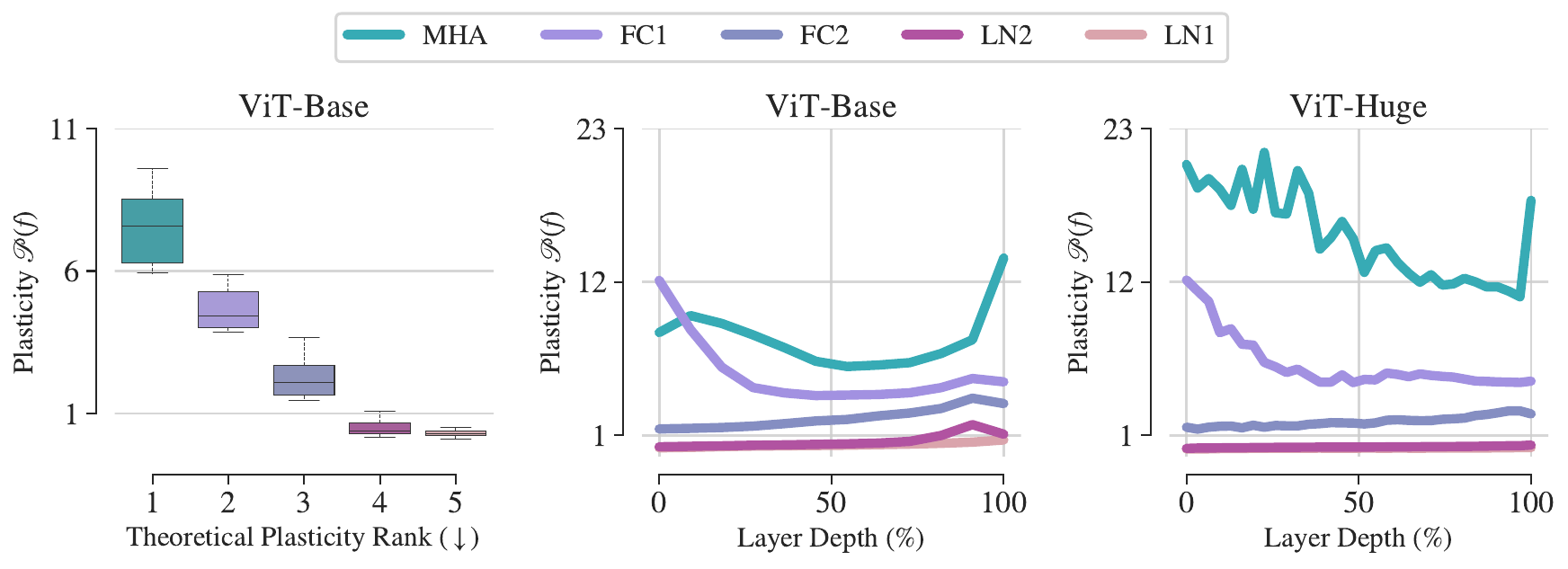}
    \caption{\textbf{Plasticity analysis on Cifar10.} The distribution of rates of change $\|f(x)-f(y)\|_\mathrm{F}/\|x-y\|_\mathrm{F}$ on ViT-Base (\textbf{left}) follow the upper bound ranking predicted by our theory in~\cref{sec:theory}. We observe along transformer blocks of ViT-Base (\textbf{middle}) that the attention module has the highest plasticity $\mathcal{P}(f)$, followed by the first and second linear layers of the feedforward. The LayerNorms are the most rigid, with a plasticity below $1$. The same pattern is obtained on ViT-Huge (\textbf{right}), where the higher attention plasticity further validates our theory (see \cref{prop:mha_bounds}) since the sequence length $n$ is larger than with ViT-Base.}
    \label{fig:analysis_cifar10}
\end{figure}

\begin{figure}[!h]
    \centering
    \includegraphics[width=\linewidth]{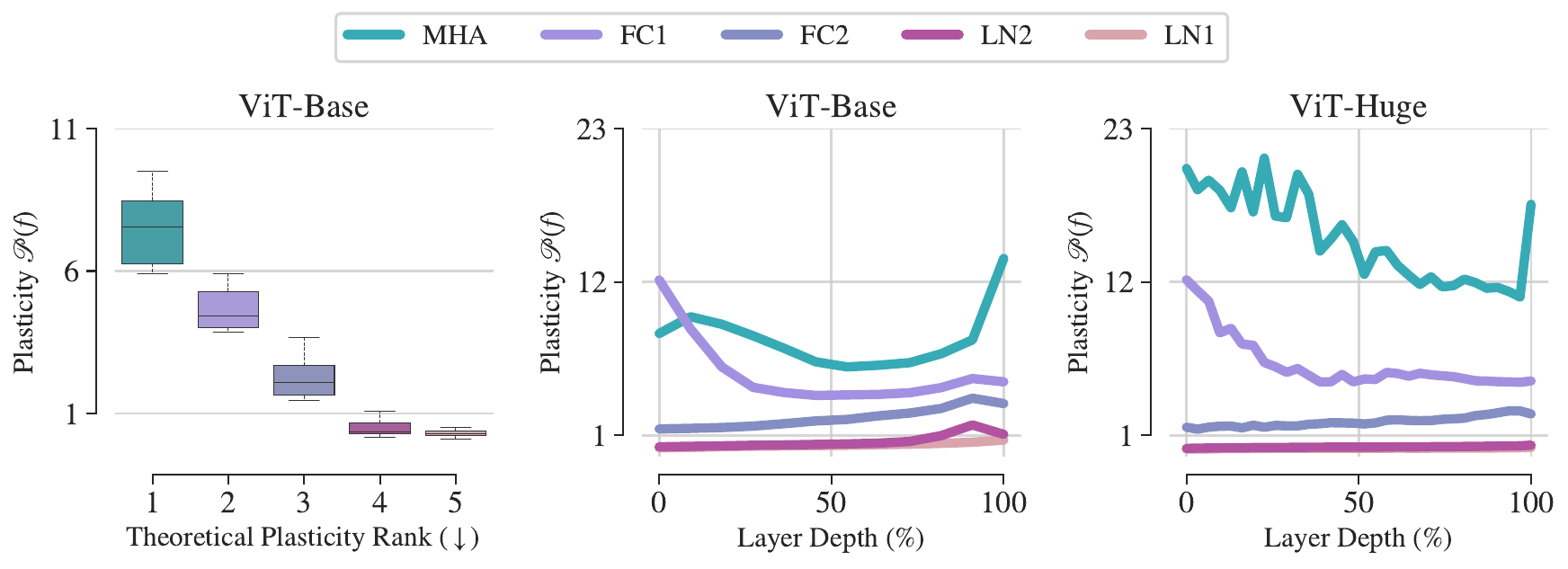}
    \caption{\textbf{Plasticity analysis on Cifar100.} The distribution of rates of change $\|f(x)-f(y)\|_\mathrm{F}/\|x-y\|_\mathrm{F}$ on ViT-Base (\textbf{left}) follow the upper bound ranking predicted by our theory in~\cref{sec:theory}. We observe along transformer blocks of ViT-Base (\textbf{middle}) that the attention module has the highest plasticity $\mathcal{P}(f)$, followed by the first and second linear layers of the feedforward. The LayerNorms are the most rigid, with a plasticity below $1$. The same pattern is obtained on ViT-Huge (\textbf{right}), where the higher attention plasticity further validates our theory (see \cref{prop:mha_bounds}) since the sequence length $n$ is larger than with ViT-Base.}
    \label{fig:analysis_cifar100}
\end{figure}

\begin{figure}[!h]
    \centering
    \includegraphics[width=\linewidth]{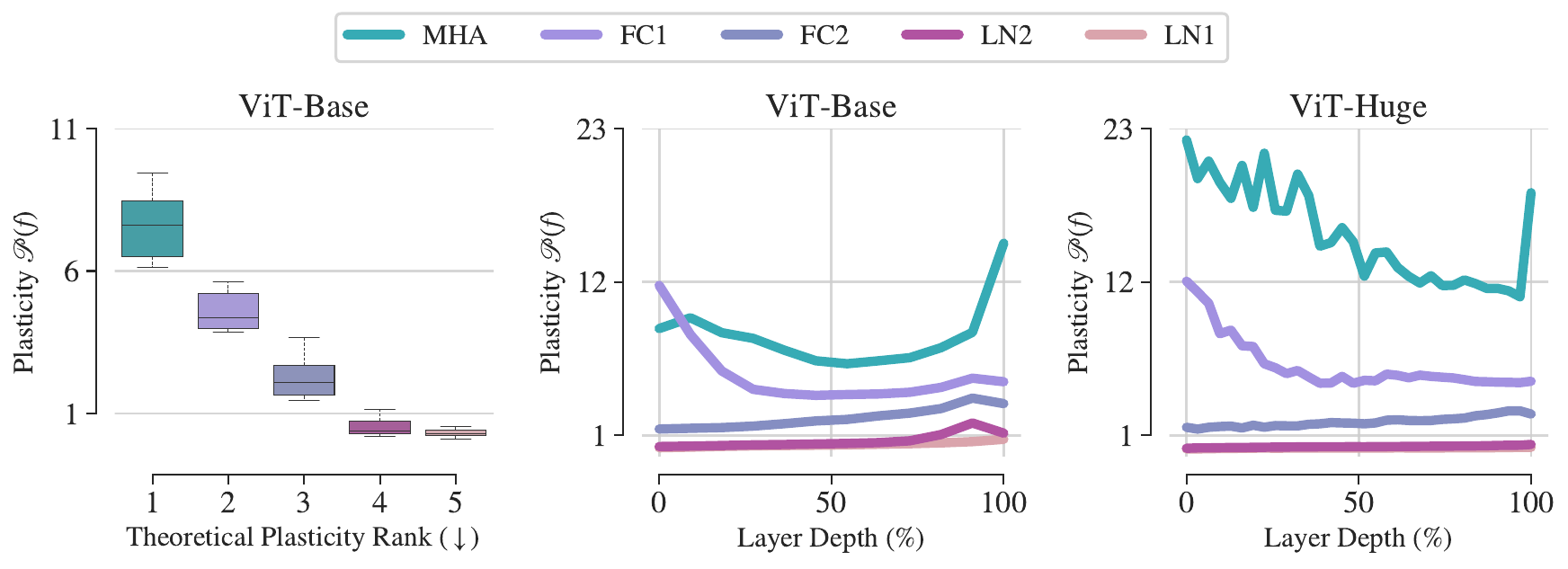}
    \caption{\textbf{Plasticity analysis on Contrast.} The distribution of rates of change $\|f(x)-f(y)\|_\mathrm{F}/\|x-y\|_\mathrm{F}$ on ViT-Base (\textbf{left}) follow the upper bound ranking predicted by our theory in~\cref{sec:theory}. We observe along transformer blocks of ViT-Base (\textbf{middle}) that the attention module has the highest plasticity $\mathcal{P}(f)$, followed by the first and second linear layers of the feedforward. The LayerNorms are the most rigid, with a plasticity below $1$. The same pattern is obtained on ViT-Huge (\textbf{right}), where the higher attention plasticity further validates our theory (see \cref{prop:mha_bounds}) since the sequence length $n$ is larger than with ViT-Base.}
    \label{fig:analysis_contrast}
\end{figure}

\begin{figure}[!h]
    \centering
    \includegraphics[width=\linewidth]{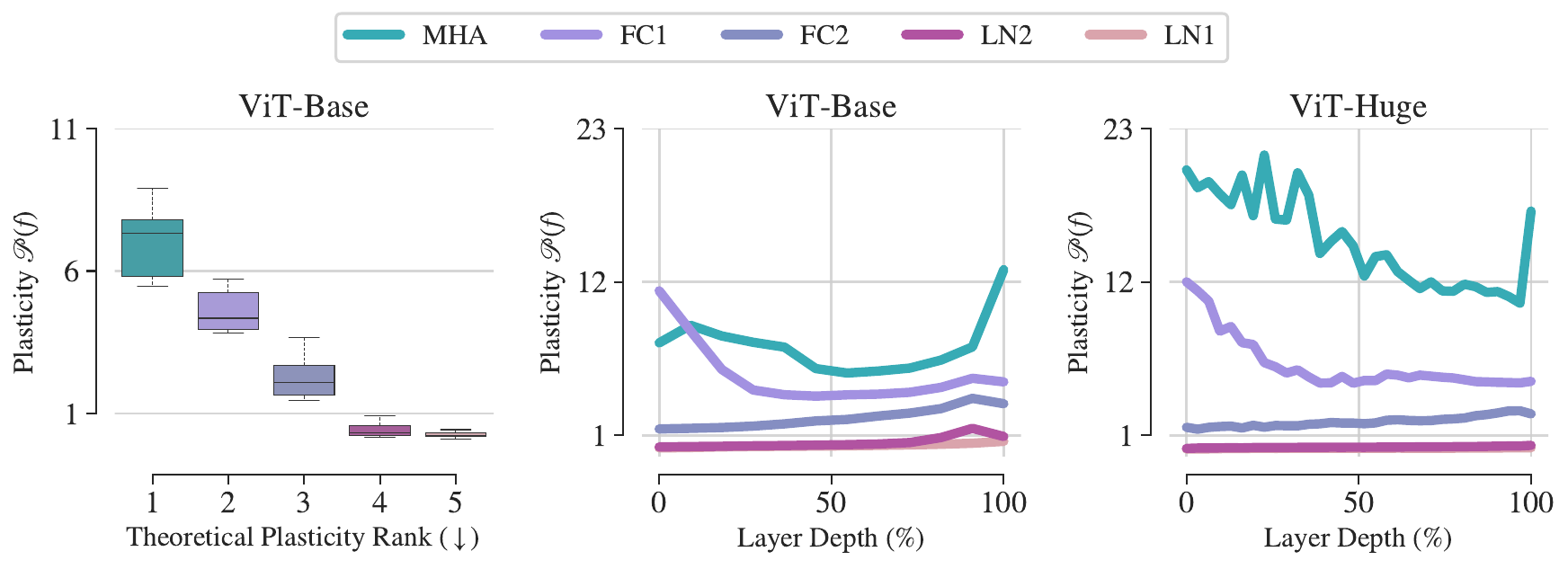}
    \caption{\textbf{Plasticity analysis on Gaussian Noise.} The distribution of rates of change $\|f(x)-f(y)\|_\mathrm{F}/\|x-y\|_\mathrm{F}$ on ViT-Base (\textbf{left}) follow the upper bound ranking predicted by our theory in~\cref{sec:theory}. We observe along transformer blocks of ViT-Base (\textbf{middle}) that the attention module has the highest plasticity $\mathcal{P}(f)$, followed by the first and second linear layers of the feedforward. The LayerNorms are the most rigid, with a plasticity below $1$. The same pattern is obtained on ViT-Huge (\textbf{right}), where the higher attention plasticity further validates our theory (see \cref{prop:mha_bounds}) since the sequence length $n$ is larger than with ViT-Base.}
    \label{fig:analysis_gaussian_noise}
\end{figure}

\begin{figure}[!h]
    \centering
    \includegraphics[width=\linewidth]{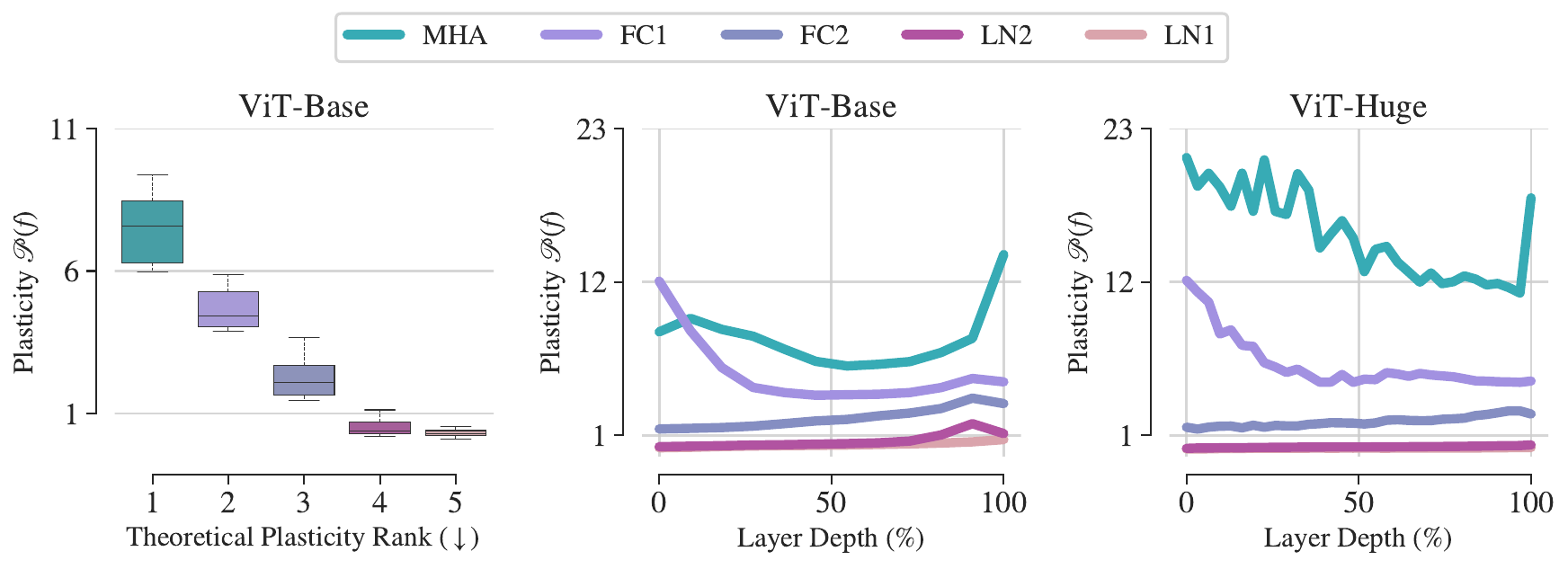}
    \caption{\textbf{Plasticity analysis on Motion Blur.} The distribution of rates of change $\|f(x)-f(y)\|_\mathrm{F}/\|x-y\|_\mathrm{F}$ on ViT-Base (\textbf{left}) follow the upper bound ranking predicted by our theory in~\cref{sec:theory}. We observe along transformer blocks of ViT-Base (\textbf{middle}) that the attention module has the highest plasticity $\mathcal{P}(f)$, followed by the first and second linear layers of the feedforward. The LayerNorms are the most rigid, with a plasticity below $1$. The same pattern is obtained on ViT-Huge (\textbf{right}), where the higher attention plasticity further validates our theory (see \cref{prop:mha_bounds}) since the sequence length $n$ is larger than with ViT-Base.}
    \label{fig:analysis_motion_blur}
\end{figure}

\begin{figure}[!h]
    \centering
    \includegraphics[width=\linewidth]{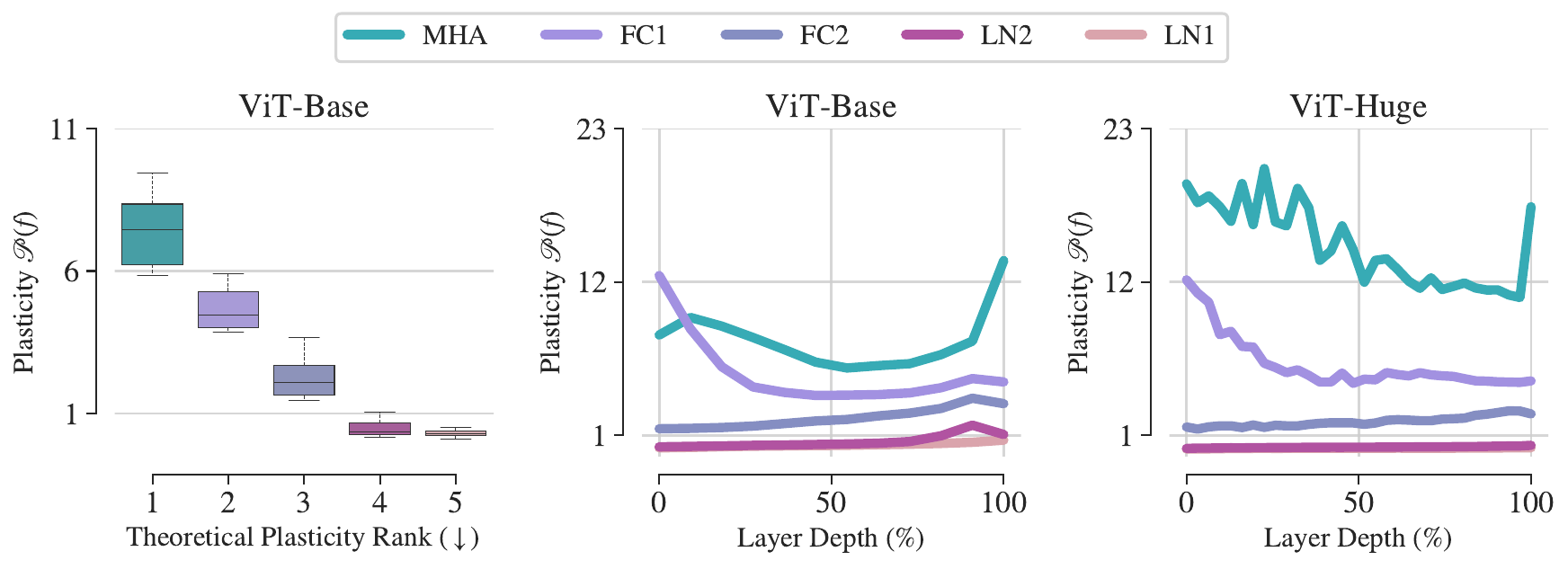}
    \caption{\textbf{Plasticity analysis on Snow.} The distribution of rates of change $\|f(x)-f(y)\|_\mathrm{F}/\|x-y\|_\mathrm{F}$ on ViT-Base (\textbf{left}) follow the upper bound ranking predicted by our theory in~\cref{sec:theory}. We observe along transformer blocks of ViT-Base (\textbf{middle}) that the attention module has the highest plasticity $\mathcal{P}(f)$, followed by the first and second linear layers of the feedforward. The LayerNorms are the most rigid, with a plasticity below $1$. The same pattern is obtained on ViT-Huge (\textbf{right}), where the higher attention plasticity further validates our theory (see \cref{prop:mha_bounds}) since the sequence length $n$ is larger than with ViT-Base.}
    \label{fig:analysis_snow}
\end{figure}

\begin{figure}[!h]
    \centering
    \includegraphics[width=\linewidth]{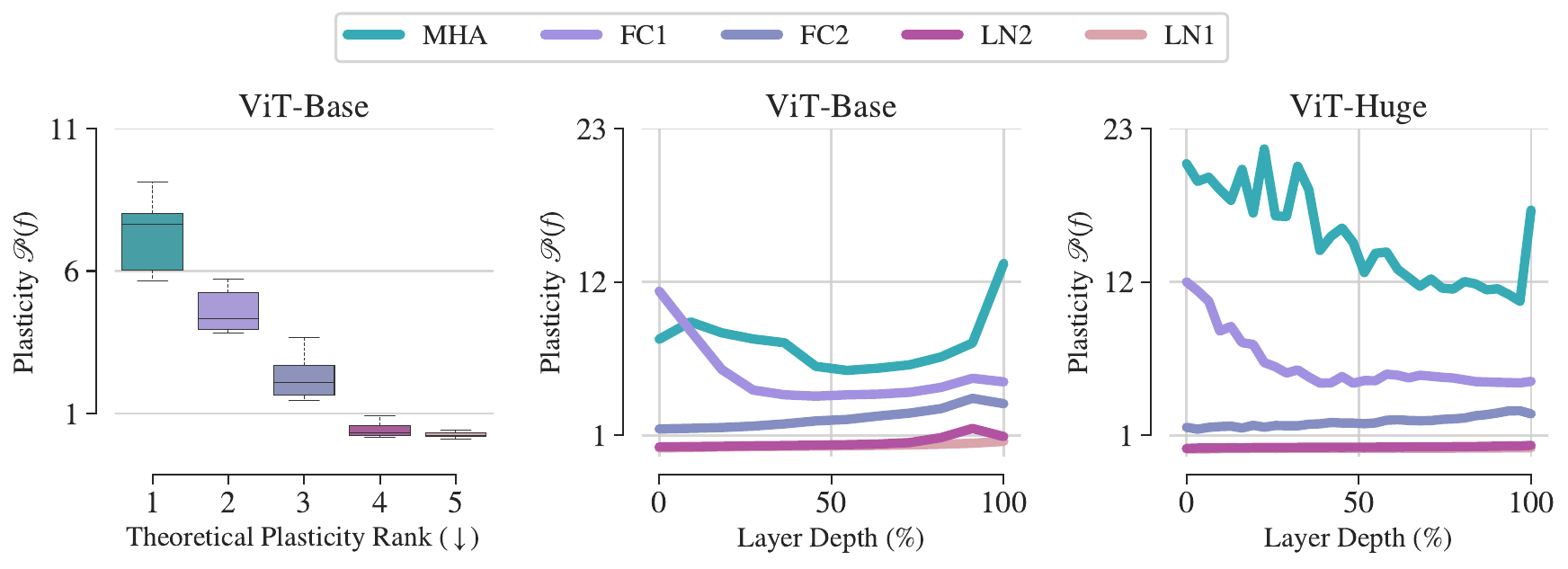}
    \caption{\textbf{Plasticity analysis on Speckle Noise.} The distribution of rates of change $\|f(x)-f(y)\|_\mathrm{F}/\|x-y\|_\mathrm{F}$ on ViT-Base (\textbf{left}) follow the upper bound ranking predicted by our theory in~\cref{sec:theory}. We observe along transformer blocks of ViT-Base (\textbf{middle}) that the attention module has the highest plasticity $\mathcal{P}(f)$, followed by the first and second linear layers of the feedforward. The LayerNorms are the most rigid, with a plasticity below $1$. The same pattern is obtained on ViT-Huge (\textbf{right}), where the higher attention plasticity further validates our theory (see \cref{prop:mha_bounds}) since the sequence length $n$ is larger than with ViT-Base.}
    \label{fig:analysis_speckle_noise}
\end{figure}

\begin{figure}[!h]
    \centering
    \includegraphics[width=\linewidth]{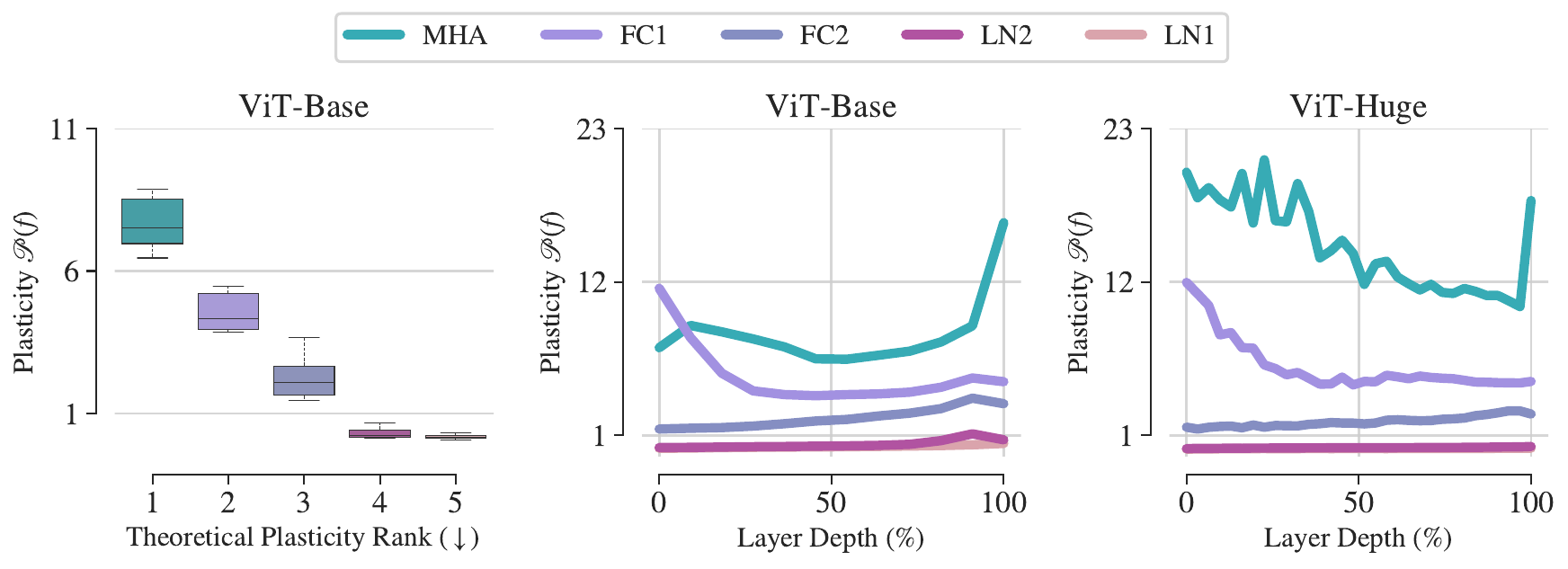}
    \caption{\textbf{Plasticity analysis on Clipart.} The distribution of rates of change $\|f(x)-f(y)\|_\mathrm{F}/\|x-y\|_\mathrm{F}$ on ViT-Base (\textbf{left}) follow the upper bound ranking predicted by our theory in~\cref{sec:theory}. We observe along transformer blocks of ViT-Base (\textbf{middle}) that the attention module has the highest plasticity $\mathcal{P}(f)$, followed by the first and second linear layers of the feedforward. The LayerNorms are the most rigid, with a plasticity below $1$. The same pattern is obtained on ViT-Huge (\textbf{right}), where the higher attention plasticity further validates our theory (see \cref{prop:mha_bounds}) since the sequence length $n$ is larger than with ViT-Base.}
    \label{fig:analysis_clipart}
\end{figure}

\begin{figure}[!h]
    \centering
    \includegraphics[width=\linewidth]{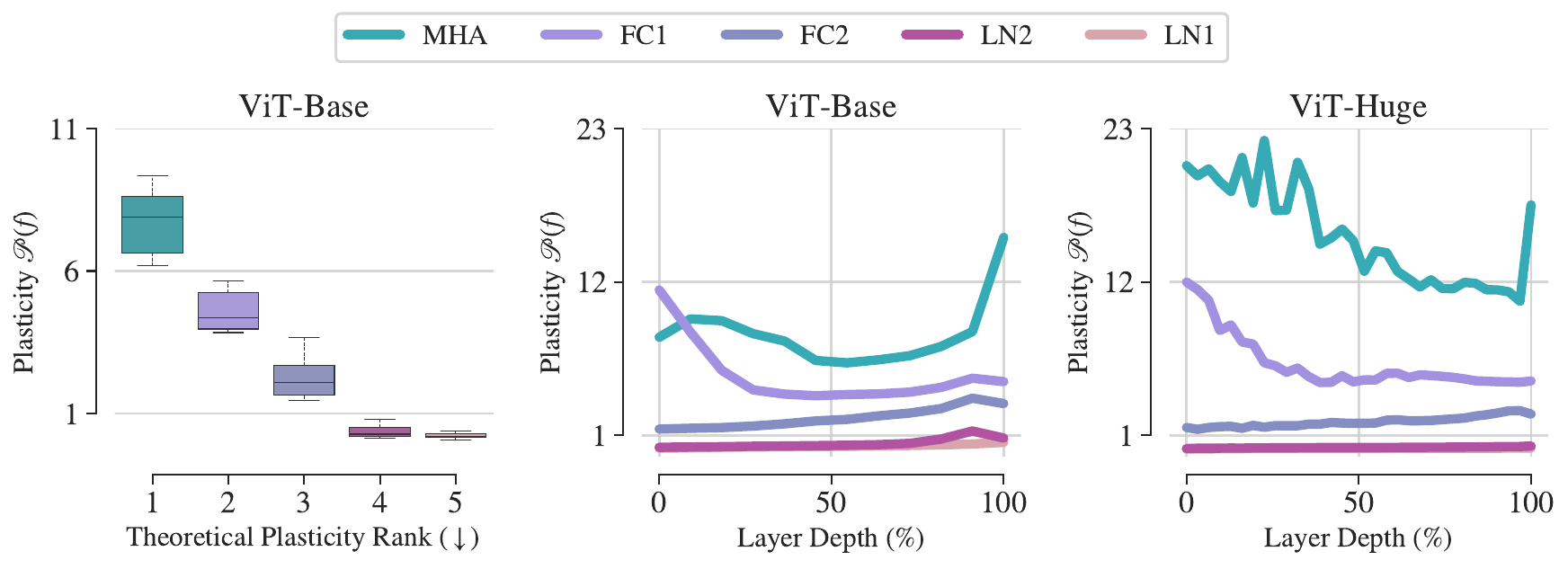}
    \caption{\textbf{Plasticity analysis on Flowers102.} The distribution of rates of change $\|f(x)-f(y)\|_\mathrm{F}/\|x-y\|_\mathrm{F}$ on ViT-Base (\textbf{left}) follow the upper bound ranking predicted by our theory in~\cref{sec:theory}. We observe along transformer blocks of ViT-Base (\textbf{middle}) that the attention module has the highest plasticity $\mathcal{P}(f)$, followed by the first and second linear layers of the feedforward. The LayerNorms are the most rigid, with a plasticity below $1$. The same pattern is obtained on ViT-Huge (\textbf{right}), where the higher attention plasticity further validates our theory (see \cref{prop:mha_bounds}) since the sequence length $n$ is larger than with ViT-Base.}
    \label{fig:analysis_flowers102}
\end{figure}

\begin{figure}[!h]
    \centering
    \includegraphics[width=\linewidth]{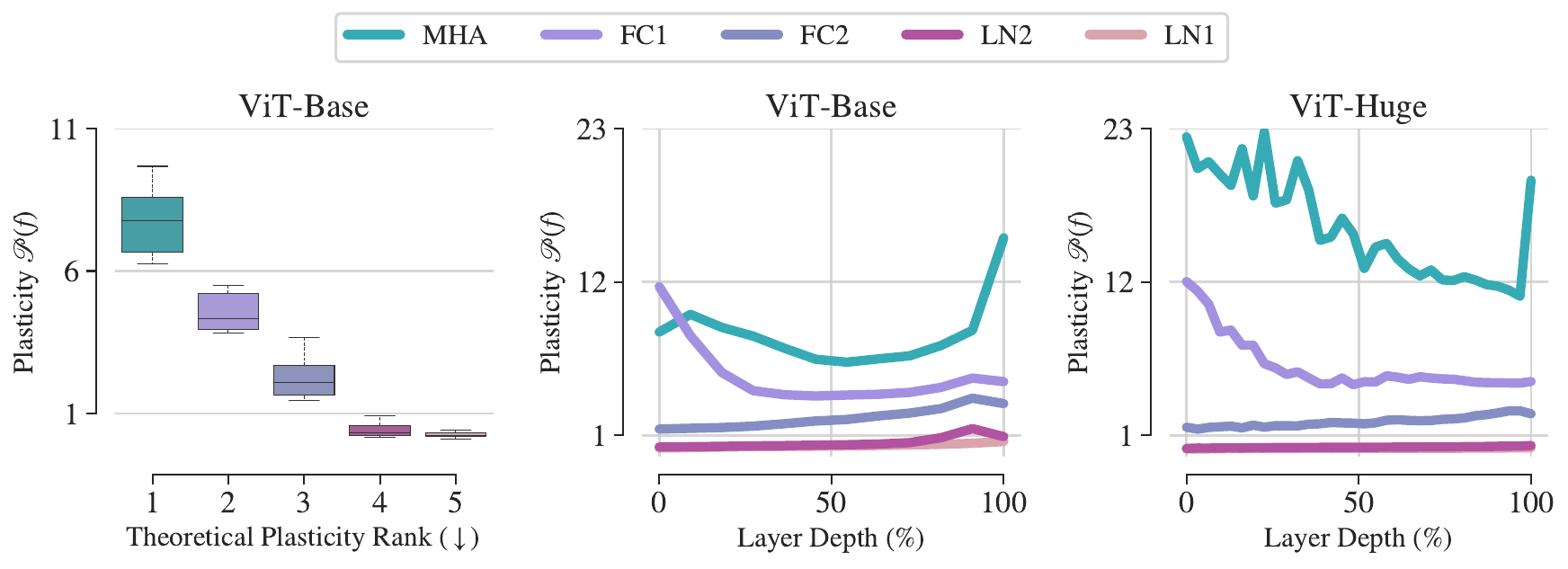}
    \caption{\textbf{Plasticity analysis on Pet.} The distribution of rates of change $\|f(x)-f(y)\|_\mathrm{F}/\|x-y\|_\mathrm{F}$ on ViT-Base (\textbf{left}) follow the upper bound ranking predicted by our theory in~\cref{sec:theory}. We observe along transformer blocks of ViT-Base (\textbf{middle}) that the attention module has the highest plasticity $\mathcal{P}(f)$, followed by the first and second linear layers of the feedforward. The LayerNorms are the most rigid, with a plasticity below $1$. The same pattern is obtained on ViT-Huge (\textbf{right}), where the higher attention plasticity further validates our theory (see \cref{prop:mha_bounds}) since the sequence length $n$ is larger than with ViT-Base.}
    \label{fig:analysis_pet}
\end{figure}

\label[secinapp]{app:plasticity_exp}
We extend the analysis of~\cref{sec:analysis} to additional datasets and display the results in~\cref{fig:analysis_cifar10,fig:analysis_cifar100,fig:analysis_contrast,fig:analysis_gaussian_noise,fig:analysis_motion_blur,fig:analysis_snow,fig:analysis_speckle_noise,fig:analysis_clipart,fig:analysis_flowers102,fig:analysis_pet}. Our findings are aligned with the theoretical analysis in~\cref{sec:theory} and shows that the attention module has the highest plasticity, followed by the first feedforward linear layer, then the second feedforward linear layer. The LayerNorms are more rigid with a plasticity below $1$.

\clearpage
\subsubsection{Plasticity experiments on DINOv3 and GPT2}
\begin{figure}[!h]
    \centering
    \includegraphics[width=\linewidth]{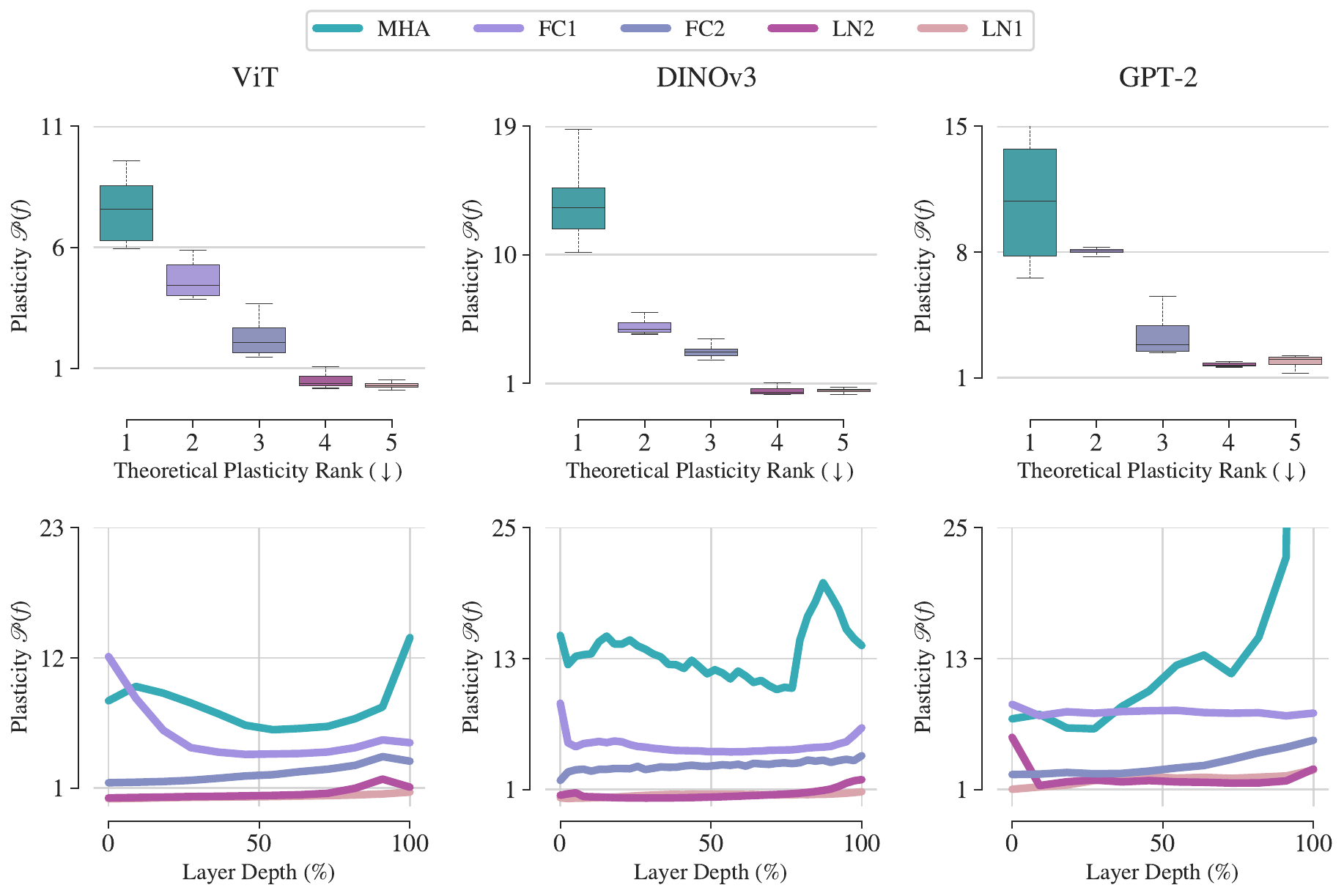}
    \caption{\textbf{Plasticity analysis on ViT-Base, DINOv3, and GPT2.} The distribution of rates of change $\|f(x)-f(y)\|_\mathrm{F}/\|x-y\|_\mathrm{F}$ on ViT-Base (\textbf{top left}), DINOv3-7B (\textbf{top middle}), and GPT2 (\textbf{top right}) follows the theoretical ranking of~\cref{sec:theory}. We observe along transformer blocks of ViT-Base (\textbf{top left}), DINOv3-7B (\textbf{bottom middle}), and GPT2 (\textbf{bottom right}) that the attention module has the highest plasticity $\mathscr{P}(f)$, followed by the first and second linear layers of the feedforward. The LayerNorms are the most rigid, with a plasticity below $1$.}
    \label{fig:plasticity_ablation_all}
\end{figure}

\label[secinapp]{app:plasticity_dinov3_gpt2}
We extend our analysis to DINOv3~\citep{simeoni2025dinov3}, a $7$B-parameter vision transformer trained in a self-supervised fashion, and GPT2~\citep{radford2019gpt2}, a $124$M-parameter decoder-only language model. Since ImageNet22k is part of the pretrained set of DINOv3, we follow the same experimental setup as with pretrained ViTs and report the results on Cifar10. To extend the analysis to GPT2, we use the fact that the pretraining set WebText~\citep{radford2019gpt2} was built by scraping the web and includes all high-quality outbound links from Reddit (at least $3$ karma). Since it contains mainstream news articles, political commentary, and tech journalism, this motivates us to use AGNews~\citep{zhang2015agnews}, a dataset of news articles about the world, sports, business, and science, as a representative proxy for the pre-training data distribution. Moreover,
the authors explicitly ``\textit{removed all Wikipedia
documents}~\citep{radford2019gpt2} from WebText. This allows us to consider the WikiText-103~\citep{merity2017pointer} as the downstream dataset. We display the distribution and evolution of plasticity across layers of DINOv3 and GPT2 in~\cref{fig:plasticity_ablation_all}. The observed patterns are consistent with those of supervised ViTs shown in~\cref{fig:analysis_sketch}. 

\clearpage
\subsection{Finetuning analysis}
\label[secinapp]{app:finetuning_exp}
In this section, we provide the additional results, figures, and experiments related to~\cref{sec:finetuning}.

\subsubsection{Performance comparison}
\begin{figure}[!h]
    \centering
    \includegraphics[width=.8\linewidth]{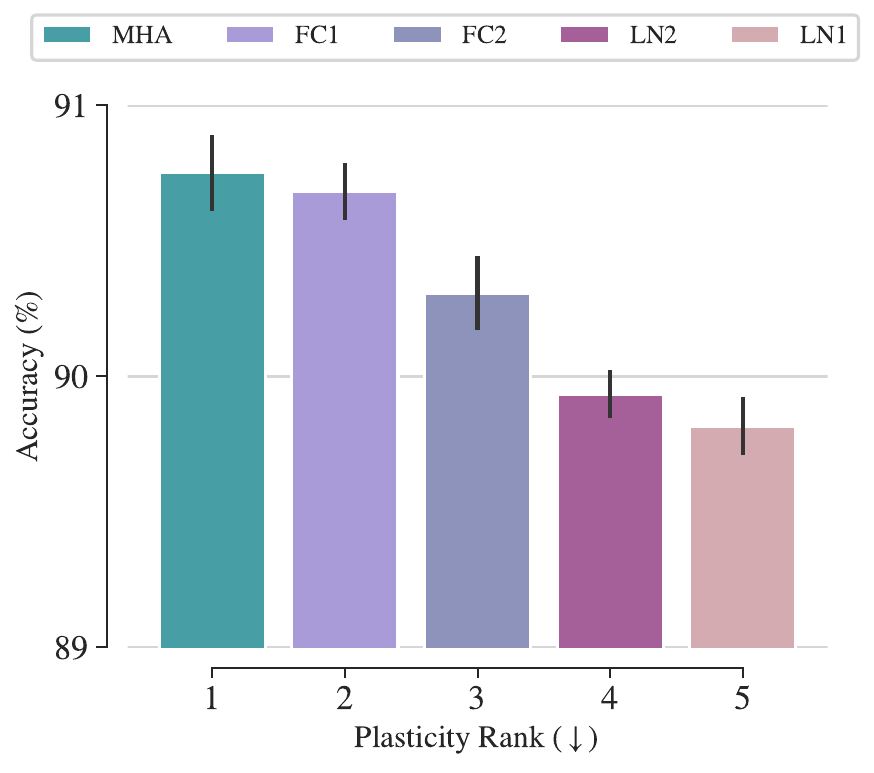}
    \caption{\textbf{Better performance (11 benchmarks).} We compare transformer components, ordered in terms of decreasing plasticity, and report the average top-$1$ accuracy over a diverse set of $11$ benchmarks, with the pooled standard error computed over $3$ finetuning runs. We can see that high plasticity results in better performance.}
    \label{fig:finetuning_all}
\end{figure}

\label[secinapp]{app:finetuning_exp}
In~\cref{tab:results}, we gather the finetuning performance for each configuration and dataset. It is used to obtain~\cref{fig:overview} (right). The relative gain metric is computed as the percentage improvement between the finetuning and the linear probing performance. 
\begin{table}[!h]
    \centering
    \caption{\textbf{Full finetuning results.}  We investigate the benefits of plasticity by evaluating finetuning the trainable components of~\cref{tab:model_configurations} on a diverse set of $11$ image classification benchmarks. We report the best top-$1$ accuracy (\%) on the test set over the learning rate grid of each dataset ($\uparrow$). Entries show the mean and standard deviation over three finetuning runs with different seeds. Transformer components are ordered in terms of decreasing plasticity. The best overall performance among the transformer components configurations of~\cref{tab:model_configurations} is in \textbf{bold}. The \emph{full-finetuning} configuration lets the $86$M parameters of ViT-Base be trainable. The \emph{linear probing} performance is obtained over the hidden representation of the last layer following~\citet{caron2021emerging}.}
    \scalebox{1}{
    \begin{NiceTabular}{lccccccc}
    \CodeBefore
        \rowcolor{tablegray}{13} 
    \Body
     configuration & MHA & FC1 & FC2 & LN2 & LN1 & \emph{full finetuning} & \emph{linear probing} \\
     \toprule[\thick pt] 
     Cifar10 & $\text{98.91}_{\tiny \pm \text{0.07}}$ & $\text{99.09}_{\tiny \pm \text{0.05}}$ & $\text{98.91}_{\tiny \pm \text{0.06}}$ & $\text{98.72}_{\tiny \pm \text{0.05}}$ & $\text{98.67}_{\tiny \pm \text{0.03}}$ & $\text{99.02}_{\tiny \pm \text{0.02}}$ & 92.07 \\
     Cifar100 & $\text{92.65}_{\tiny \pm \text{0.07}}$ & $\text{92.85}_{\tiny \pm \text{0.07}}$ & $\text{92.31}_{\tiny \pm \text{0.11}}$ & $\text{91.93}_{\tiny \pm \text{0.11}}$ & $\text{91.43}_{\tiny \pm \text{0.07}}$ & $\text{92.74}_{\tiny \pm \text{0.05}}$ & 67.56\\
     Contrast & $\text{97.09}_{\tiny \pm \text{0.11}}$ & $\text{97.06}_{\tiny \pm \text{0.08}}$ & $\text{96.28}_{\tiny \pm \text{0.11}}$ & $\text{96.67}_{\tiny \pm \text{0.20}}$ & $\text{96.89}_{\tiny \pm \text{0.19}}$ & $\text{97.23}_{\tiny \pm \text{0.18}}$ & 75.55\\
     Gaussian Noise & $\text{89.41}_{\tiny \pm \text{0.53}}$ & $\text{89.49}_{\tiny \pm \text{0.16}}$ & $\text{88.49}_{\tiny \pm \text{0.51}}$ & $\text{89.55}_{\tiny \pm \text{0.04}}$ & $\text{88.99}_{\tiny \pm \text{0.24}}$ & $\text{87.14}_{\tiny \pm \text{1.16}}$ & 50.35\\
     Motion Blur & $\text{94.72}_{\tiny \pm \text{0.21}}$ & $\text{94.53}_{\tiny \pm \text{0.06}}$ & $\text{94.04}_{\tiny \pm \text{0.16}}$ & $\text{93.95}_{\tiny \pm \text{0.34}}$ & $\text{93.25}_{\tiny \pm \text{0.29}}$ & $\text{94.67}_{\tiny \pm \text{0.14}}$ & 62.75\\ 
     Snow & $\text{95.47}_{\tiny \pm \text{0.13}}$ & $\text{95.52}_{\tiny \pm \text{0.20}}$ & $\text{95.27}_{\tiny \pm \text{0.29}}$ & $\text{95.51}_{\tiny \pm \text{0.11}}$ & $\text{95.15}_{\tiny \pm \text{0.10}}$ & $\text{95.42}_{\tiny \pm \text{0.13}}$ & 61.85\\ 
     Speckle Noise & $\text{90.07}_{\tiny \pm \text{0.32}}$ & $\text{89.85}_{\tiny \pm \text{0.34}}$ & $\text{89.22}_{\tiny \pm \text{0.31}}$ & $\text{89.71}_{\tiny \pm \text{0.17}}$ & $\text{89.74}_{\tiny \pm \text{0.31}}$ & $\text{89.58}_{\tiny \pm \text{0.43}}$ & 52.10\\
     Clipart & 
    $\text{77.31}_{\tiny \pm \text{0.41}}$ & $\text{76.47}_{\tiny \pm \text{0.24}}$ & $\text{76.54}_{\tiny \pm \text{0.17}}$ & $\text{74.37}_{\tiny \pm \text{0.08}}$ & $\text{74.65}_{\tiny \pm \text{0.16}}$ & $\text{78.50}_{\tiny \pm \text{0.49}}$ & 45.86\\
     Sketch & $\text{69.23}_{\tiny \pm \text{0.05}}$ & $\text{69.31}_{\tiny \pm \text{0.18}}$ & $\text{69.49}_{\tiny \pm \text{0.20}}$ & $\text{65.27}_{\tiny \pm \text{0.15}}$ & $\text{65.76}_{\tiny \pm \text{0.10}}$ & $\text{71.30}_{\tiny \pm \text{0.26}}$ & 30.16\\
     Flowers102 & 
    $\text{99.03}_{\tiny \pm \text{0.08}}$ & $\text{99.05}_{\tiny \pm \text{0.06}}$ & $\text{98.86}_{\tiny \pm \text{0.06}}$ & $\text{99.21}_{\tiny \pm \text{0.07}}$ & $\text{98.99}_{\tiny \pm \text{0.20}}$ & $\text{99.15}_{\tiny \pm \text{0.05}}$ & 96.62\\
     Pet & $\text{94.37}_{\tiny \pm \text{0.13}}$ & $\text{94.26}_{\tiny \pm \text{0.26}}$ & $\text{93.98}_{\tiny \pm \text{0.20}}$ & $\text{94.39}_{\tiny \pm \text{0.13}}$ & $\text{94.46}_{\tiny \pm \text{0.11}}$ & $\text{94.57}_{\tiny \pm \text{0.29}}$ & 89.18\\
     \midrule[\midthick pt]
     Avg. & \textbf{90.75} & 90.68 & 90.31 & 89.93 & 89.82 & 90.90 & 64.22 \\
    \end{NiceTabular}
    }
    \label{tab:results}
\end{table}
For visualization purposes, we display in~\cref{fig:finetuning_all} the overall performance of each configuration on the $11$ benchmarks. We observe similar patterns than with the relative gain in~\cref{fig:overview} (right): the higher the plasticity, the better the finetuning. 
\subsubsection{Extension to ViT-Large ($307$M)}
\label[secinapp]{app:vit_large}
We extend the finetuning analysis to ViT-Large ($307$M parameters) and obtain similar conclusions. Results on Clipart are displayed in~\cref{tab:base_vs_large}.
\begin{table}[!h]
\centering
    \caption{\textbf{Consistent benefits across model sizes.} We report the average top-$1$ accuracy (\%) on the test set over the learning rate grid of Clipart ($\uparrow$). Entries show the mean and standard deviation over learning rates. The best overall performance among the transformer components configurations of~\cref{tab:model_configurations} is in \textbf{bold} and \colorbox{tablegray}{non-smooth components} are highlighted in gray.}
    \label{tab:base_vs_large}
    \scalebox{1}{
    \begin{NiceTabular}{lccccc}
    \CodeBefore
    \columncolor{tablegray}{2-4} 
    \rowcolor{white}{1}
    \Body
    configuration & MHA & FC1 & FC2 & LN2 & LN1 \\
    \toprule[\thick pt]
    ViT-Base ($86$M) & \textbf{76.9} $_{\tiny \pm \text{0.6}}$  & 75.7 $ _{\tiny \pm \text{0.6}}$ &  75.9 $ _{\tiny \pm \text{0.8}}$ & 73.6 $ _{\tiny \pm \text{0.8}}$ & 74.0 $ _{\tiny \pm \text{0.5}}$ \\
    ViT-Large ($307$M) & \textbf{78.4} $ _{\tiny \pm \text{0.2}}$& 77.3 $ _{\tiny \pm \text{0.2}}$ & 78.3 $ _{\tiny \pm \text{0.3}}$& 76.1 $ _{\tiny \pm \text{0.3}}$ & 76.8 $ _{\tiny \pm \text{0.3}}$\\
    \end{NiceTabular}
    }
\end{table}

\clearpage
\subsubsection{Robustness analysis}
\begin{figure}[!h]
    \centering
    \includegraphics[width=.95\linewidth]{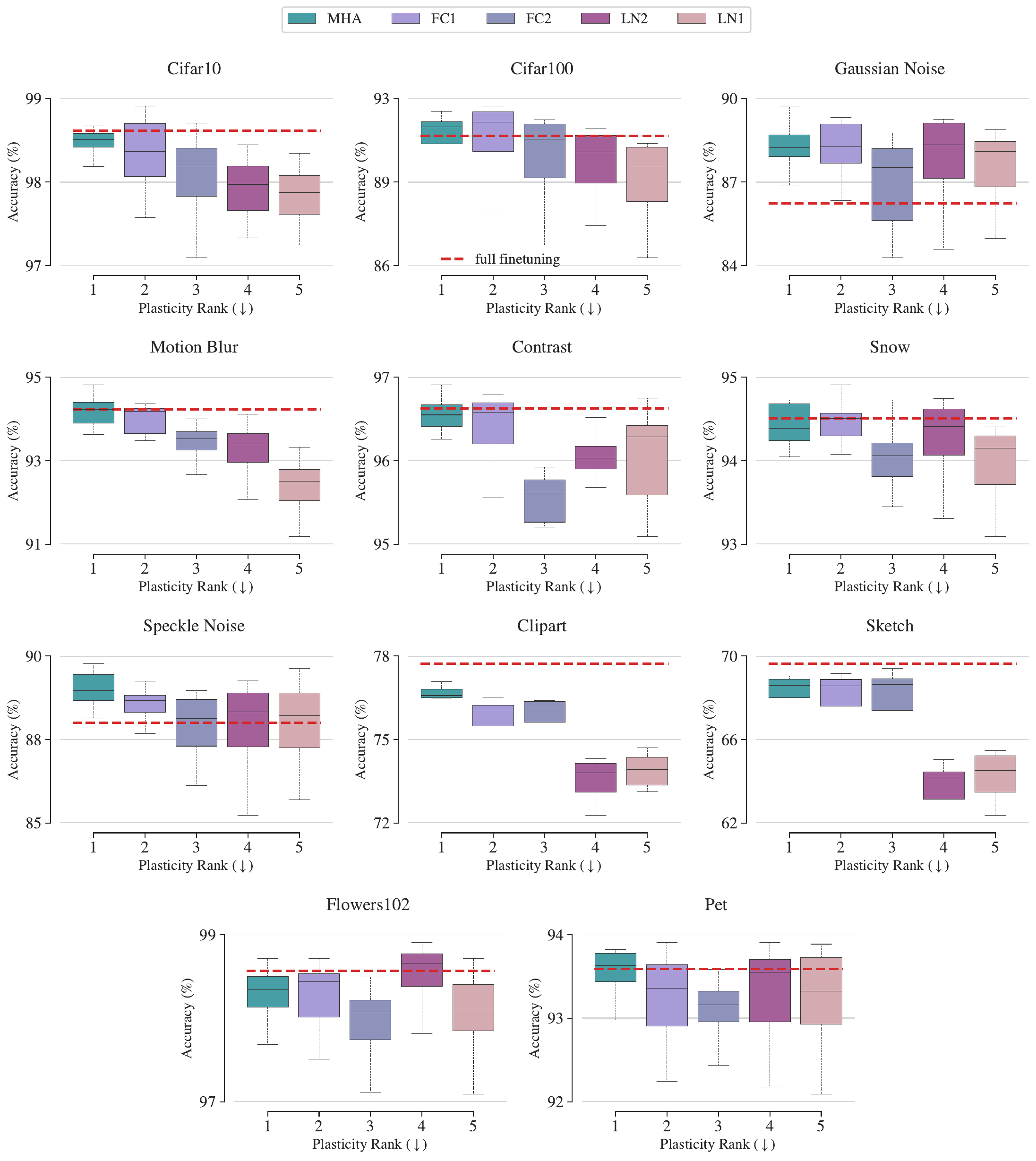}
    \caption{\textbf{Robustness comparison (11 benchmarks).} We display the distribution of the finetuning performance over the learning rates from~\cref{tab:training_details} and $3$ seeds relative to network initialization and dataloaders. We compare transformer components, ordered in terms of decreasing plasticity. Overall, plastic components result in more stable performance. In particular, the attention module consistently displays a small performance variation.}
    \label{fig:robustness_all}
\end{figure}

\label[secinapp]{app:robustness_exp}
In~\cref{fig:robustness_all}, we display the finetuning performance over learning rates and seeds for all benchmarks. Overall, we observe similar patterns to those in~\cref{fig:robustness_training} (left), with plastic components resulting in more stable performance. In particular, we consistently see that finetuning the attention module leads to better and more stable performance.

\clearpage
\subsubsection{Gradient norm analysis}
\label[secinapp]{app:grad_norm_exp}
In~\cref{fig:training_evolution_cifar10_seed_0,fig:training_evolution_cifar10_seed_42,fig:training_evolution_cifar10_seed_3407,fig:training_evolution_cifar100_seed_0,fig:training_evolution_cifar100_seed_42,fig:training_evolution_cifar100_seed_3407,fig:training_evolution_cifar10_c_contrast_seed_0,fig:training_evolution_cifar10_c_contrast_seed_42,fig:training_evolution_cifar10_c_contrast_seed_3407,fig:training_evolution_cifar10_c_gaussian_noise_seed_0,fig:training_evolution_cifar10_c_gaussian_noise_seed_42,fig:training_evolution_cifar10_c_gaussian_noise_seed_3407,fig:training_evolution_cifar10_c_motion_blur_seed_0,fig:training_evolution_cifar10_c_motion_blur_seed_42,fig:training_evolution_cifar10_c_motion_blur_seed_3407,fig:training_evolution_cifar10_c_snow_seed_0,fig:training_evolution_cifar10_c_snow_seed_42,fig:training_evolution_cifar10_c_snow_seed_3407,fig:training_evolution_cifar10_c_speckle_noise_seed_0,fig:training_evolution_cifar10_c_speckle_noise_seed_42,fig:training_evolution_cifar10_c_speckle_noise_seed_3407,fig:training_evolution_domainnet_clipart_seed_0,fig:training_evolution_domainnet_clipart_seed_42,fig:training_evolution_domainnet_clipart_seed_3407,fig:training_evolution_domainnet_sketch_seed_0,fig:training_evolution_domainnet_sketch_seed_42,fig:training_evolution_domainnet_sketch_seed_3407,fig:training_evolution_flowers102_seed_0,fig:training_evolution_flowers102_seed_42,fig:training_evolution_flowers102_seed_3407,fig:training_evolution_pet_seed_0,fig:training_evolution_pet_seed_42,fig:training_evolution_pet_seed_3407}, we display the evolution of the gradient norms and validation loss on all benchmarks, learning rates and seeds. We consistently see that the scale of gradient norms and validation loss descent follows the plasticity ranking from~\cref{sec:analysis}: we observe a faster and better convergence for plastic components, with even more salient benefits on challenging datasets such as Cifar100, Clipart, and Sketch. In particular, we observe lower gradient norms and a less steep descent for the LayerNorms, which is aggravated for low learning rates $\eta$. This showcases the plasticity of LayerNorms to the choice of learning rates compared to components with higher plasticity. For components with high plasticity,  we observe a rather expected training evolution for low learning rates with increasing gradient norms and decreasing validation loss. However, for higher learning rates, we can see that the gradient norms first increase in a steep fashion before slowly decreasing. This can be understood by the model escaping the pretraining minima, passing through sharp regions of the loss landscape before converging to a flat local minima. This behavior can be seen on Clipart~\cref{fig:training_evolution_domainnet_clipart_seed_0}, for instance, and is particularly salient for the attention module. This is reminiscent of~\citet{park2022how}, who showed that the multihead self-attention module flattens the loss landscape, which leads to better generalization~\citep{ilbert2024samformer,foret2021sharpnessaware,chen2022when}. 
\begin{figure}[!h]
    \centering
    \includegraphics[width=\linewidth]{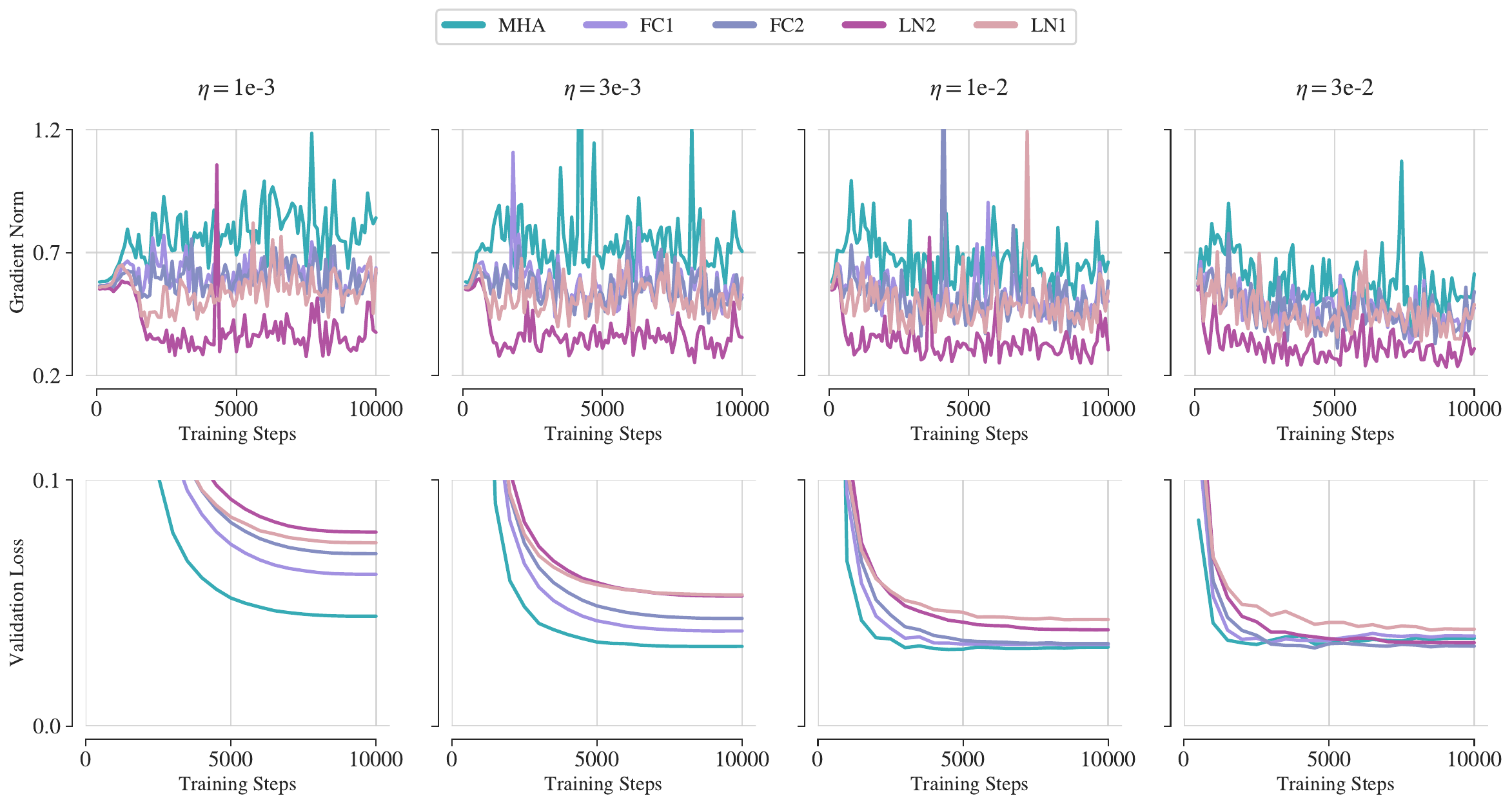}
    \caption{\textbf{Training dynamics on Cifar10 with seed $0$.} We display the evolution during training of the gradient norms (\textbf{top}) and the validation loss (\textbf{bottom}) of each finetuning configuration of~\cref{tab:model_configurations}, with increasing learning rate $\eta$ from \textbf{left to right}. Components are ordered in terms of decreasing plasticity in the legend. Plastic components have higher gradient norms, which leads to a steeper descent in the validation loss and better downstream performance. The benefits of plasticity are even more salient with low learning rates. Overall, higher plasticity leads to better optimization and generalization.
    }
    \label{fig:training_evolution_cifar10_seed_0}
\end{figure}

\begin{figure}[!h]
    \centering
    \includegraphics[width=\linewidth]{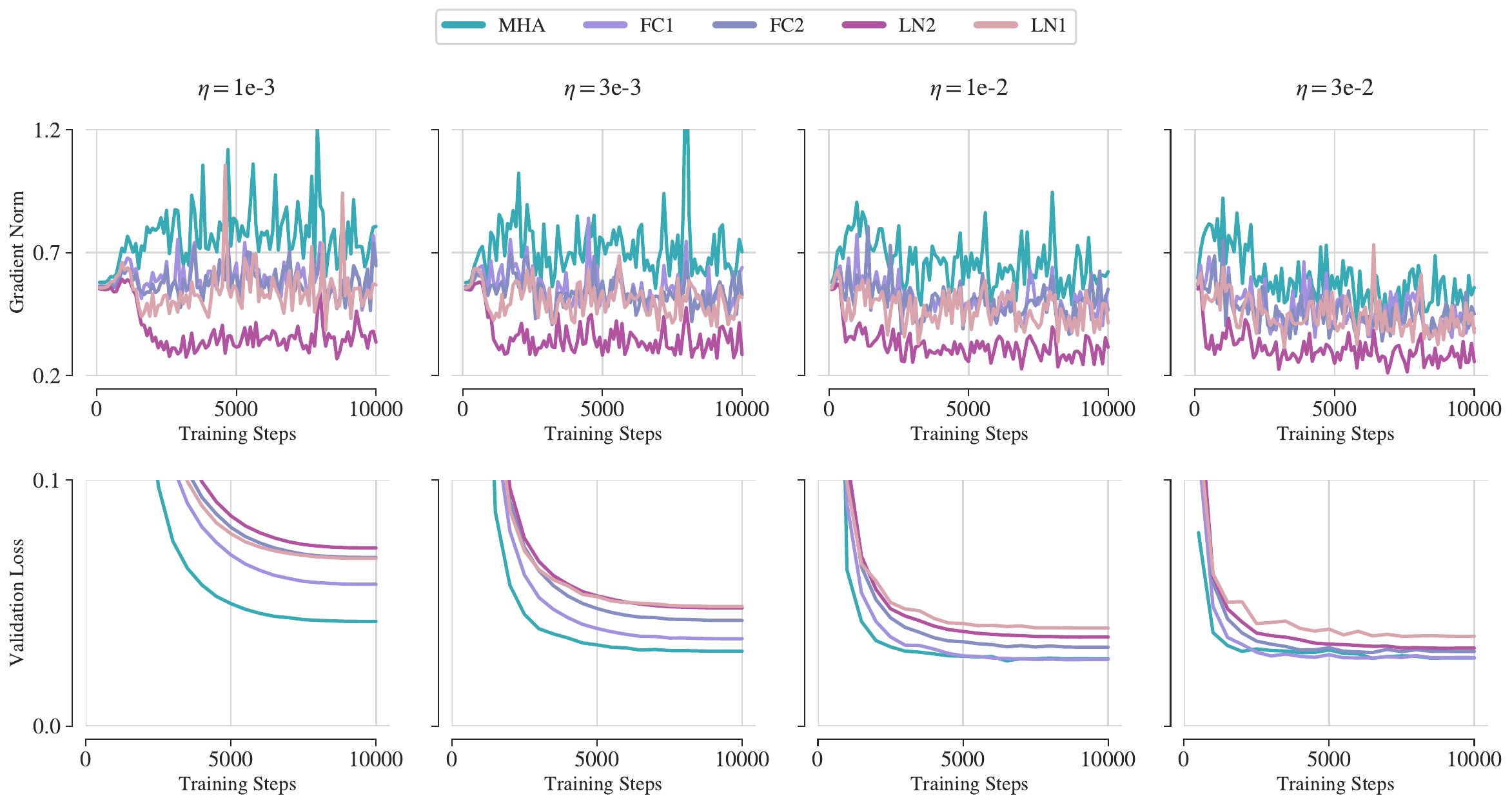}
    \caption{\textbf{Training dynamics on Cifar10 with seed $42$.} Akin to~\cref{fig:training_evolution_cifar10_seed_0}, we observe the same consistent pattern of faster and better convergence for components with high plasticity.}
    \label{fig:training_evolution_cifar10_seed_42}
\end{figure}

\begin{figure}[!h]
    \centering
    \includegraphics[width=\linewidth]{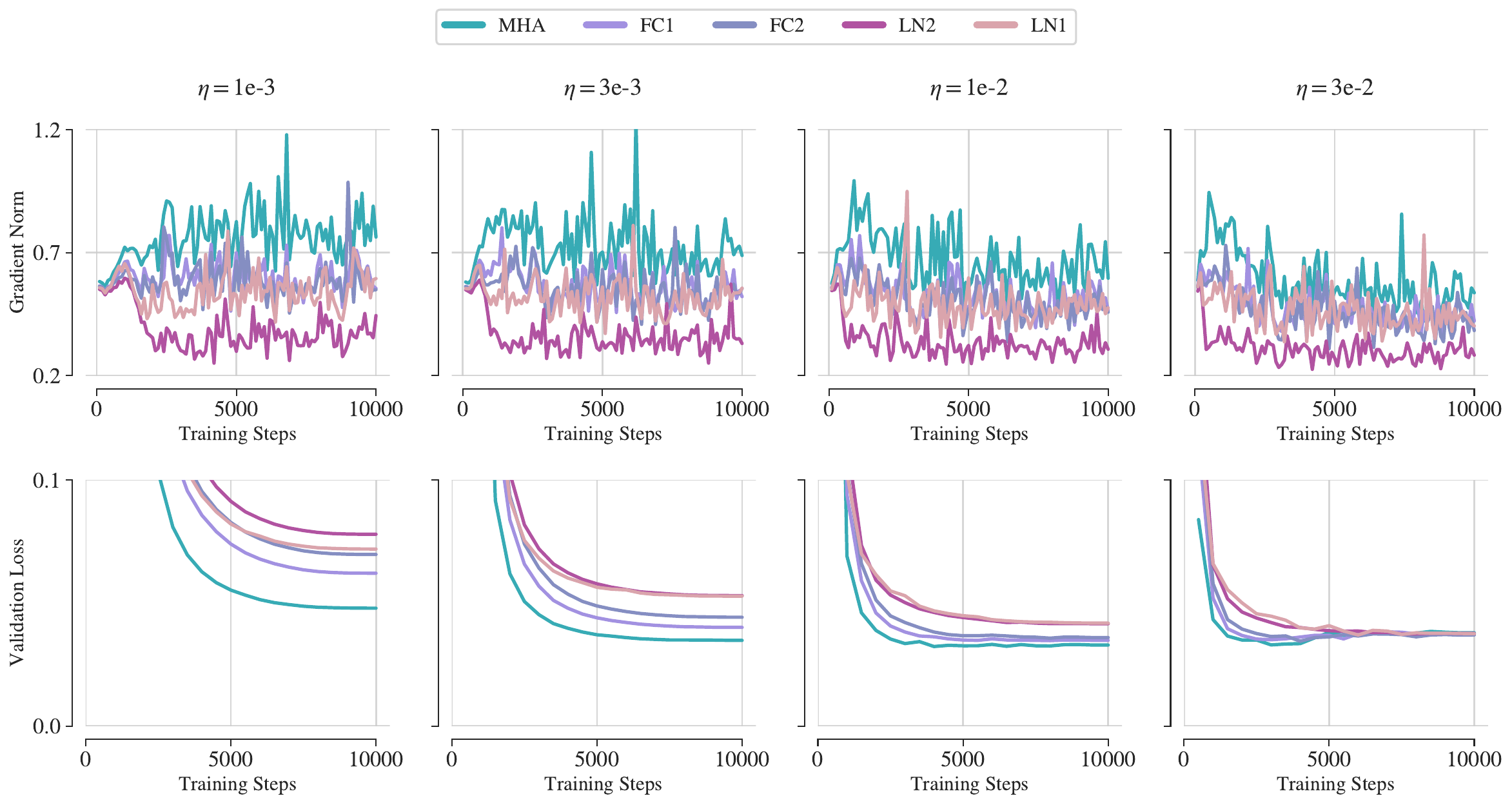}
    \caption{\textbf{Training dynamics on Cifar10 with seed $3407$.} Akin to~\cref{fig:training_evolution_cifar10_seed_0}, we observe the same consistent pattern of faster and better convergence for components with high plasticity.}
    \label{fig:training_evolution_cifar10_seed_3407}
\end{figure}

\begin{figure}[!h]
    \centering
    \includegraphics[width=\linewidth]{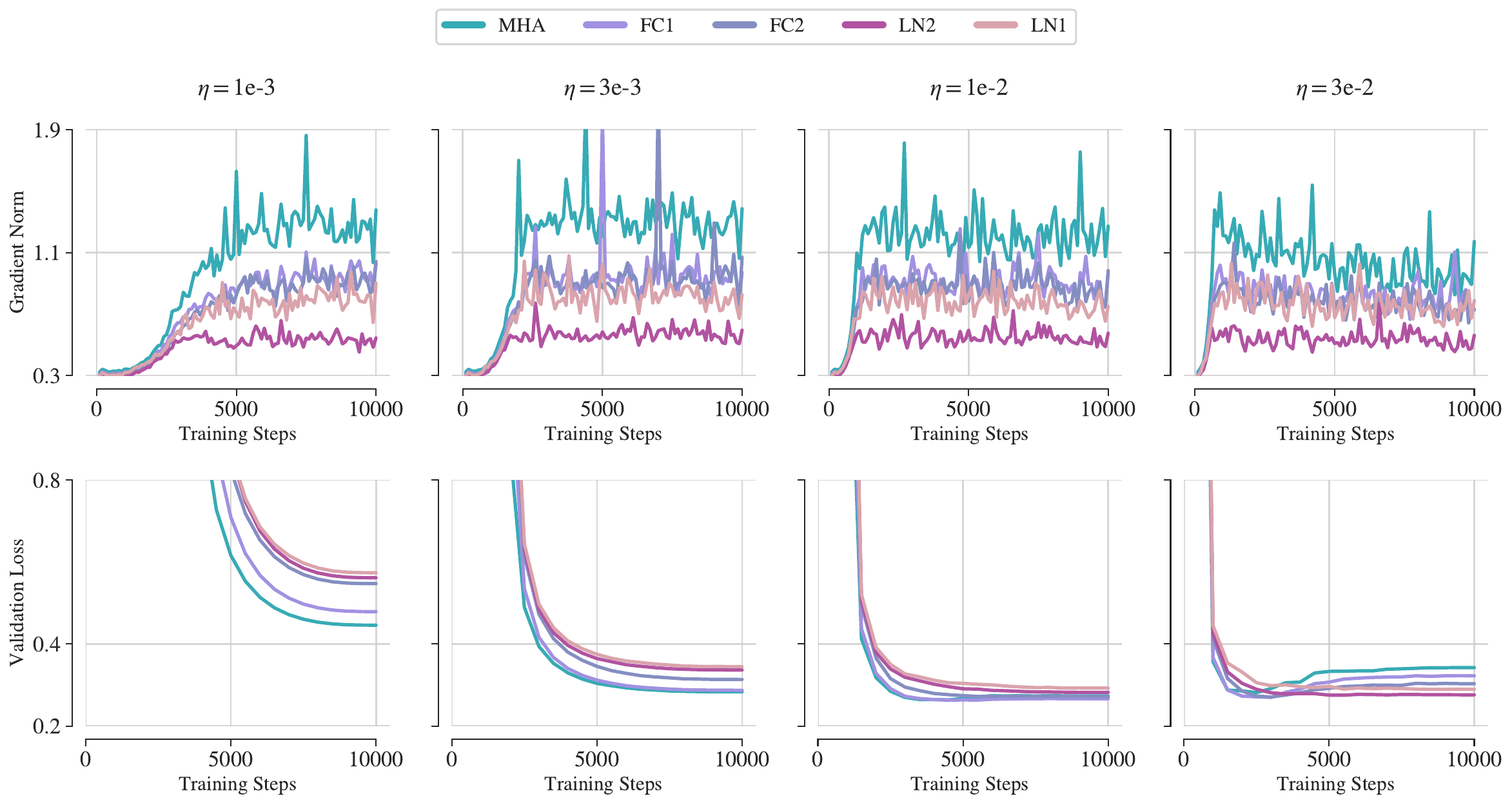}
    \caption{\textbf{Training dynamics on Cifar100 with seed $0$.} We display the evolution during training of the gradient norms (\textbf{top}) and the validation loss (\textbf{bottom}) of each finetuning configuration of~\cref{tab:model_configurations}, with increasing learning rate $\eta$ from \textbf{left to right}. Components are ordered in terms of decreasing plasticity in the legend. Plastic components have higher gradient norms, which leads to a steeper descent in the validation loss and better downstream performance. The benefits of plasticity are even more salient with low learning rates. Overall, higher plasticity leads to better optimization and generalization.
    }
    \label{fig:training_evolution_cifar100_seed_0}
\end{figure}

\begin{figure}[!h]
    \centering
    \includegraphics[width=\linewidth]{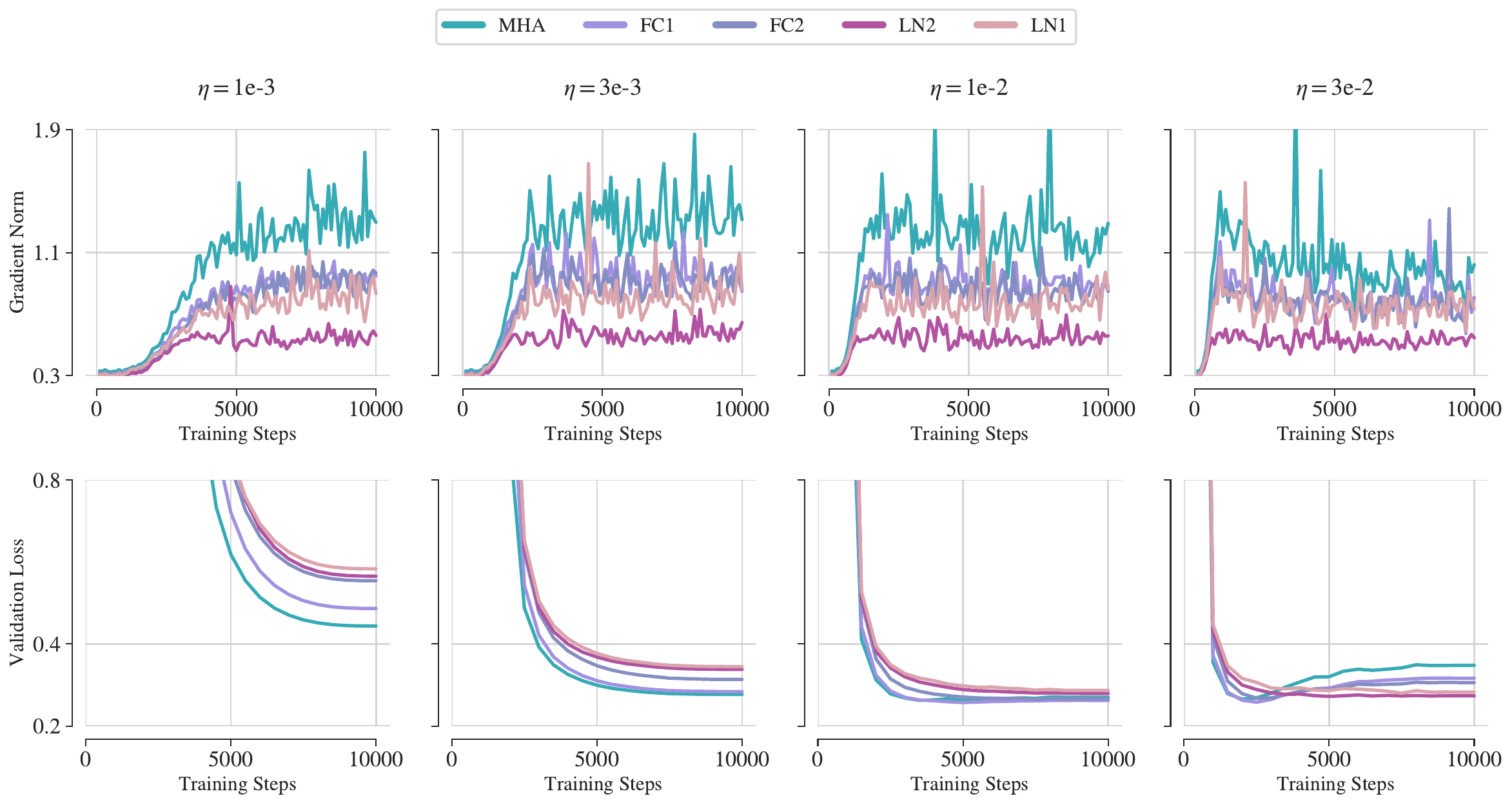}
    \caption{\textbf{Training dynamics on Cifar100 with seed $42$.} Akin to~\cref{fig:training_evolution_cifar100_seed_0}, we observe the same consistent pattern of faster and better convergence for components with high plasticity.}
    \label{fig:training_evolution_cifar100_seed_42}
\end{figure}

\begin{figure}[!h]
    \centering
    \includegraphics[width=\linewidth]{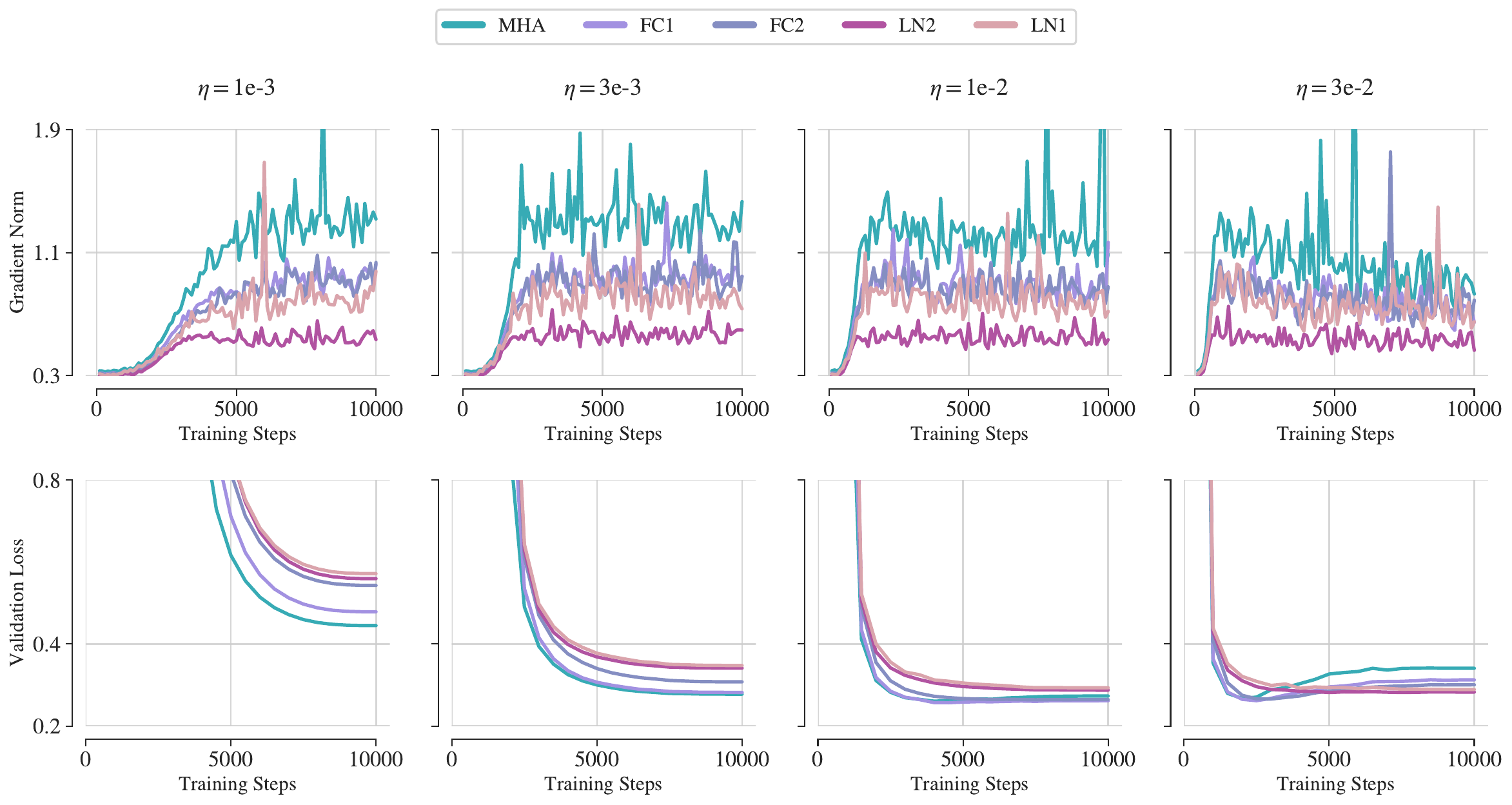}
    \caption{\textbf{Training dynamics on Cifar100 with seed $3407$.} Akin to~\cref{fig:training_evolution_cifar100_seed_0}, we observe the same consistent pattern of faster and better convergence for components with high plasticity.}
    \label{fig:training_evolution_cifar100_seed_3407}
\end{figure}

\begin{figure}[!h]
    \centering
    \includegraphics[width=\linewidth]{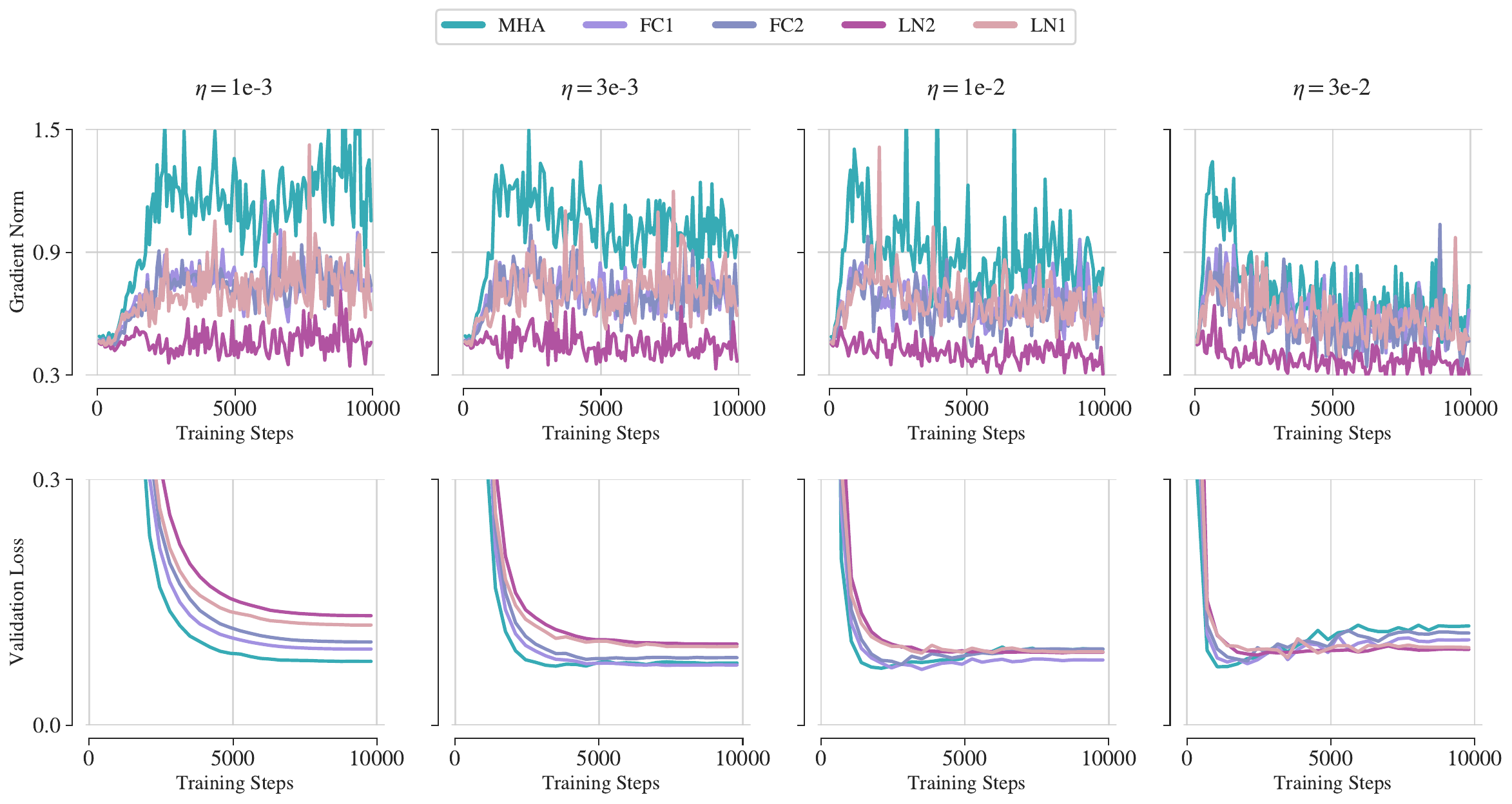}
    \caption{\textbf{Training dynamics on Contrast with seed $0$.} We display the evolution during training of the gradient norms (\textbf{top}) and the validation loss (\textbf{bottom}) of each finetuning configuration of~\cref{tab:model_configurations}, with increasing learning rate $\eta$ from \textbf{left to right}. Components are ordered in terms of decreasing plasticity in the legend. Plastic components have higher gradient norms, which leads to a steeper descent in the validation loss and better downstream performance. The benefits of plasticity are even more salient with low learning rates. Overall, higher plasticity leads to better optimization and generalization.
    }
    \label{fig:training_evolution_cifar10_c_contrast_seed_0}
\end{figure}

\begin{figure}[!h]
    \centering
    \includegraphics[width=\linewidth]{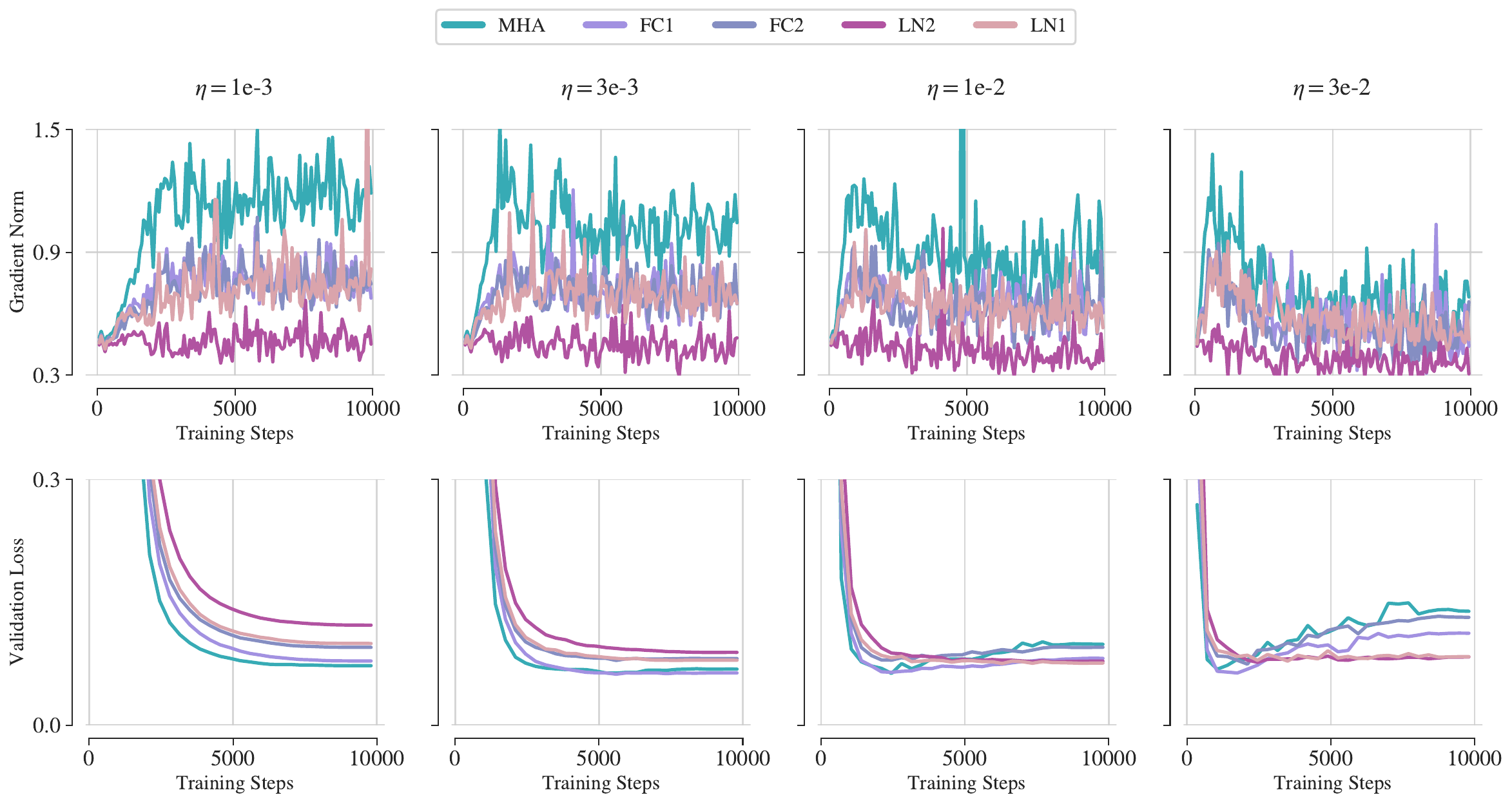}
    \caption{\textbf{Training dynamics on Contrast with seed $42$.} Akin to~\cref{fig:training_evolution_cifar10_c_contrast_seed_0}, we observe the same consistent pattern of faster and better convergence for components with high plasticity.}
    \label{fig:training_evolution_cifar10_c_contrast_seed_42}
\end{figure}

\begin{figure}[!h]
    \centering
    \includegraphics[width=\linewidth]{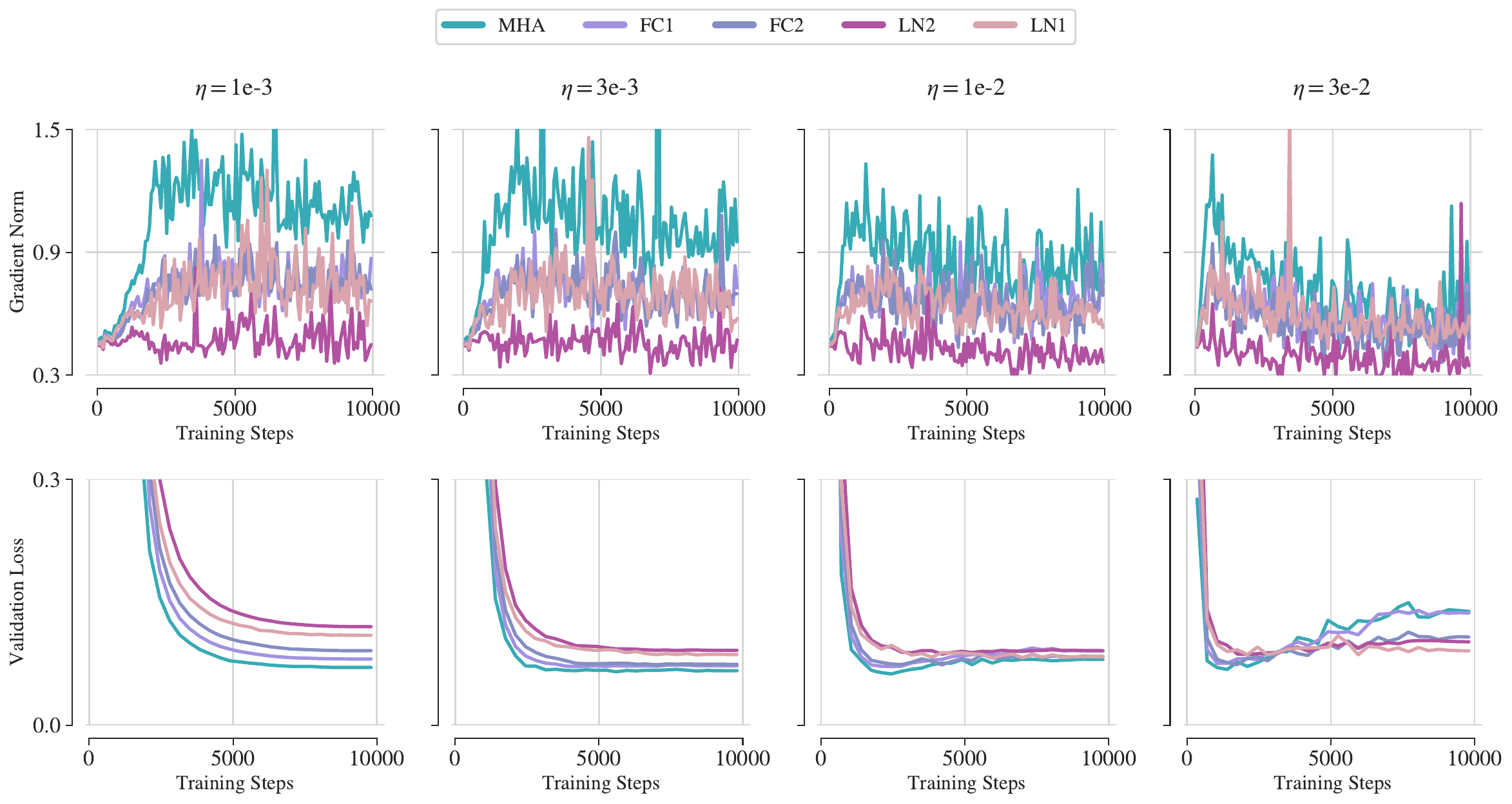}
    \caption{\textbf{Training dynamics on Contrast with seed $3407$.} Akin to~\cref{fig:training_evolution_cifar10_c_contrast_seed_0}, we observe the same consistent pattern of faster and better convergence for components with high plasticity.}
    \label{fig:training_evolution_cifar10_c_contrast_seed_3407}
\end{figure}

\begin{figure}[!h]
    \centering
    \includegraphics[width=\linewidth]{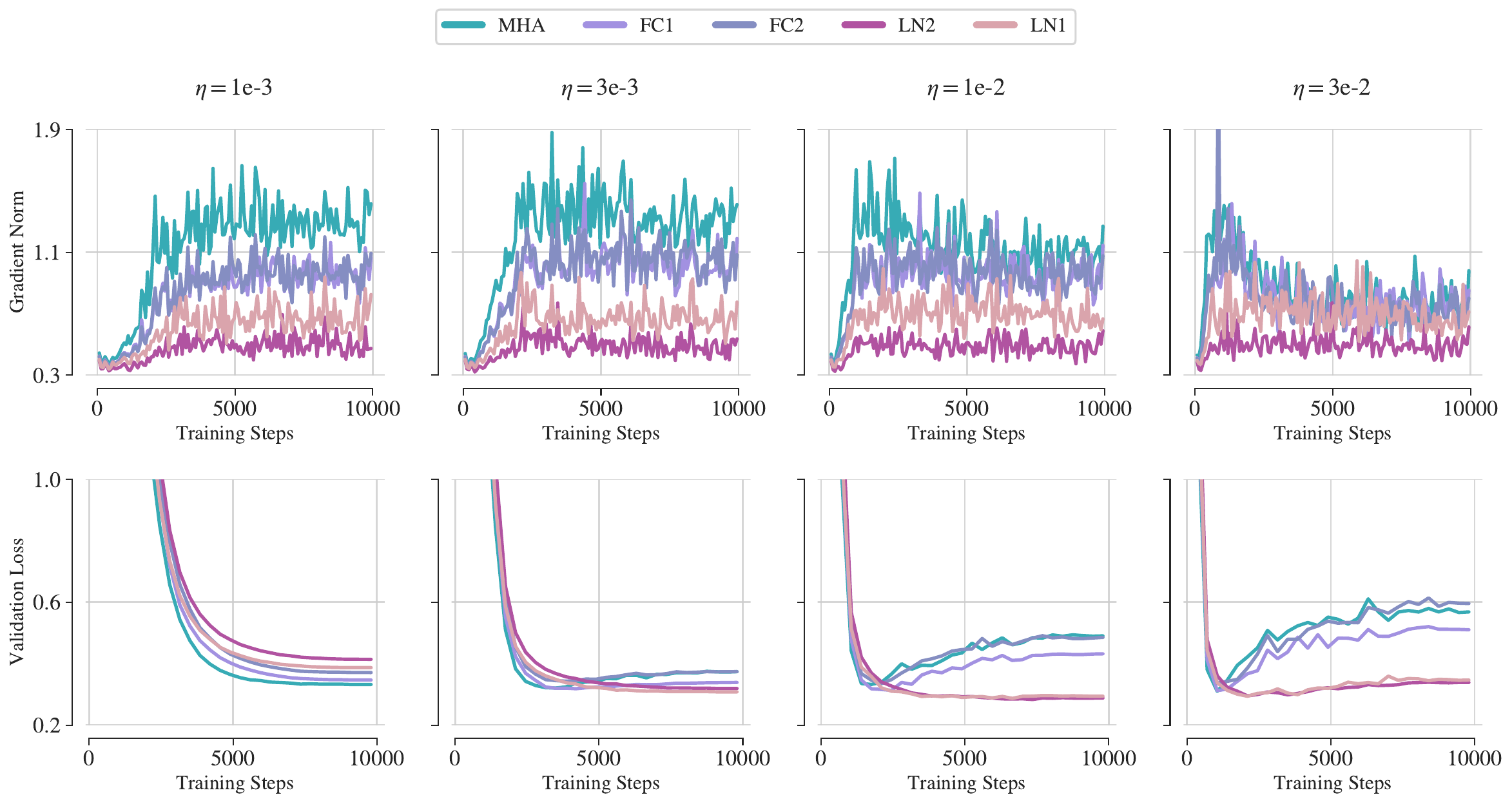}
    \caption{\textbf{Training dynamics on Gaussian Noise with seed $0$.} We display the evolution during training of the gradient norms (\textbf{top}) and the validation loss (\textbf{bottom}) of each finetuning configuration of~\cref{tab:model_configurations}, with increasing learning rate $\eta$ from \textbf{left to right}. Components are ordered in terms of decreasing plasticity in the legend. Plastic components have higher gradient norms, which leads to a steeper descent in the validation loss and better downstream performance. The benefits of plasticity are even more salient with low learning rates. Overall, higher plasticity leads to better optimization and generalization.
    }
    \label{fig:training_evolution_cifar10_c_gaussian_noise_seed_0}
\end{figure}

\begin{figure}[!h]
    \centering
    \includegraphics[width=\linewidth]{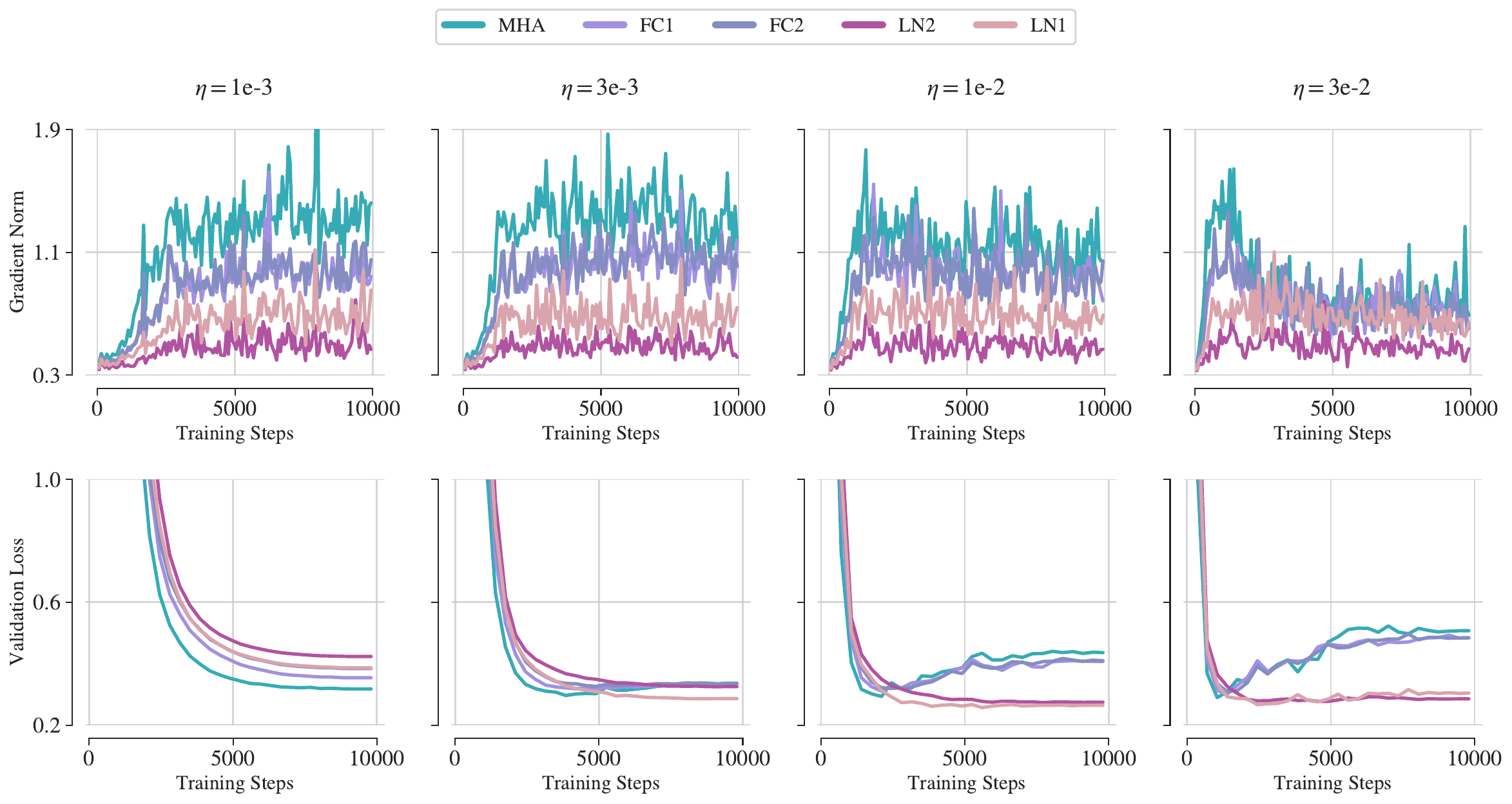}
    \caption{\textbf{Training dynamics on Gaussian Noise with seed $42$.} Akin to~\cref{fig:training_evolution_cifar10_c_gaussian_noise_seed_0}, we observe the same consistent pattern of faster and better convergence for components with high plasticity.}
    \label{fig:training_evolution_cifar10_c_gaussian_noise_seed_42}
\end{figure}

\begin{figure}[!h]
    \centering
    \includegraphics[width=\linewidth]{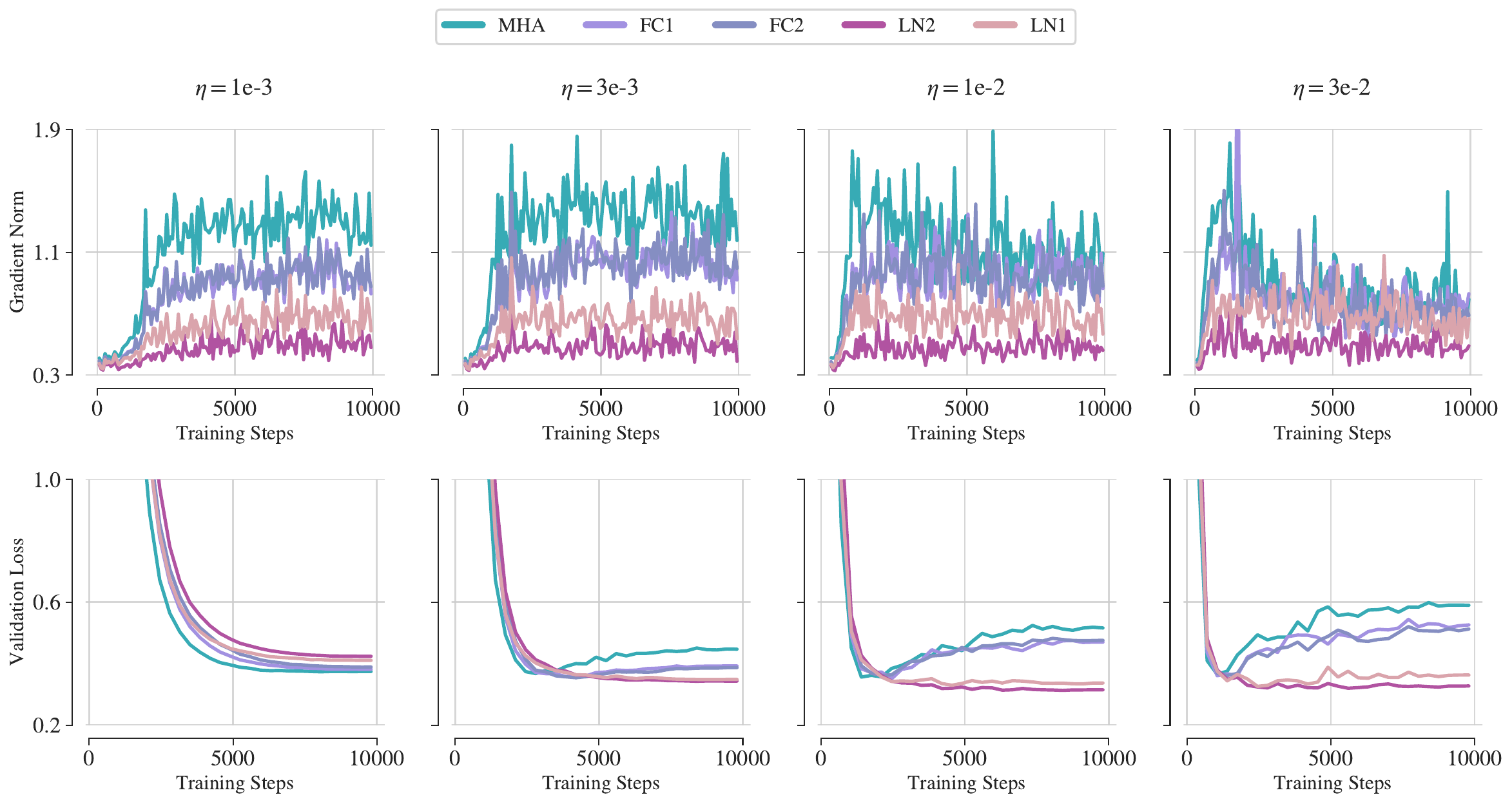}
    \caption{\textbf{Training dynamics on Gaussian Noise with seed $3407$.} Akin to~\cref{fig:training_evolution_cifar10_c_gaussian_noise_seed_0}, we observe the same consistent pattern of faster and better convergence for components with high plasticity.}
    \label{fig:training_evolution_cifar10_c_gaussian_noise_seed_3407}
\end{figure}

\begin{figure}[!h]
    \centering
    \includegraphics[width=\linewidth]{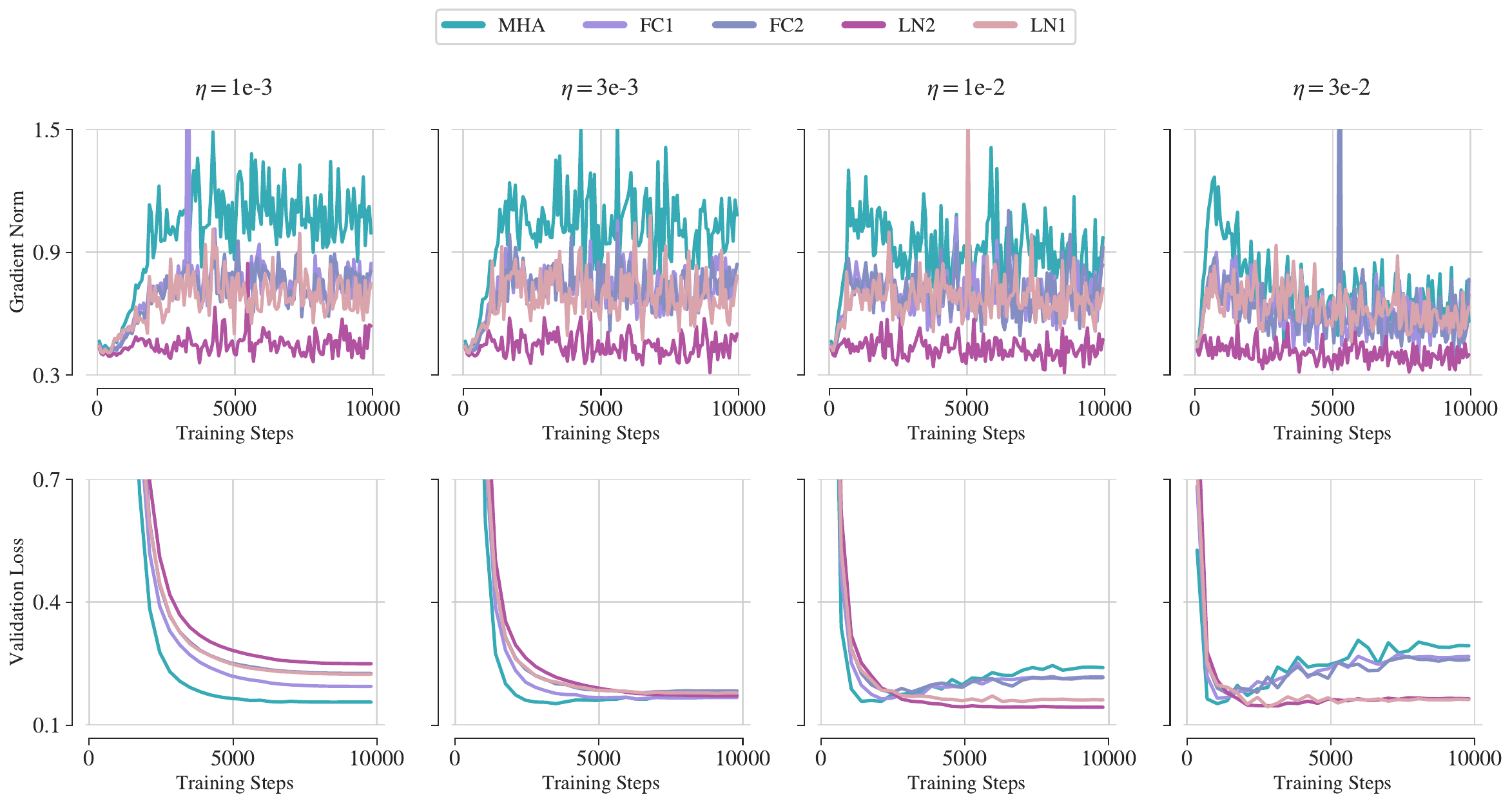}
    \caption{\textbf{Training dynamics on Motion Blur with seed $0$.} We display the evolution during training of the gradient norms (\textbf{top}) and the validation loss (\textbf{bottom}) of each finetuning configuration of~\cref{tab:model_configurations}, with increasing learning rate $\eta$ from \textbf{left to right}. Components are ordered in terms of decreasing plasticity in the legend. Plastic components have higher gradient norms, which leads to a steeper descent in the validation loss and better downstream performance. The benefits of plasticity are even more salient with low learning rates. Overall, higher plasticity leads to better optimization and generalization.
    }
    \label{fig:training_evolution_cifar10_c_motion_blur_seed_0}
\end{figure}

\begin{figure}[!h]
    \centering
    \includegraphics[width=\linewidth]{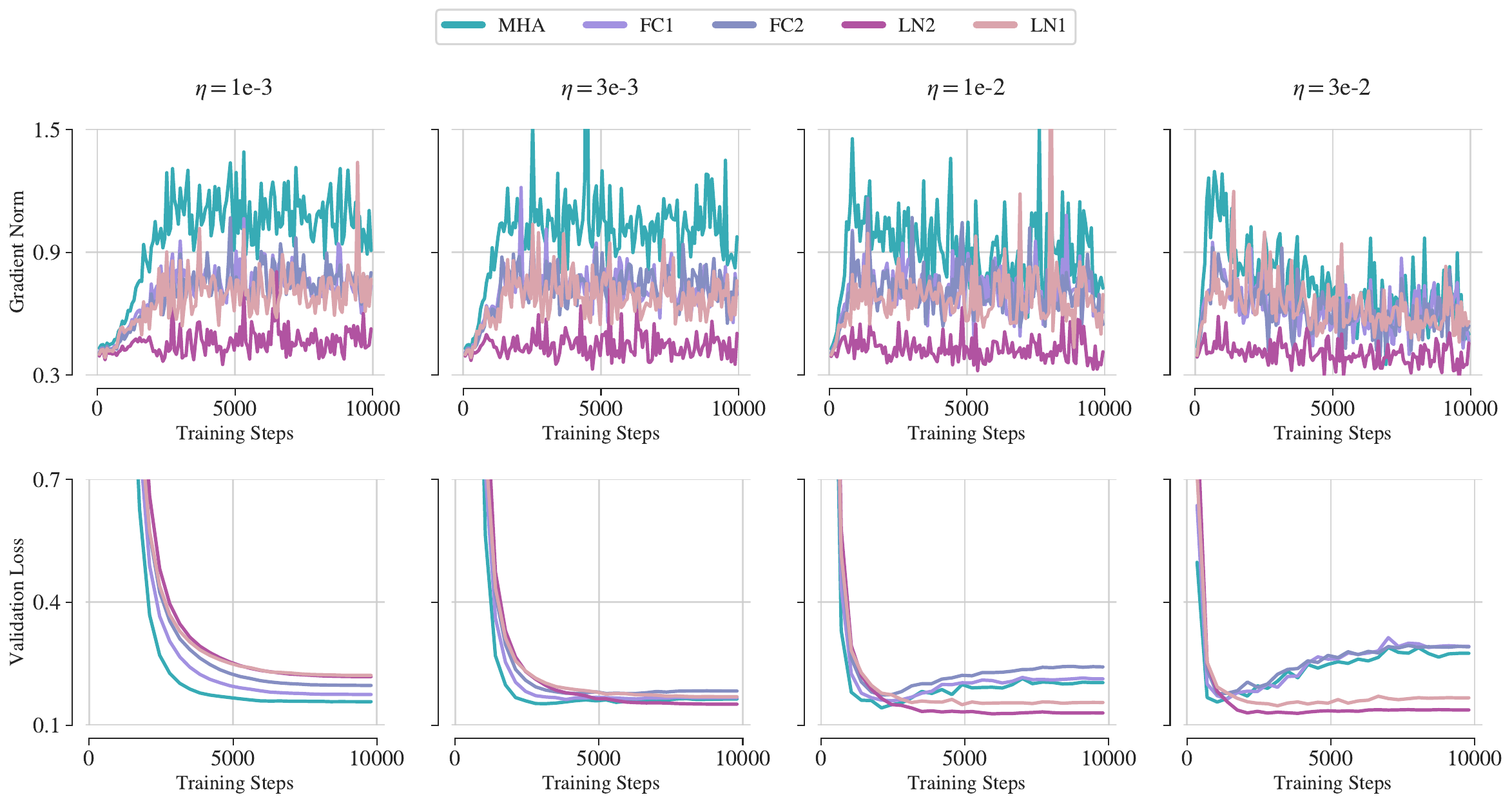}
    \caption{\textbf{Training dynamics on Motion Blur with seed $42$.} Akin to~\cref{fig:training_evolution_cifar10_c_motion_blur_seed_0}, we observe the same consistent pattern of faster and better convergence for components with high plasticity.}
    \label{fig:training_evolution_cifar10_c_motion_blur_seed_42}
\end{figure}

\begin{figure}[!h]
    \centering
    \includegraphics[width=\linewidth]{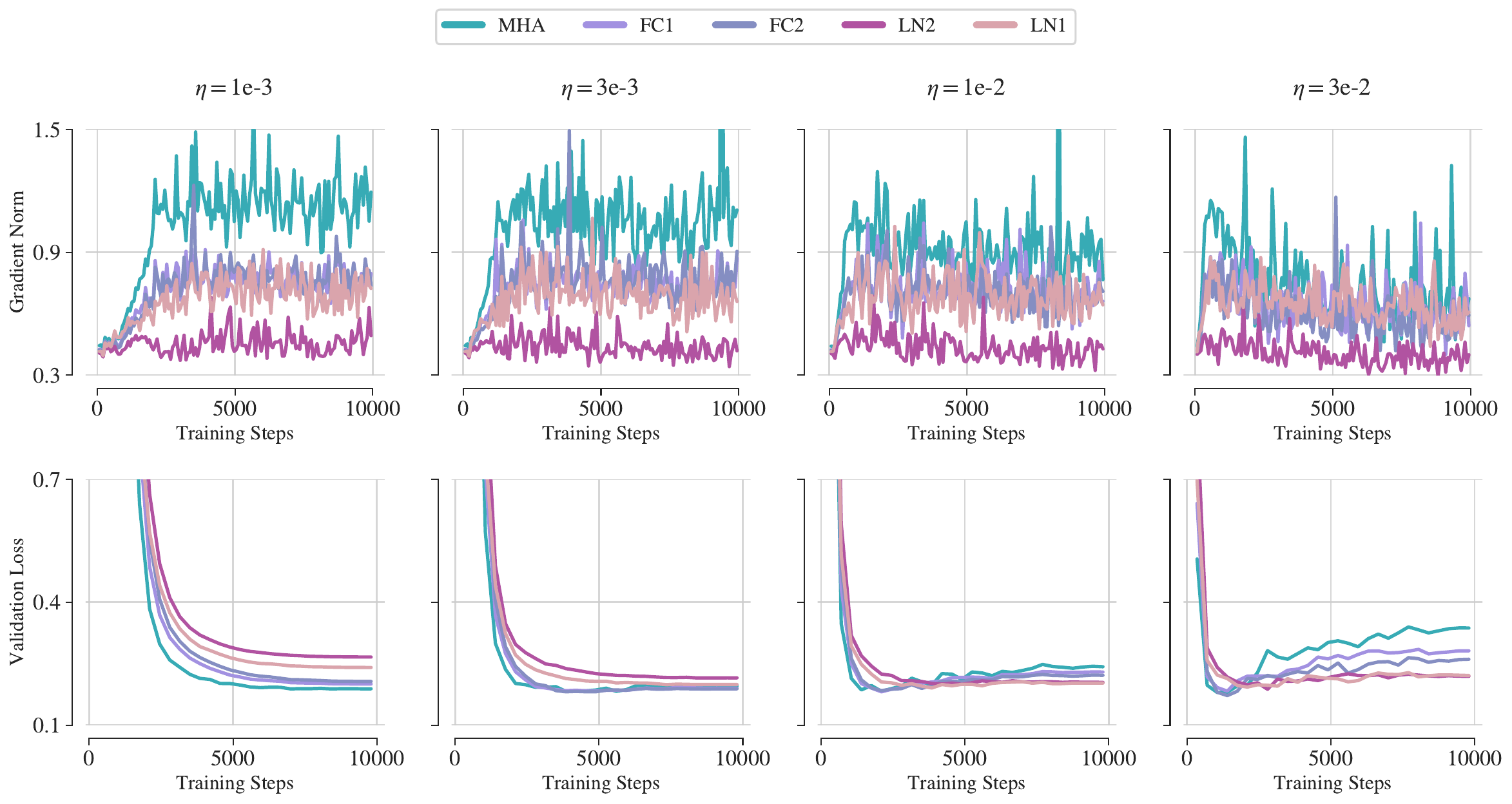}
    \caption{\textbf{Training dynamics on Motion Blur with seed $3407$.} Akin to~\cref{fig:training_evolution_cifar10_c_motion_blur_seed_0}, we observe the same consistent pattern of faster and better convergence for components with high plasticity.}
    \label{fig:training_evolution_cifar10_c_motion_blur_seed_3407}
\end{figure}

\begin{figure}[!h]
    \centering
    \includegraphics[width=\linewidth]{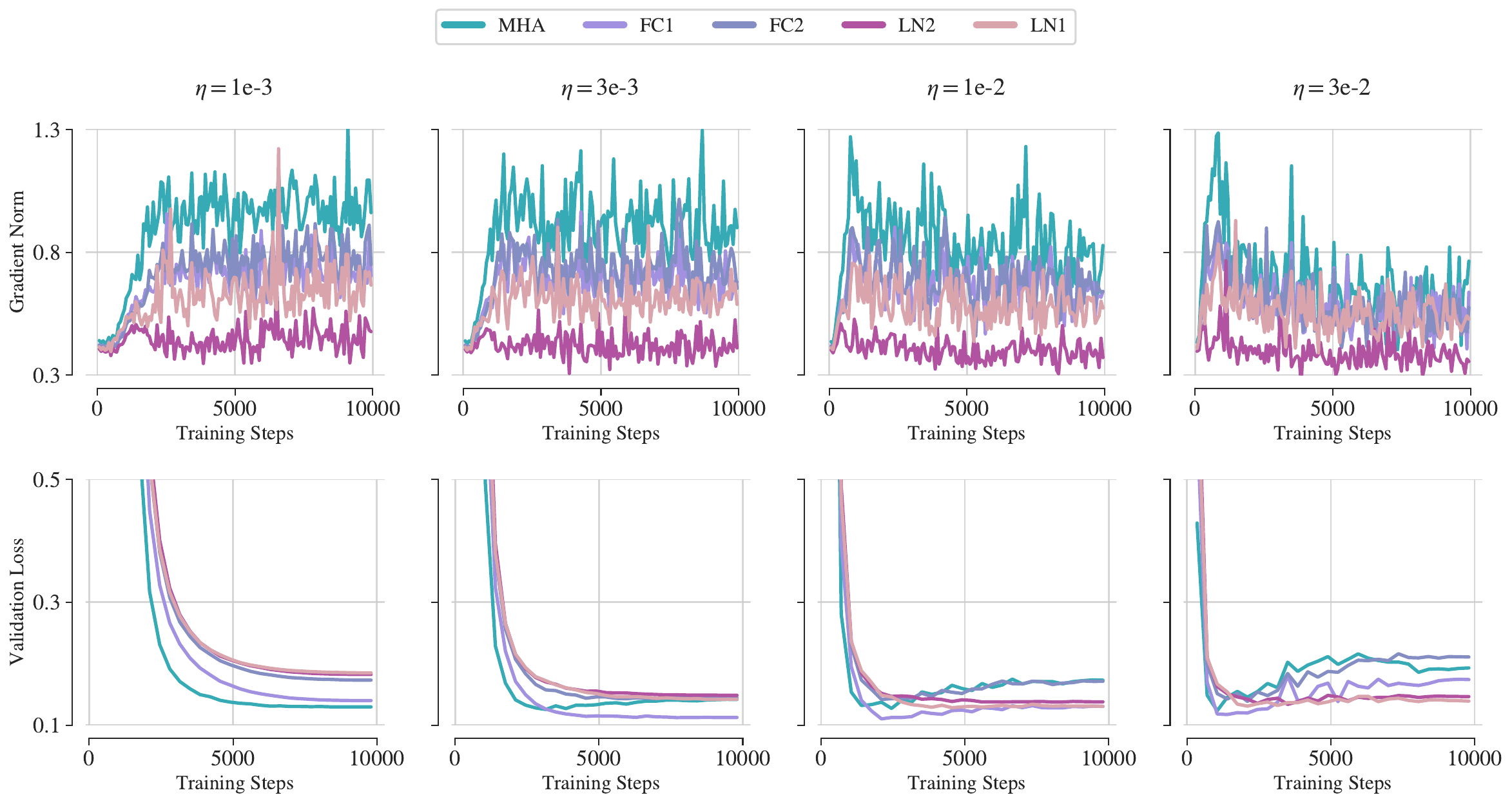}
    \caption{\textbf{Training dynamics on Snow with seed $0$.} We display the evolution during training of the gradient norms (\textbf{top}) and the validation loss (\textbf{bottom}) of each finetuning configuration of~\cref{tab:model_configurations}, with increasing learning rate $\eta$ from \textbf{left to right}. Components are ordered in terms of decreasing plasticity in the legend. Plastic components have higher gradient norms, which leads to a steeper descent in the validation loss and better downstream performance. The benefits of plasticity are even more salient with low learning rates. Overall, higher plasticity leads to better optimization and generalization.
    }
    \label{fig:training_evolution_cifar10_c_snow_seed_0}
\end{figure}

\begin{figure}[!h]
    \centering
    \includegraphics[width=\linewidth]{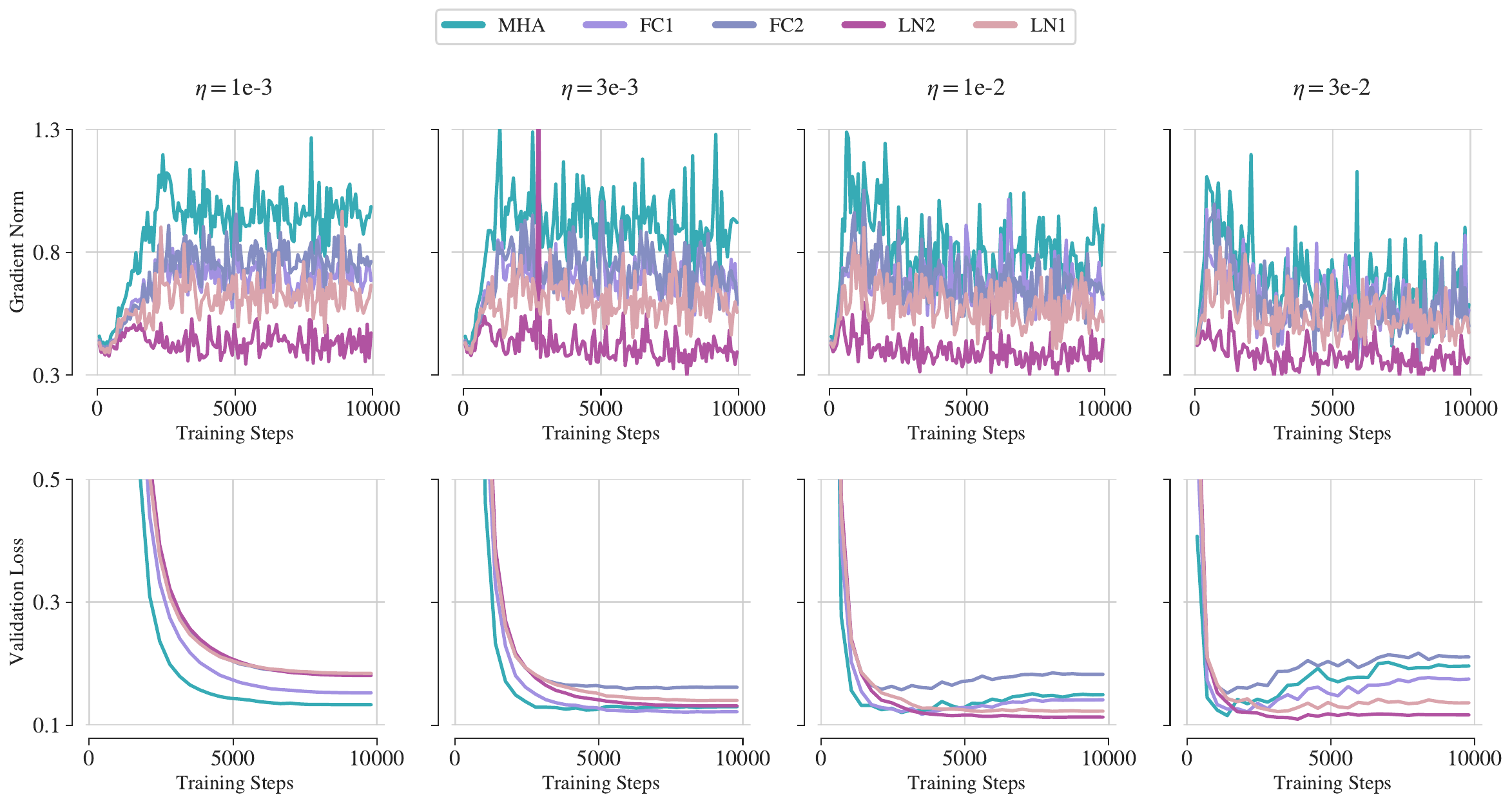}
    \caption{\textbf{Training dynamics on Snow with seed $42$.} Akin to~\cref{fig:training_evolution_cifar10_c_snow_seed_0}, we observe the same consistent pattern of faster and better convergence for components with high plasticity.}
    \label{fig:training_evolution_cifar10_c_snow_seed_42}
\end{figure}

\begin{figure}[!h]
    \centering
    \includegraphics[width=\linewidth]{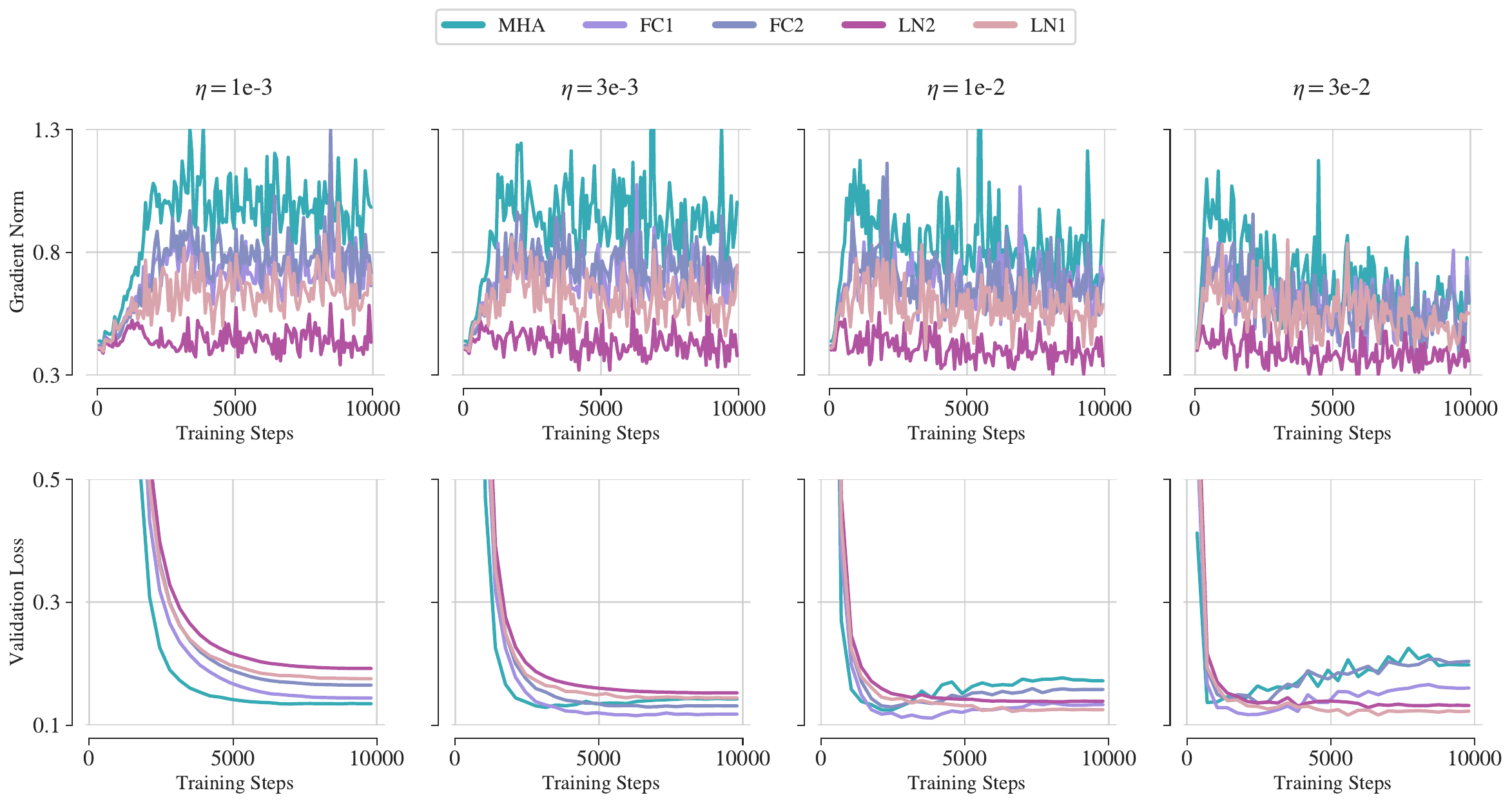}
    \caption{\textbf{Training dynamics on Snow with seed $3407$.} Akin to~\cref{fig:training_evolution_cifar10_c_snow_seed_0}, we observe the same consistent pattern of faster and better convergence for components with high plasticity.}
    \label{fig:training_evolution_cifar10_c_snow_seed_3407}
\end{figure}

\begin{figure}[!h]
    \centering
    \includegraphics[width=\linewidth]{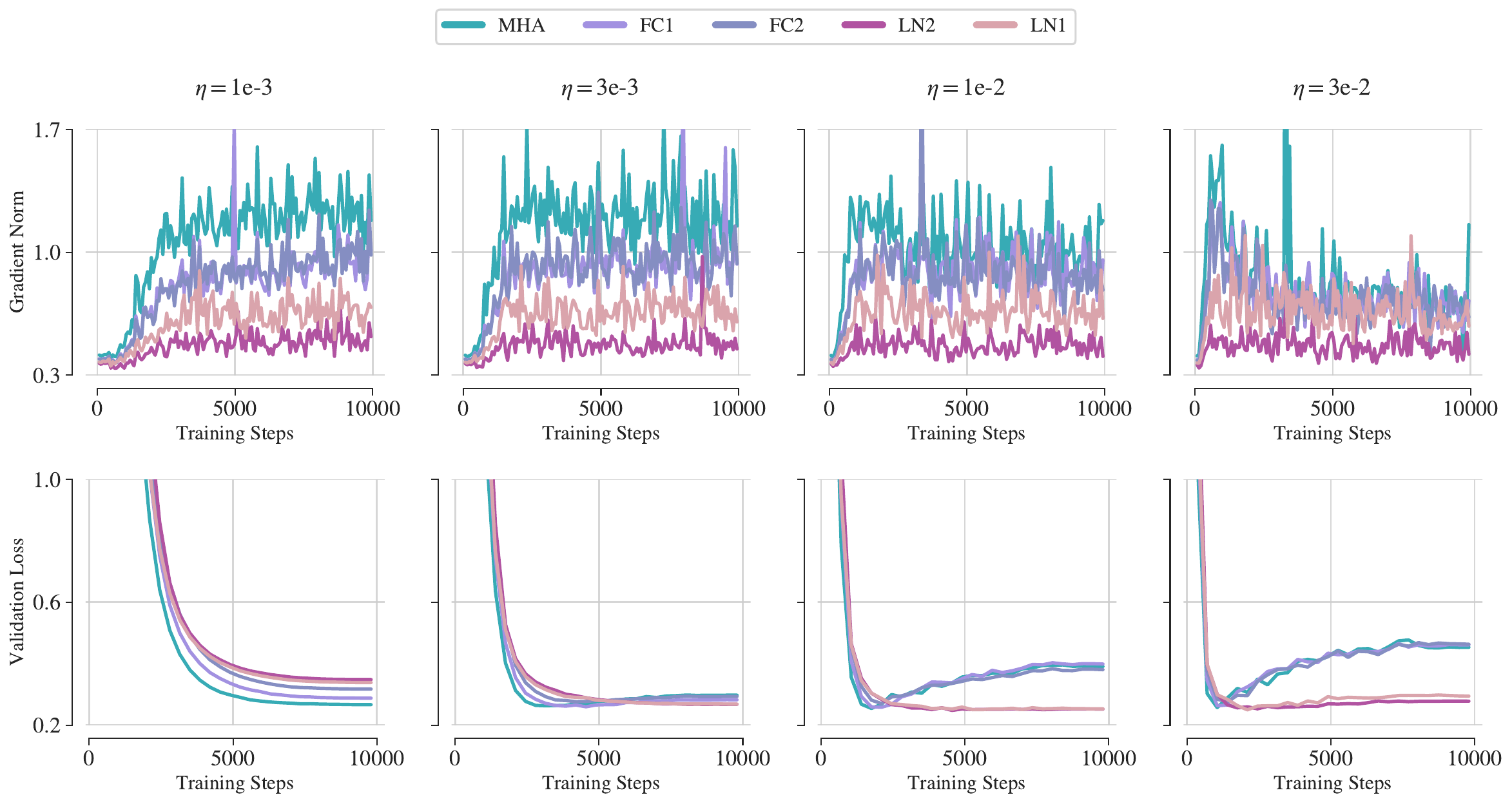}
    \caption{\textbf{Training dynamics on Speckle Noise with seed $0$.} We display the evolution during training of the gradient norms (\textbf{top}) and the validation loss (\textbf{bottom}) of each finetuning configuration of~\cref{tab:model_configurations}, with increasing learning rate $\eta$ from \textbf{left to right}. Components are ordered in terms of decreasing plasticity in the legend. Plastic components have higher gradient norms, which leads to a steeper descent in the validation loss and better downstream performance. The benefits of plasticity are even more salient with low learning rates. Overall, higher plasticity leads to better optimization and generalization.
    }
    \label{fig:training_evolution_cifar10_c_speckle_noise_seed_0}
\end{figure}

\begin{figure}[!h]
    \centering
    \includegraphics[width=\linewidth]{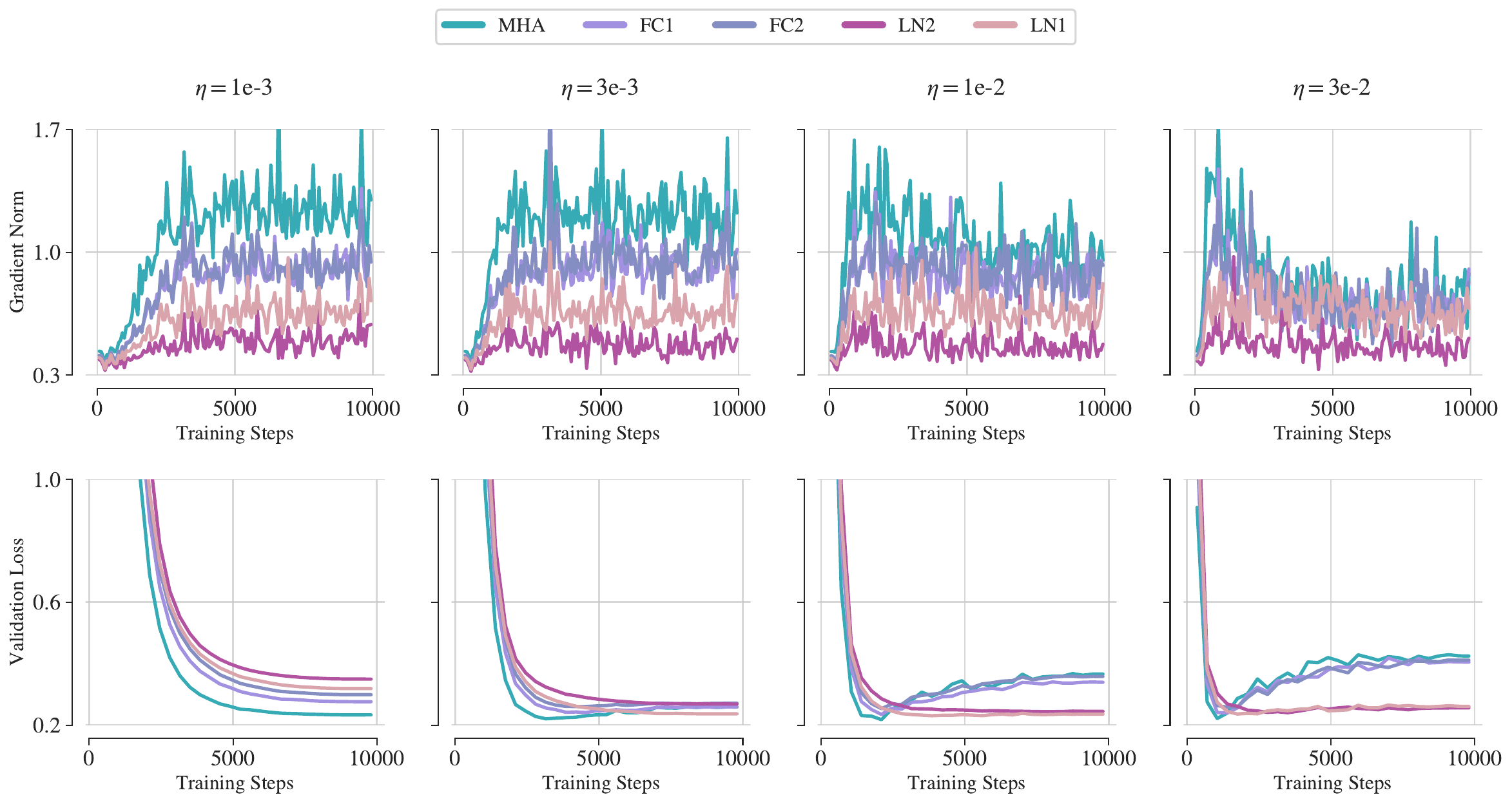}
    \caption{\textbf{Training dynamics on Speckle Noise with seed $42$.} Akin to~\cref{fig:training_evolution_cifar10_c_speckle_noise_seed_0}, we observe the same consistent pattern of faster and better convergence for components with high plasticity.}
    \label{fig:training_evolution_cifar10_c_speckle_noise_seed_42}
\end{figure}

\begin{figure}[!h]
    \centering
    \includegraphics[width=\linewidth]{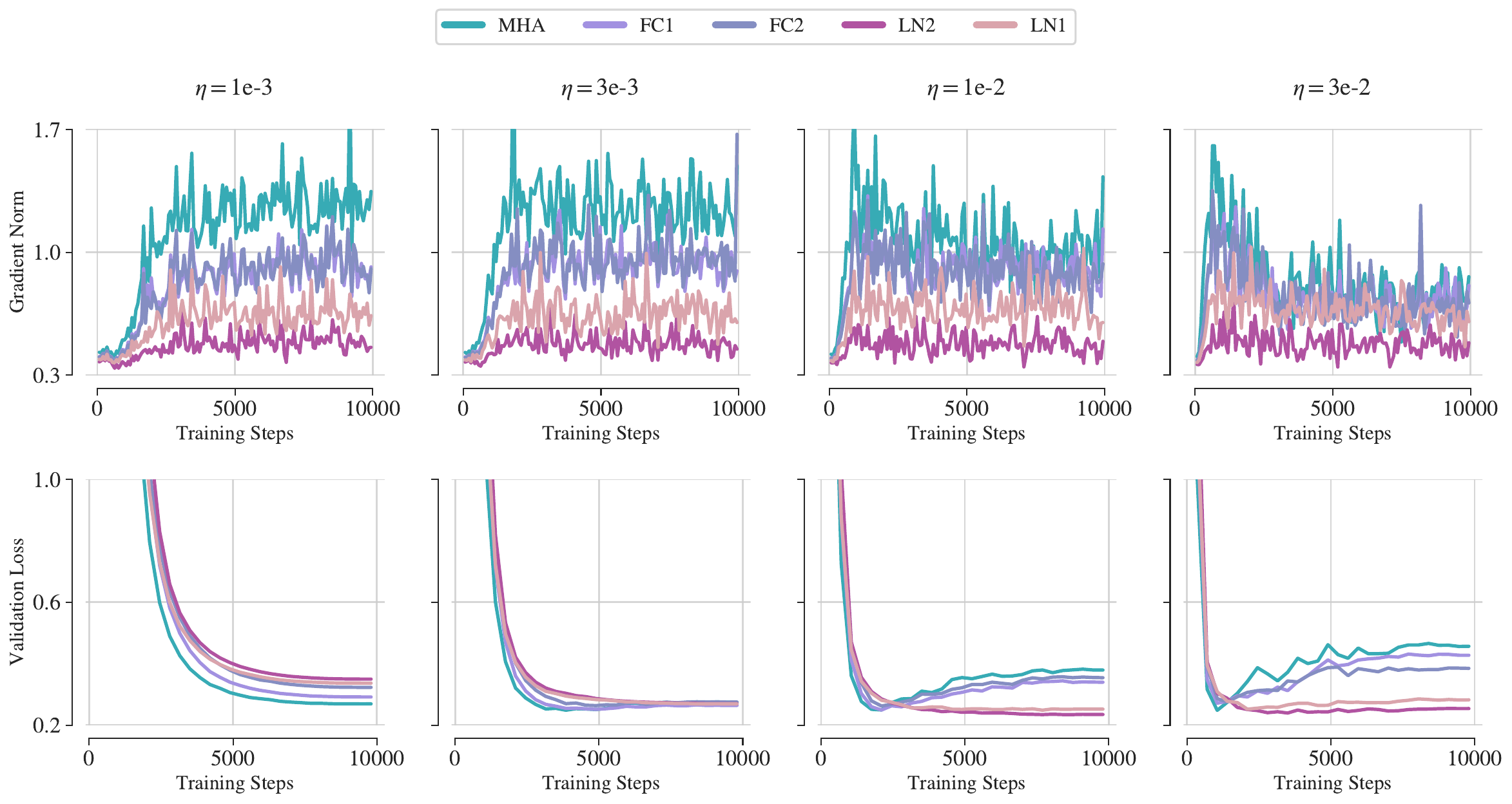}
    \caption{\textbf{Training dynamics on Speckle Noise with seed $3407$.} Akin to~\cref{fig:training_evolution_cifar10_c_speckle_noise_seed_0}, we observe the same consistent pattern of faster and better convergence for components with high plasticity.}
    \label{fig:training_evolution_cifar10_c_speckle_noise_seed_3407}
\end{figure}

\begin{figure}[!h]
    \centering
    \includegraphics[width=\linewidth]{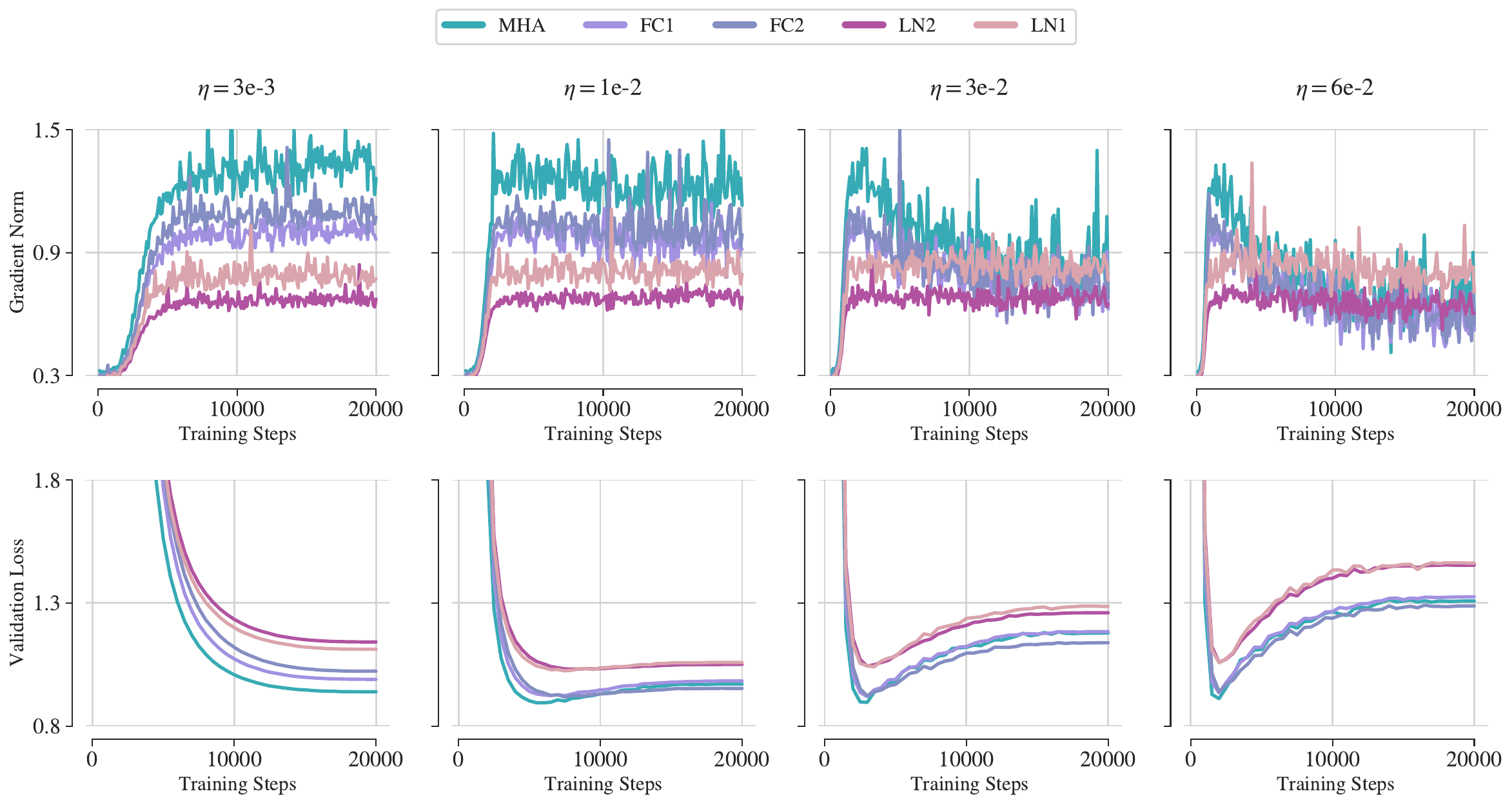}
    \caption{\textbf{Training dynamics on Clipart with seed $0$.} We display the evolution during training of the gradient norms (\textbf{top}) and the validation loss (\textbf{bottom}) of each finetuning configuration of~\cref{tab:model_configurations}, with increasing learning rate $\eta$ from \textbf{left to right}. Components are ordered in terms of decreasing plasticity in the legend. Plastic components have higher gradient norms, which leads to a steeper descent in the validation loss and better downstream performance. The benefits of plasticity are salient across all learning rates. Overall, higher plasticity leads to better optimization and generalization.
    }
    \label{fig:training_evolution_domainnet_clipart_seed_0}
\end{figure}

\begin{figure}[!h]
    \centering
    \includegraphics[width=\linewidth]{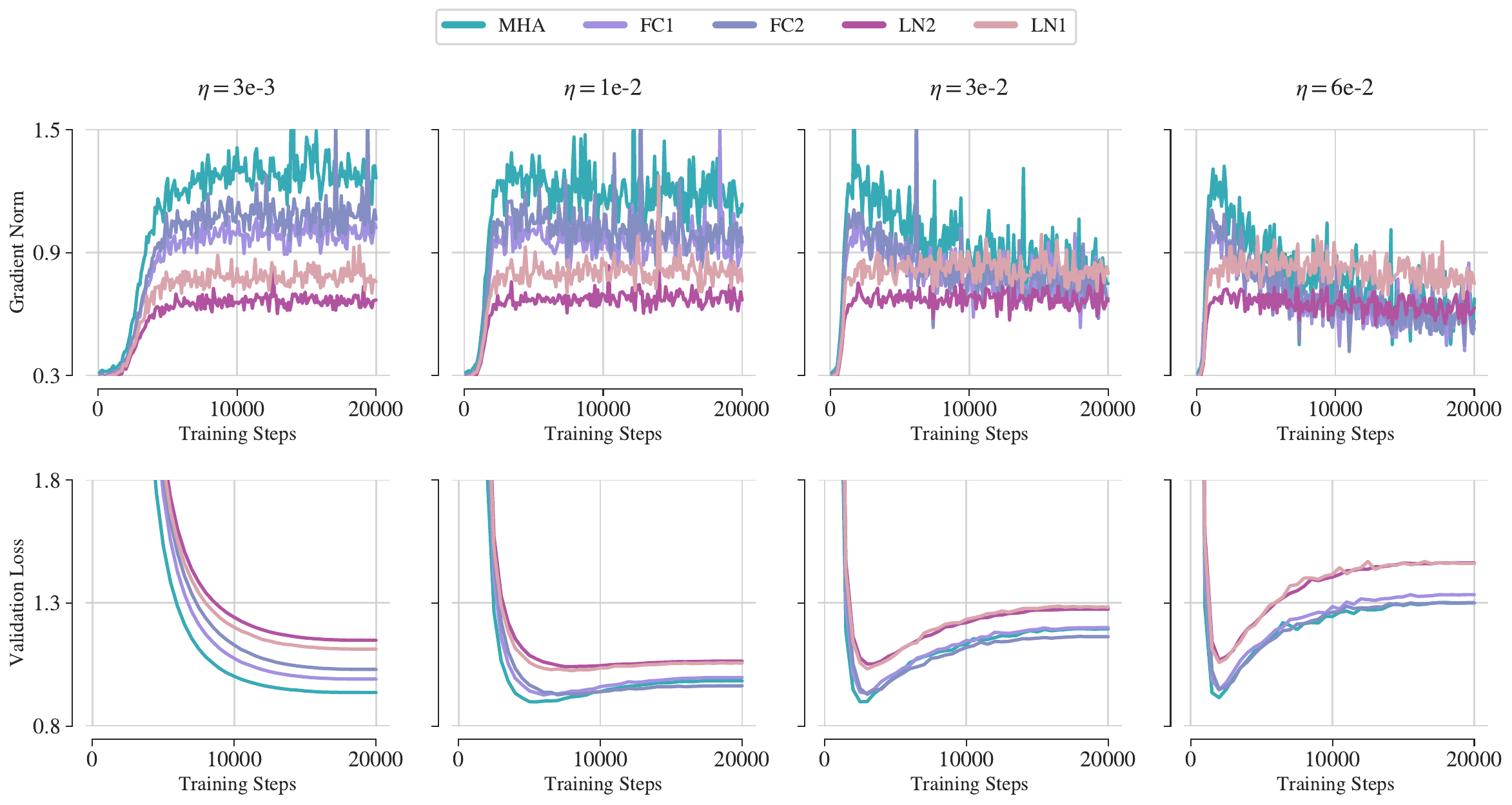}
    \caption{\textbf{Training dynamics on Clipart with seed $42$.} Akin to~\cref{fig:training_evolution_domainnet_clipart_seed_0}, we observe the same consistent pattern of faster and better convergence for components with high plasticity.}
    \label{fig:training_evolution_domainnet_clipart_seed_42}
\end{figure}

\begin{figure}[!h]
    \centering
    \includegraphics[width=\linewidth]{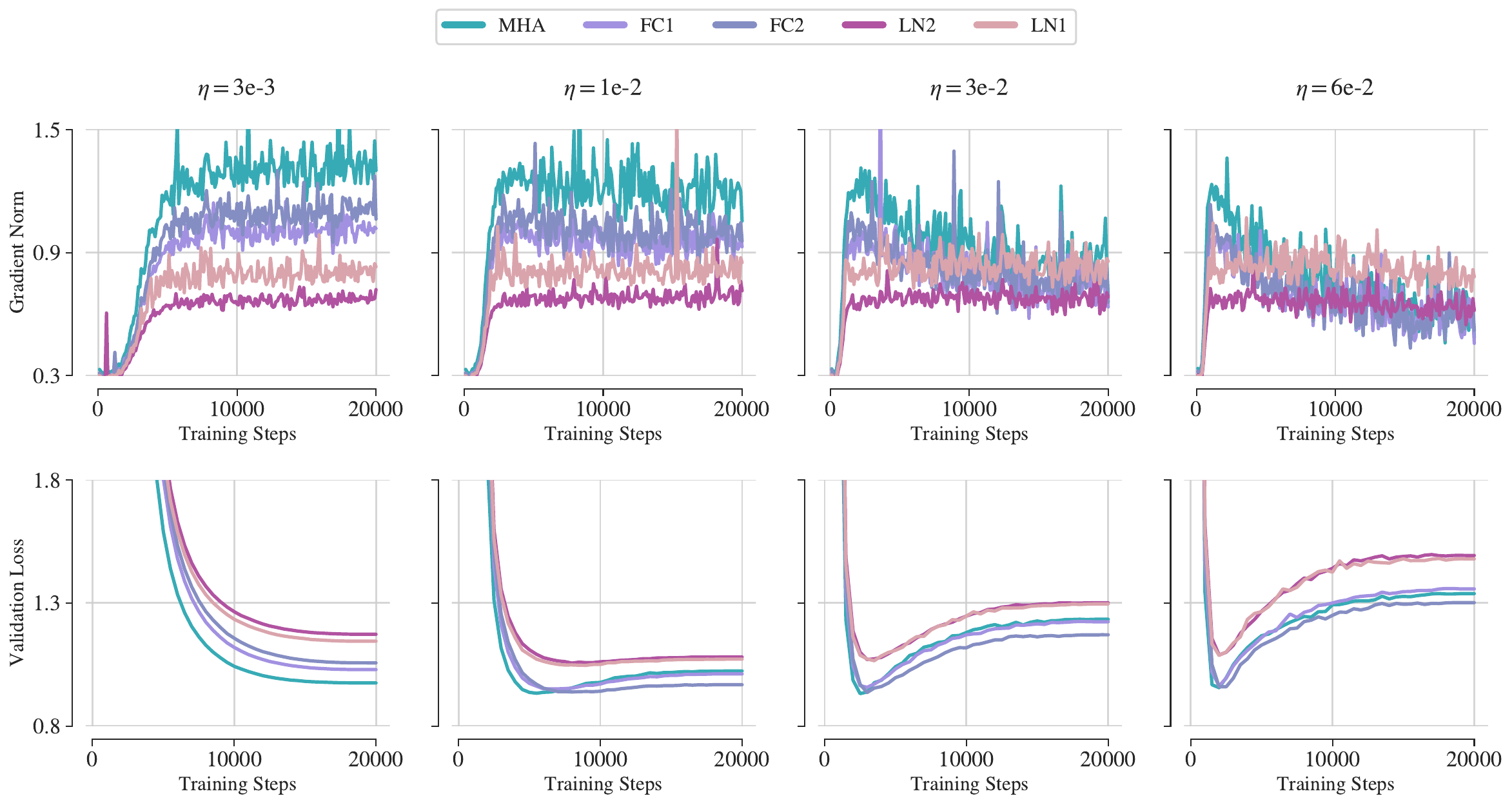}
    \caption{\textbf{Training dynamics on Clipart with seed $3407$.} Akin to~\cref{fig:training_evolution_domainnet_clipart_seed_0}, we observe the same consistent pattern of faster and better convergence for components with high plasticity.}
    \label{fig:training_evolution_domainnet_clipart_seed_3407}
\end{figure}

\begin{figure}[!h]
    \centering
    \includegraphics[width=\linewidth]{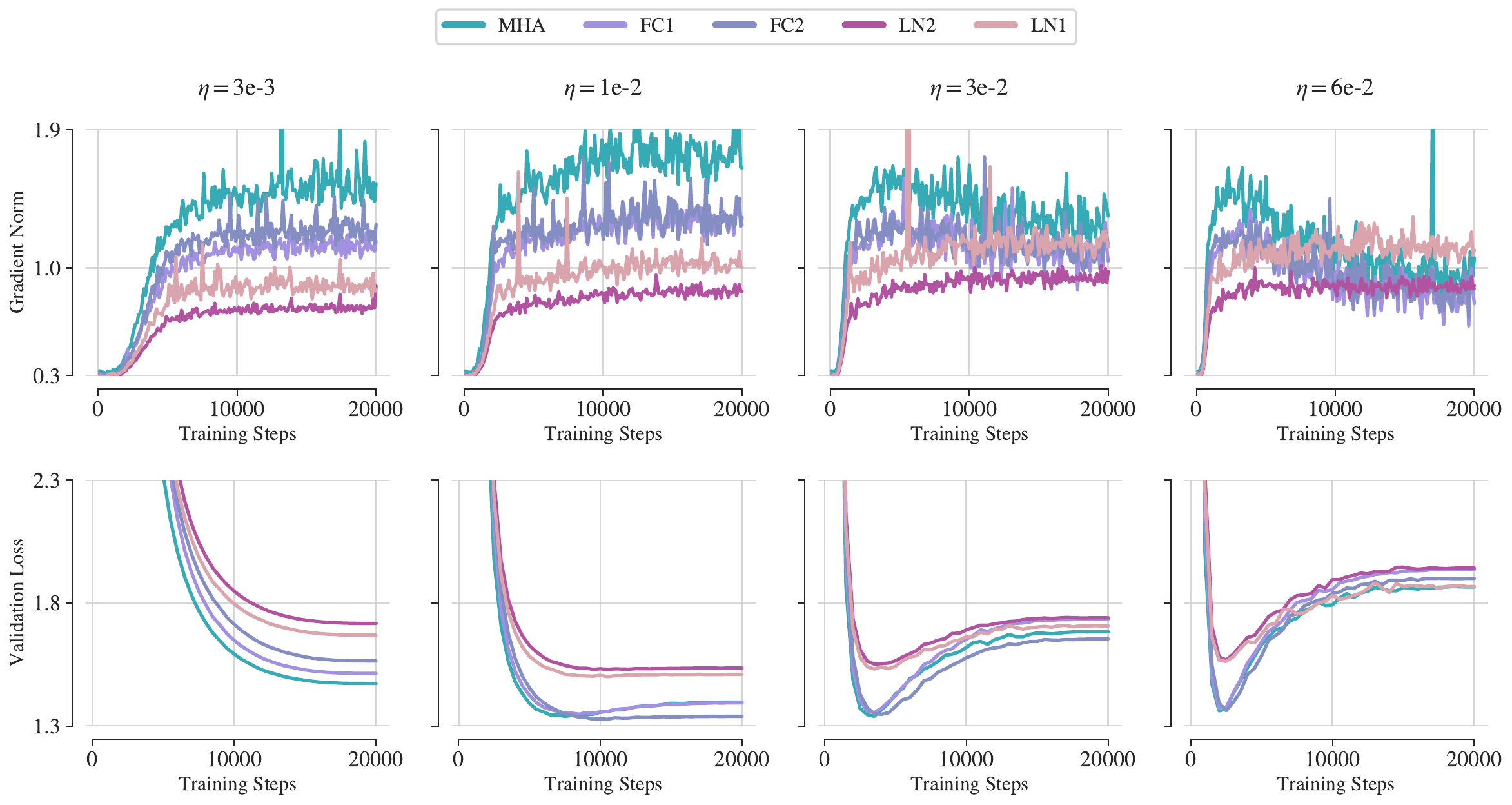}
    \caption{\textbf{Training dynamics on Sketch with seed $0$.} We display the evolution during training of the gradient norms (\textbf{top}) and the validation loss (\textbf{bottom}) of each finetuning configuration of~\cref{tab:model_configurations}, with increasing learning rate $\eta$ from \textbf{left to right}. Components are ordered in terms of decreasing plasticity in the legend. Plastic components have higher gradient norms, which leads to a steeper descent in the validation loss and better downstream performance. The benefits of plasticity are salient across all learning rates. Overall, higher plasticity leads to better optimization and generalization.
    }
    \label{fig:training_evolution_domainnet_sketch_seed_0}
\end{figure}

\begin{figure}[!h]
    \centering
    \includegraphics[width=\linewidth]{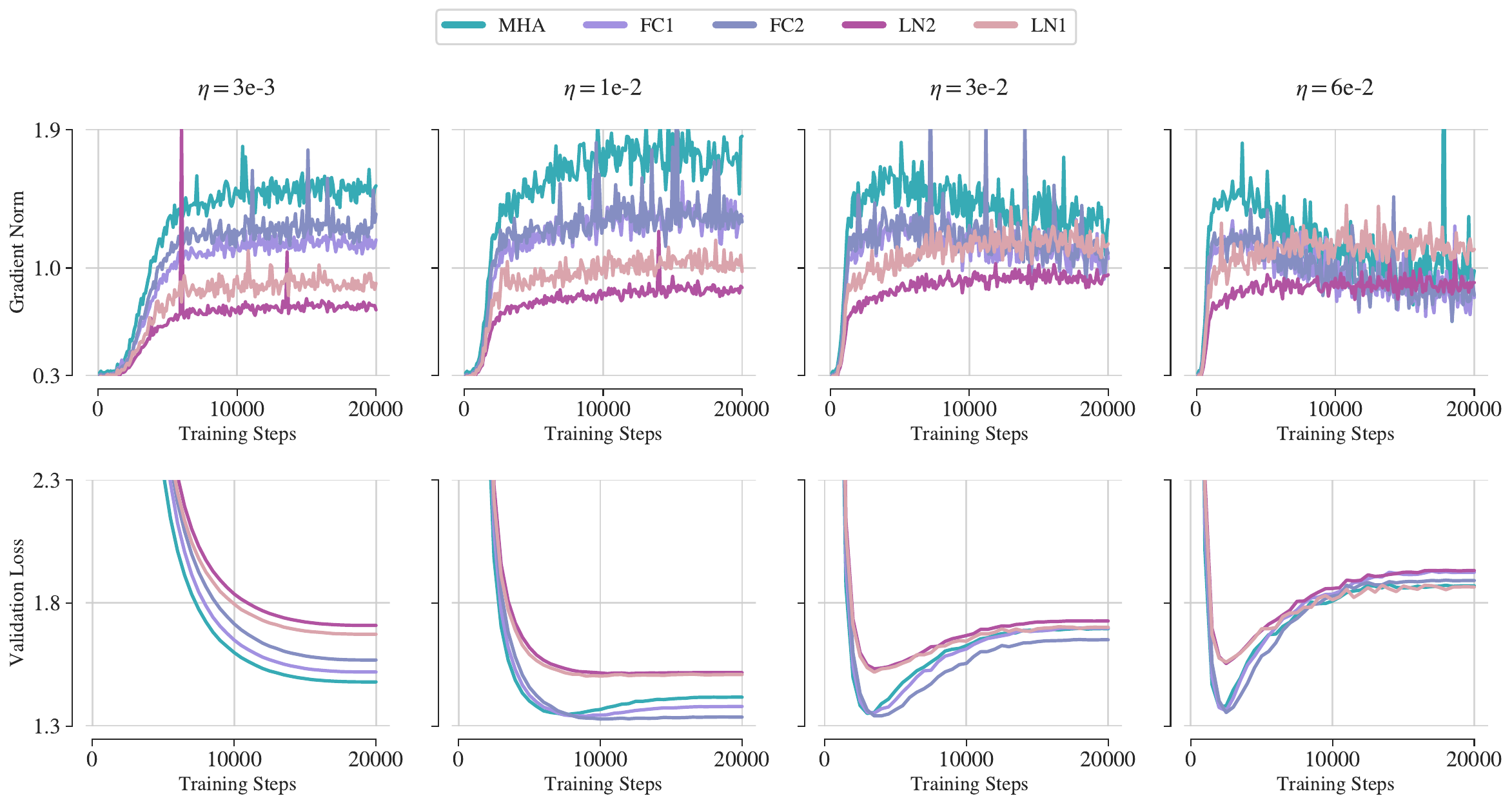}
    \caption{\textbf{Training dynamics on Sketch with seed $42$.} Akin to~\cref{fig:training_evolution_domainnet_sketch_seed_0}, we observe the same consistent pattern of faster and better convergence for components with high plasticity.}
    \label{fig:training_evolution_domainnet_sketch_seed_42}
\end{figure}

\begin{figure}[!h]
    \centering
    \includegraphics[width=\linewidth]{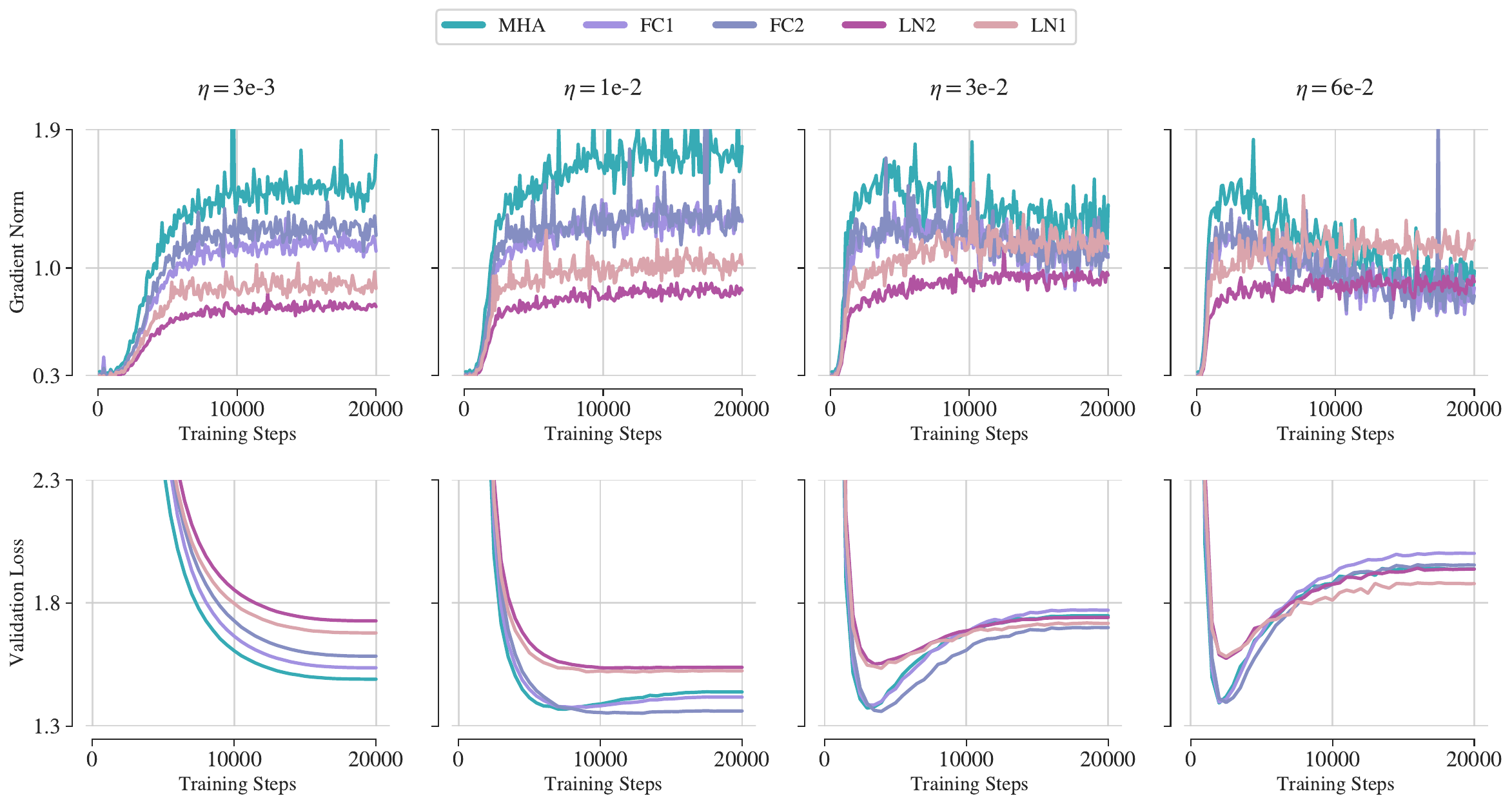}
    \caption{\textbf{Training dynamics on Sketch with seed $3407$.} Akin to~\cref{fig:training_evolution_domainnet_sketch_seed_0}, we observe the same consistent pattern of faster and better convergence for components with high plasticity.}
    \label{fig:training_evolution_domainnet_sketch_seed_3407}
\end{figure}

\begin{figure}[!h]
    \centering
    \includegraphics[width=\linewidth]{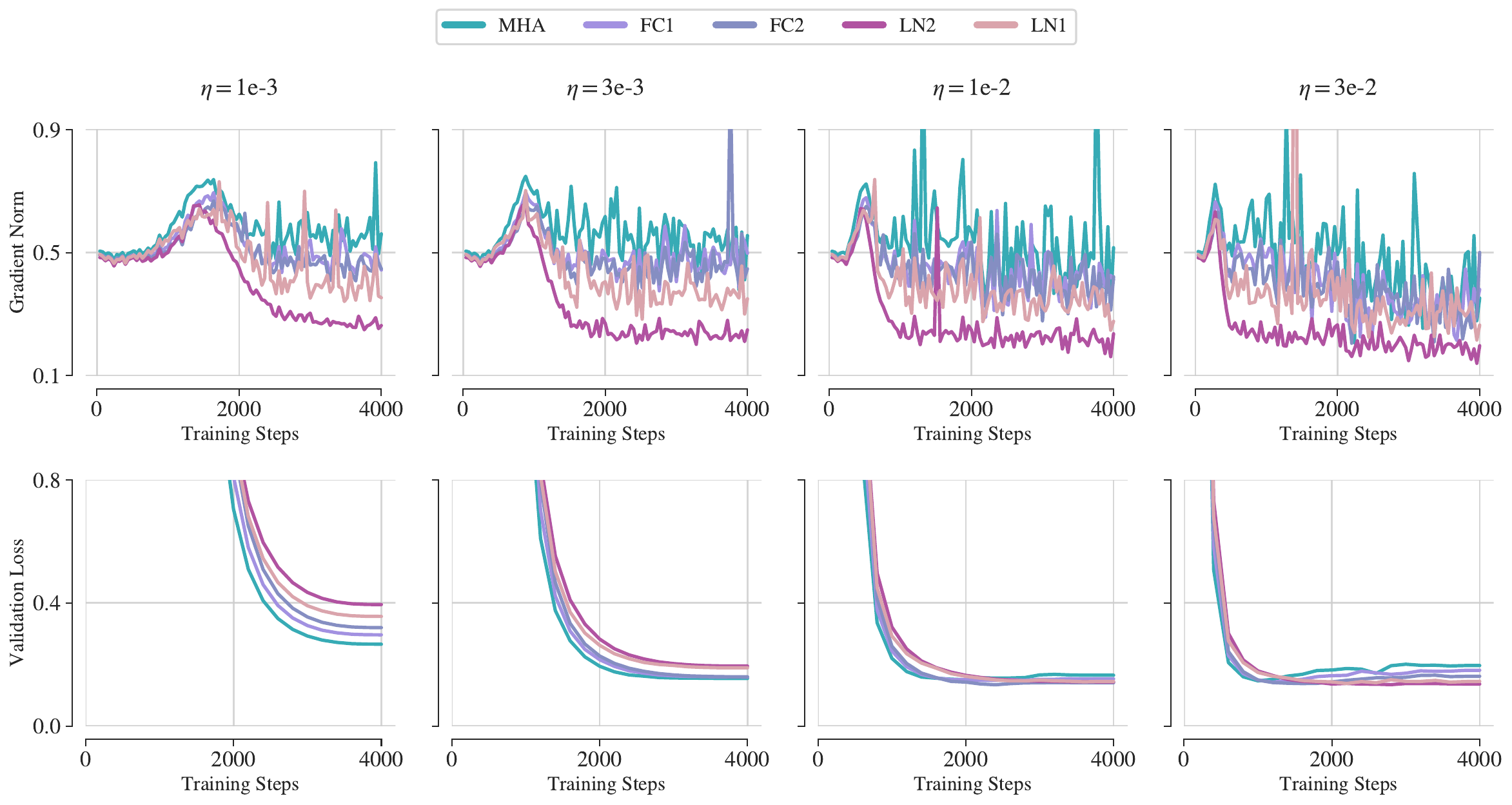}
    \caption{\textbf{Training dynamics on Pet with seed $0$.} We display the evolution during training of the gradient norms (\textbf{top}) and the validation loss (\textbf{bottom}) of each finetuning configuration of~\cref{tab:model_configurations}, with increasing learning rate $\eta$ from \textbf{left to right}. Components are ordered in terms of decreasing plasticity in the legend. On Pet, the pretrained model already achieves good linear probing performance as shown in~\cref{tab:results}. This leads to lower gradient norms for all components compared to more challenging datasets. That being said, we observe for low learning rate that plastic components have higher gradient norms ans a steeper descent in the validation loss.}
    \label{fig:training_evolution_pet_seed_0}
\end{figure}

\begin{figure}[!h]
    \centering
    \includegraphics[width=\linewidth]{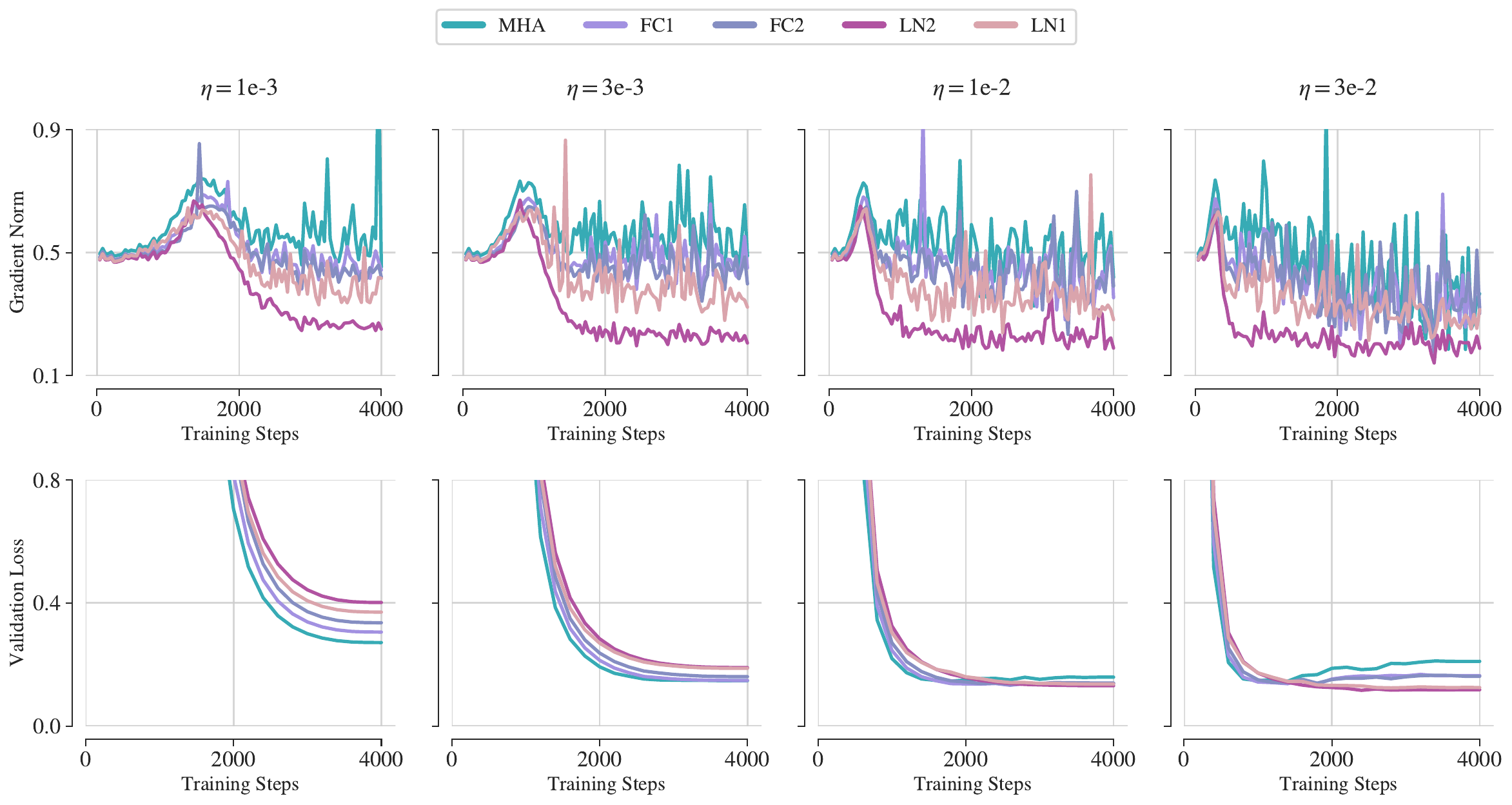}
    \caption{\textbf{Training dynamics on Pet with seed $42$.} Akin to~\cref{fig:training_evolution_pet_seed_0}, we observe the same consistent patterns.}
    \label{fig:training_evolution_pet_seed_42}
\end{figure}

\begin{figure}[!h]
    \centering
    \includegraphics[width=\linewidth]{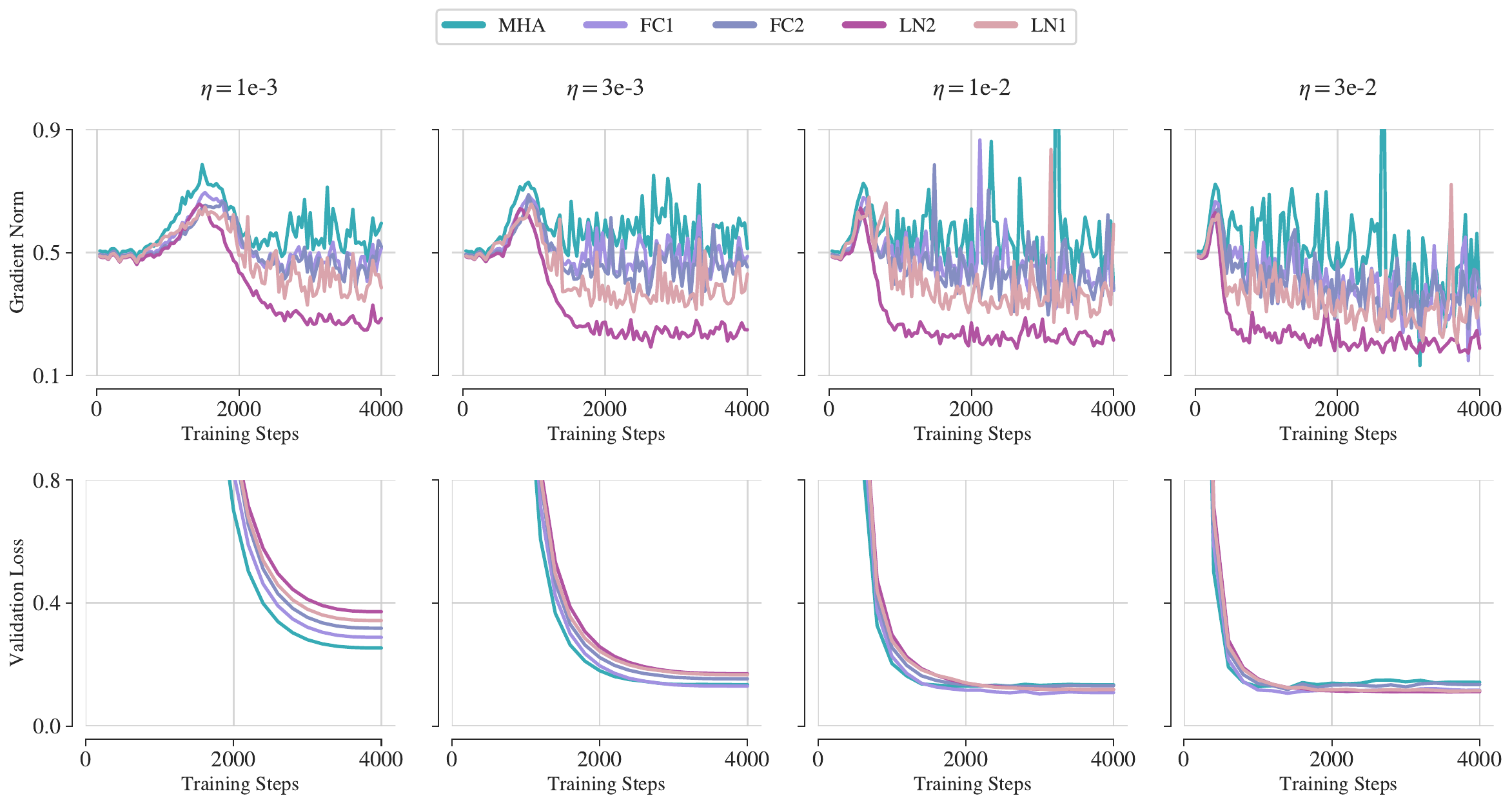}
    \caption{\textbf{Training dynamics on Pet with seed $3407$.} Akin to~\cref{fig:training_evolution_pet_seed_0}, we observe the same consistent patterns.}
    \label{fig:training_evolution_pet_seed_3407}
\end{figure}

\begin{figure}[!h]
    \centering
    \includegraphics[width=\linewidth]{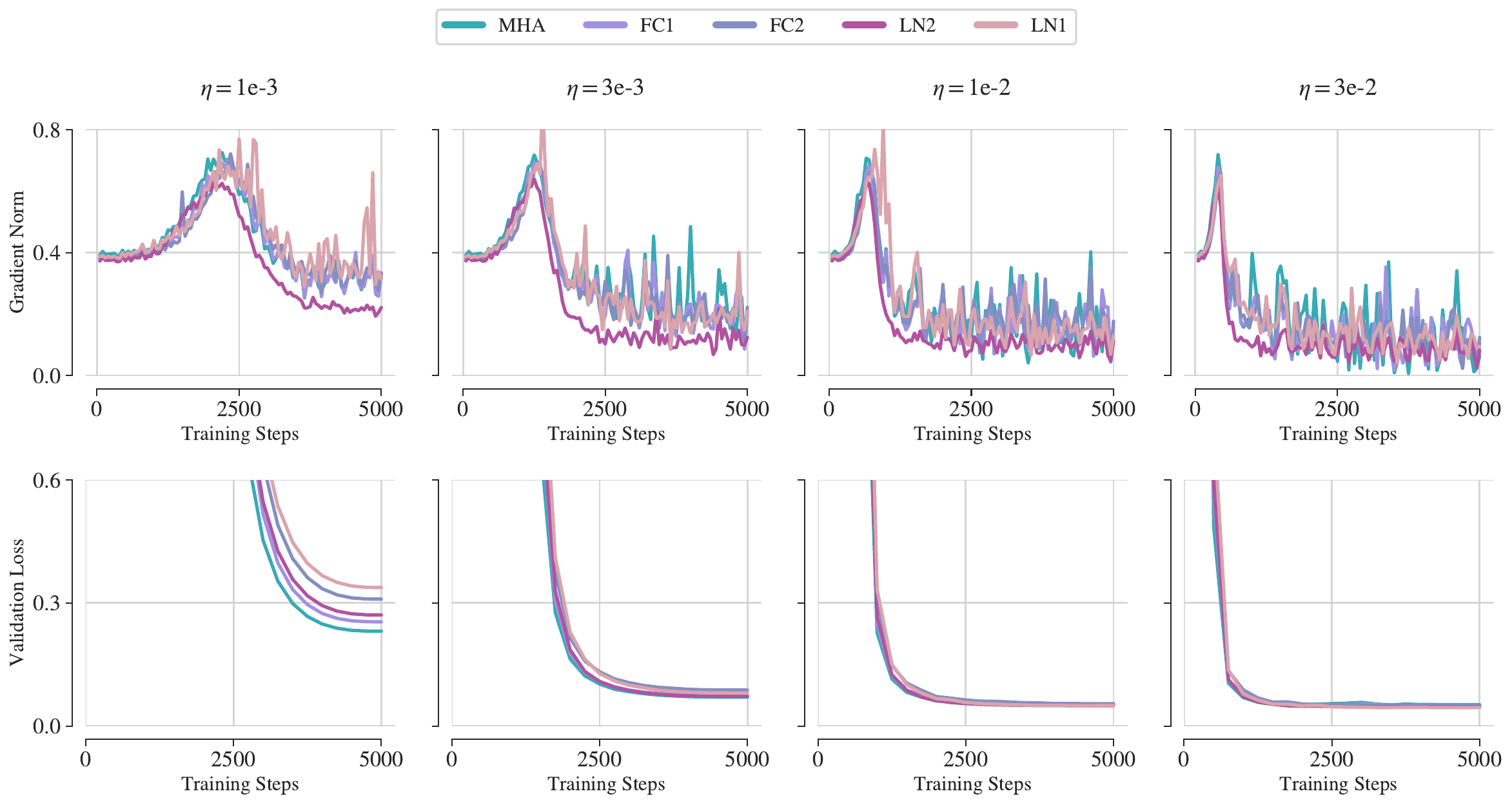}
    \caption{\textbf{Training dynamics on Flowers102 with seed $0$.} We display the evolution during training of the gradient norms (\textbf{top}) and the validation loss (\textbf{bottom}) of each finetuning configuration of~\cref{tab:model_configurations}, with increasing learning rate $\eta$ from \textbf{left to right}. Components are ordered in terms of decreasing plasticity in the legend. On Flowers102, the pretrained model already achieves good linear probing performance as shown in~\cref{tab:results}. This leads to lower gradient norms for all components compared to more challenging datasets. That being said, we observe for low learning rate that plastic components have higher gradient norms ans a steeper descent in the validation loss.}
    \label{fig:training_evolution_flowers102_seed_0}
\end{figure}

\begin{figure}[!h]
    \centering
    \includegraphics[width=\linewidth]{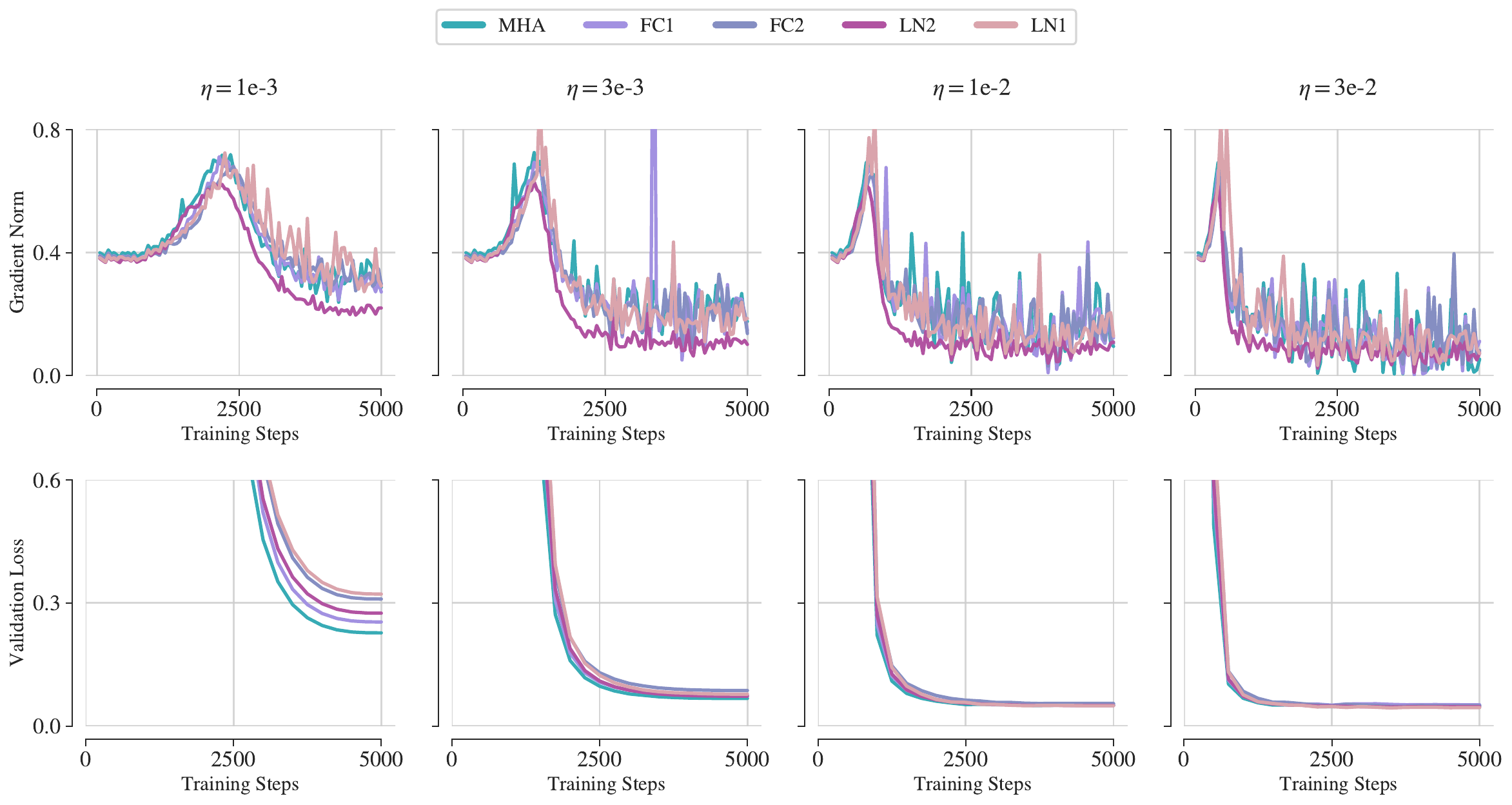}
    \caption{\textbf{Training dynamics on Flowers102 with seed $42$.} Akin to~\cref{fig:training_evolution_flowers102_seed_0}, we observe the same consistent patterns.}
    \label{fig:training_evolution_flowers102_seed_42}
\end{figure}

\begin{figure}[!h]
    \centering
    \includegraphics[width=\linewidth]{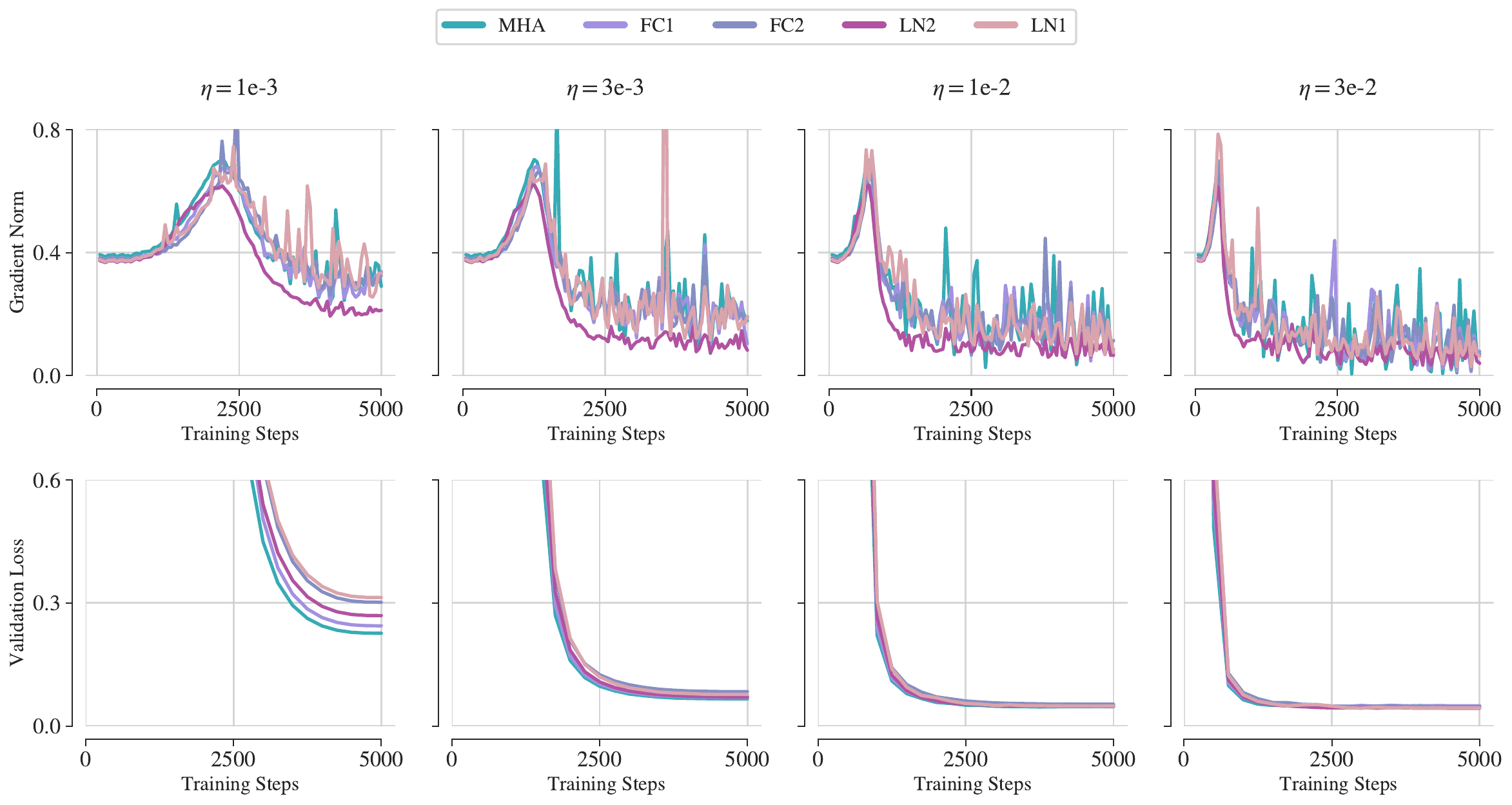}
    \caption{\textbf{Training dynamics on Flowers102 with seed $42$.} Akin to~\cref{fig:training_evolution_flowers102_seed_0}, we observe the same consistent patterns.}
    \label{fig:training_evolution_flowers102_seed_3407}
\end{figure}

\clearpage
\subsubsection{Extension to Adam} 
\label[secinapp]{app:adam_exp}
In our work, we use the SGD optimizer~\citep{bottou2010sgd} following the original ViT paper~\citep{dosovitskiy2021vit}. The Adam optimizer~\citep{kingma2014adam}, more precisely its decoupled weight-decay version from~\citet{loshchilov2018decoupled}, is also a prevalent choice, notably to pretrain large language models~\citep{orvieto2026adam}. As such, it is natural to wonder whether our component-wise analysis holds with Adam. We first note that full-finetuning leads to similar performance for both Adam and SGD~\citep{touvron2021deit}. We keep Adam's default parameters $\beta_1=0.9, \beta_2=0.999$ and, for a fair comparison, we choose smaller learning rates as is standard in the literature on adaptive methods~\citep{kumar2023finetunevisionmodelssgd, dosovitskiy2021vit,touvron2021deit}. More precisely, we follow~\citet{kumar2023finetunevisionmodelssgd} and scale the values with the formula $\mathrm{lr}_\mathrm{adam} = 1\mathrm{e}{-2} \times \mathrm{lr}_\mathrm{sgd}$, where the SGD learning rates of $\mathrm{lr}_\mathrm{sgd}$ are given in~\cref{tab:training_details}. The full results comparison between Adam and SGD is displayed in~\cref{tab:adam_sgd_performance}. We observe that finetuning non-smooth components (in \colorbox{tablegray}{gray shade}) consistently yields higher accuracy and superior stability across learning rates compared to smoother modules. This trend is even more salient with Adam. This showcases that the core thesis that non-smoothness is beneficial to finetuning generalizes across optimizers. Finally, the robust adaptation and interplay between plasticity and optimization are also similar between Adam and SGD as shown in~\cref{fig:adamw_sgd_robustness_domainnet_sketch}. The evolution of the gradient norms (middle) and the validation loss (right) is given for the finetuning run on Sketch that achieves the highest accuracy (this corresponds to learning rate $\eta=1\mathrm{e-}2$). This confirms our intuition: the ordering of gradient norms is aligned with the \emph{plasticity ranking} established in~\cref{sec:analysis} and the loss descent is steeper for components with high plasticity, such as the attention modules and the feedforward layers. These patterns are consistent with those of SGD displayed in~\cref{fig:robustness_training} and hold across learning rates as shown in~\cref{fig:adam_training_evolution_cifar100_seed_0,fig:adam_training_evolution_motion_blur_seed_0,fig:adam_training_evolution_clipart_seed_0,fig:adam_training_evolution_sketch_seed_0}.
\begin{table*}[!h]
\centering
    \caption{\textbf{Adam vs SGD}. Average top-$1$ finetuning accuracy on the test set with standard deviation across learning rates. Transformer components are ordered in terms of decreasing plasticity. The best performance among components for each optimizer is in \textbf{bold} and non-smooth components are highlighted in \colorbox{tablegray}{gray}. The takeaways are similar for SGD and Adam: non-smooth components lead to better and more stable finetuning performance. We note that, although the accuracy is slightly lower with Adam, the stability of non-smooth components is more salient.}
    \label{tab:adam_sgd_performance}
    \scalebox{.9}{
    \begin{NiceTabular}{lcccccccccc}
    \CodeBefore
    \columncolor{tablegray}{2-7} 
    \rowcolor{white}{1,2}
    \Body
    \multirow{2}{*}{configuration} 
    & \multicolumn{2}{c}{MHA} & \multicolumn{2}{c}{FC1} & \multicolumn{2}{c}{FC2} & \multicolumn{2}{c}{LN2} & \multicolumn{2}{c}{LN1} \\
    \cmidrule(l{1em}r{1em}){2-3} \cmidrule(l{1em}r{1em}){4-5} \cmidrule(l{1em}r{1em}){6-7} \cmidrule(l{1em}r{1em}){8-9} \cmidrule(l{1em}r{1em}){10-11} 
     & \small Adam & \small SGD & \small Adam & \small SGD & \small Adam & \small SGD & \small Adam & \small SGD & \small Adam & \small SGD \\
    \toprule[\thick pt]
    Cifar100 & 91.0 $_{\tiny \pm \text{0.2}}$ & \textbf{91.7} $_{\tiny \pm \text{1.1}}$ & \textbf{91.3} $ _{\tiny \pm \text{0.6}}$ & 91.6 $ _{\tiny \pm \text{1.6}}$ & 90.6 $ _{\tiny \pm \text{1.4}}$ & 91.2 $ _{\tiny \pm \text{0.7}}$ & 89.4 $ _{\tiny \pm \text{2.7}}$ & 90.6 $ _{\tiny \pm \text{1.4}}$ & 88.4 $ _{\tiny \pm \text{3.2}}$ & 89.9 $ _{\tiny \pm \text{1.7}}$ \\
    Motion Blur & \textbf{92.4} $_{\tiny \pm \text{0.2}}$ & 94.1 $_{\tiny \pm \text{0.2}}$ & 91.8 $ _{\tiny \pm \text{0.4}}$ & \textbf{94.4} $ _{\tiny \pm \text{0.2}}$ & 90.7 $ _{\tiny \pm \text{0.9}}$ & 93.7 $ _{\tiny \pm \text{0.3}}$ & 91.7 $ _{\tiny \pm \text{2.4}}$ & 93.7 $ _{\tiny \pm \text{0.6}}$ & 90.7 $ _{\tiny \pm \text{2.4}}$ & 92.6 $ _{\tiny \pm \text{0.4}}$ \\
    Clipart & \textbf{76.4} $_{\tiny \pm \text{0.5}}$ & \textbf{76.9} $_{\tiny \pm \text{0.6}}$ & 74.8 $ _{\tiny \pm \text{0.8}}$ & 75.7 $ _{\tiny \pm \text{0.6}}$ & 75.8 $ _{\tiny \pm \text{0.3}}$ & 75.9 $ _{\tiny \pm \text{0.8}}$ & 73.3 $ _{\tiny \pm \text{1.9}}$ & 73.6 $ _{\tiny \pm \text{0.8}}$ & 73.5 $ _{\tiny \pm \text{1.5}}$ & 74.0 $ _{\tiny \pm \text{0.5}}$ \\
    Sketch & 69.0 $_{\tiny \pm \text{0.5}}$ & \textbf{68.5} $_{\tiny \pm \text{1.1}}$ & 67.9 $ _{\tiny \pm \text{0.6}}$ & 68.1 $ _{\tiny \pm \text{1.6}}$ & \textbf{69.2} $ _{\tiny \pm \text{0.4}}$ & 68.1 $ _{\tiny \pm \text{2.0}}$ & 63.1 $ _{\tiny \pm \text{2.9}}$ & 63.9 $ _{\tiny \pm \text{1.6}}$ & 64.0 $ _{\tiny \pm \text{2.6}}$ & 64.5 $ _{\tiny \pm \text{1.0}}$ \\
    \midrule[\midthick pt]
    Average & \textbf{82.2} & \textbf{82.8} & 81.4 & 82.5 & 81.7 & 82.1  & 79.4  & 80.5  & 79.2 & 80.2 \\
    \end{NiceTabular}
    }
\end{table*}
\begin{figure}[!h]
    \centering
    \includegraphics[width=\linewidth]{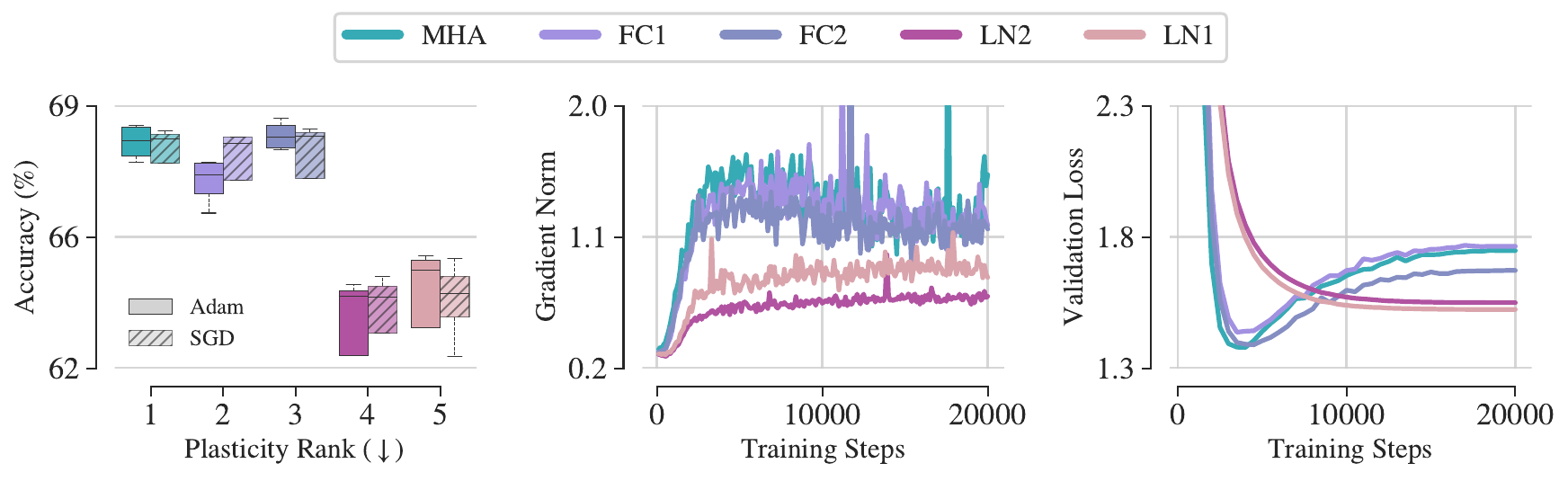}
    \caption{\textbf{Benefits of plasticity with Adam on Sketch.} Transformer components are ordered in terms of decreasing plasticity. We can see that the performance across learning rates (\textbf{left}) is better and more stable for plastic components, with similar trends for both Adam and SGD. The evolution of Adam's gradient norms (\textbf{middle}) and validation loss (\textbf{right}) throughout training is consistent with those of SGD shown in~\cref{fig:robustness_training}: the higher plasticity, the larger gradient norms, and the better the generalization.}
    \label{fig:robustness_training_adam}
\end{figure}
\begin{figure}[!h]
    \centering
    \includegraphics[width=\linewidth]{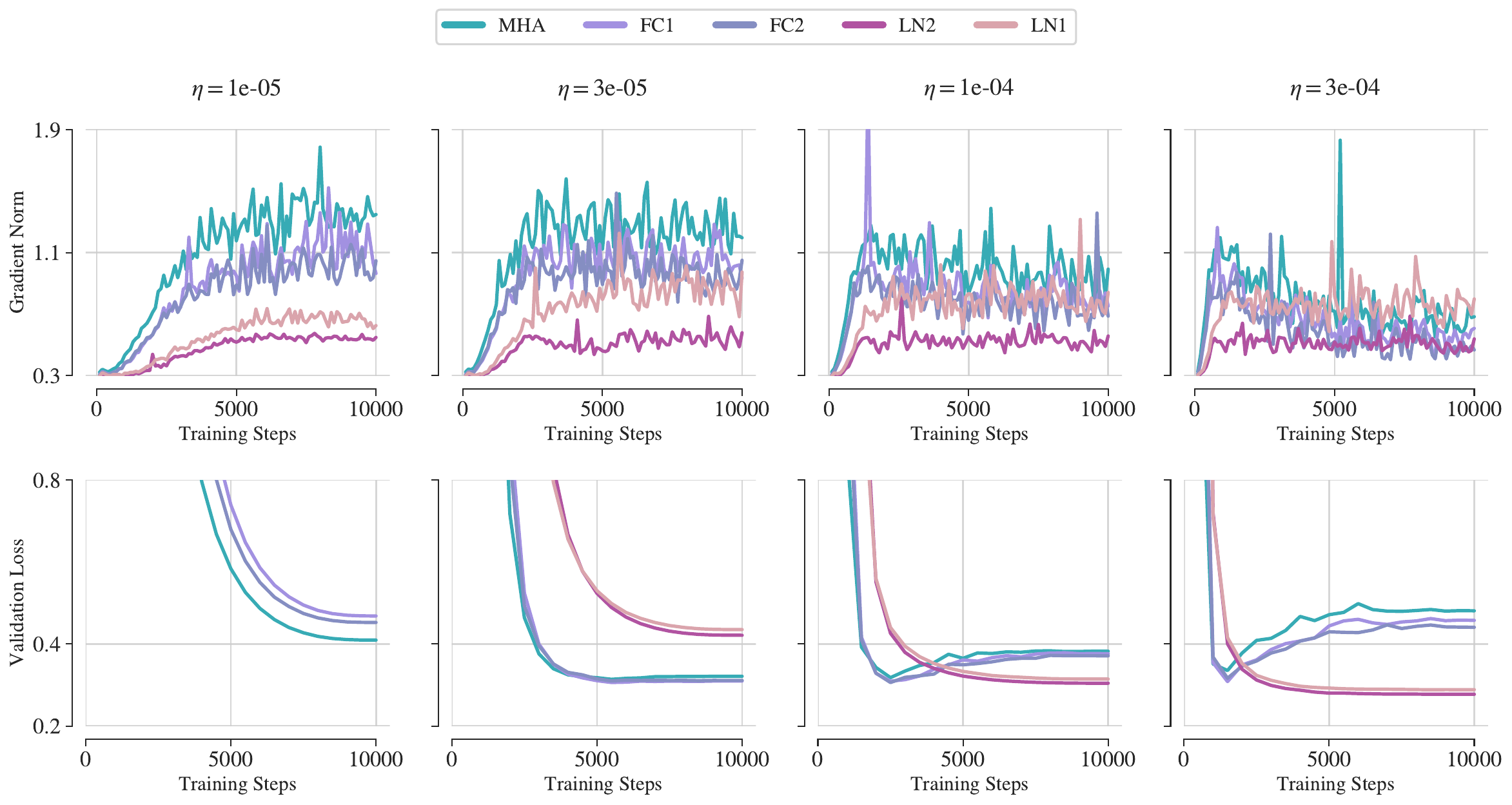}
    \caption{\textbf{Training dynamics on Cifar100 with Adam and seed $0$.} We display the evolution during training of the gradient norms (\textbf{top}) and the validation loss (\textbf{bottom}) of each finetuning configuration of~\cref{tab:model_configurations}, with increasing learning rate $\eta$ from \textbf{left to right}. Components are ordered in terms of decreasing plasticity in the legend. Plastic components have higher gradient norms, which leads to a steeper descent in the validation loss and better downstream performance. The benefits of plasticity are even more salient with low learning rates. Overall, higher plasticity leads to better optimization and generalization.
    }
    \label{fig:adam_training_evolution_cifar100_seed_0}
\end{figure}

\begin{figure}[!h]
    \centering
    \includegraphics[width=\linewidth]{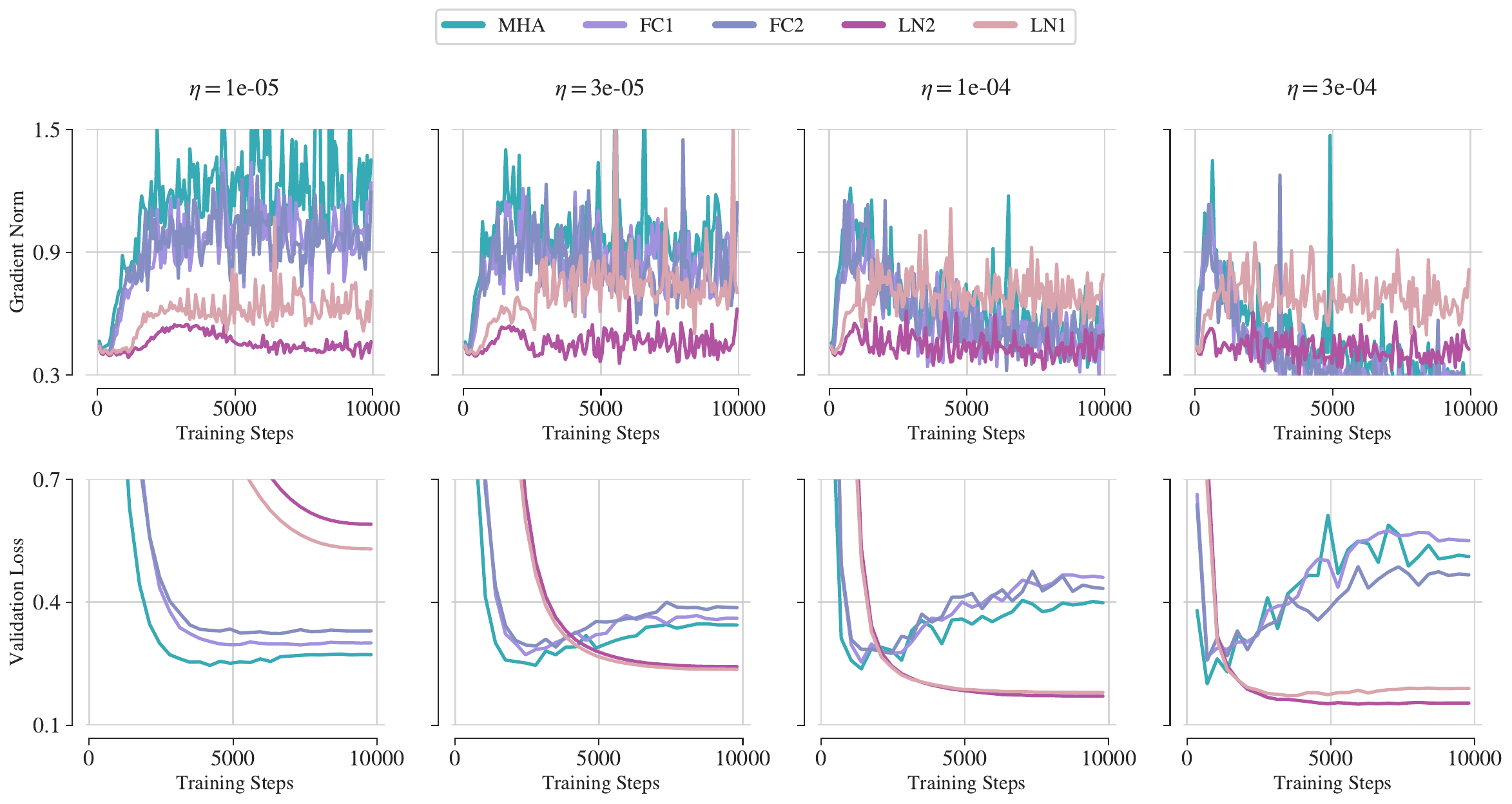}
    \caption{\textbf{Training dynamics on Motion Blur with Adam and seed $0$.} We display the evolution during training of the gradient norms (\textbf{top}) and the validation loss (\textbf{bottom}) of each finetuning configuration of~\cref{tab:model_configurations}, with increasing learning rate $\eta$ from \textbf{left to right}. Components are ordered in terms of decreasing plasticity in the legend. Plastic components have higher gradient norms, which leads to a steeper descent in the validation loss and better downstream performance. The benefits of plasticity are even more salient with low learning rates. Overall, higher plasticity leads to better optimization and generalization.
    }
    \label{fig:adam_training_evolution_motion_blur_seed_0}
\end{figure}

\begin{figure}[!h]
    \centering
    \includegraphics[width=\linewidth]{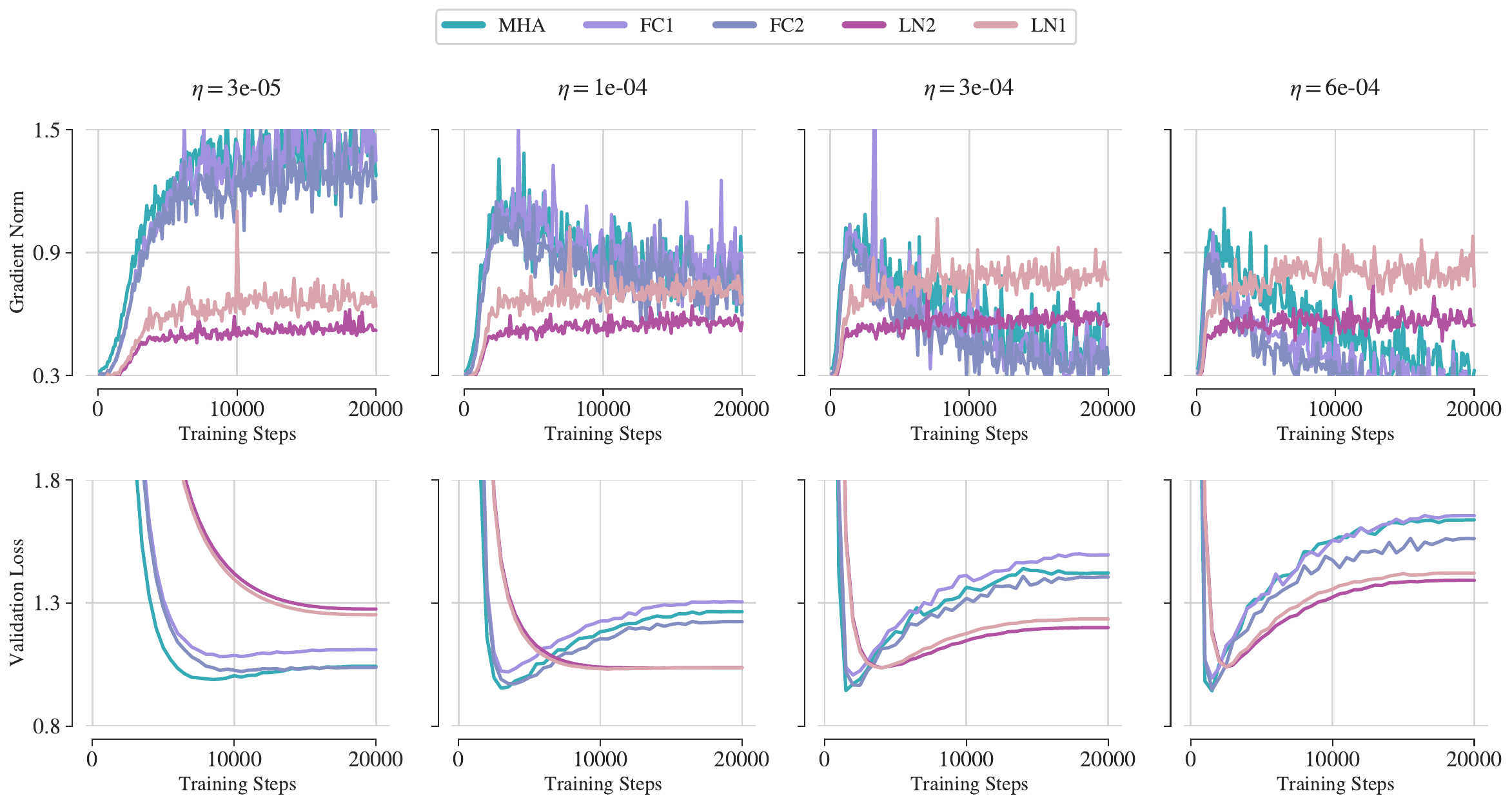}
    \caption{\textbf{Training dynamics on Clipart with Adam and seed $0$.} We display the evolution during training of the gradient norms (\textbf{top}) and the validation loss (\textbf{bottom}) of each finetuning configuration of~\cref{tab:model_configurations}, with increasing learning rate $\eta$ from \textbf{left to right}. Components are ordered in terms of decreasing plasticity in the legend. Plastic components have higher gradient norms, which leads to a steeper descent in the validation loss and better downstream performance. The benefits of plasticity are even more salient with low learning rates. Overall, higher plasticity leads to better optimization and generalization.
    }
    \label{fig:adam_training_evolution_clipart_seed_0}
\end{figure}

\begin{figure}[!h]
    \centering
    \includegraphics[width=\linewidth]{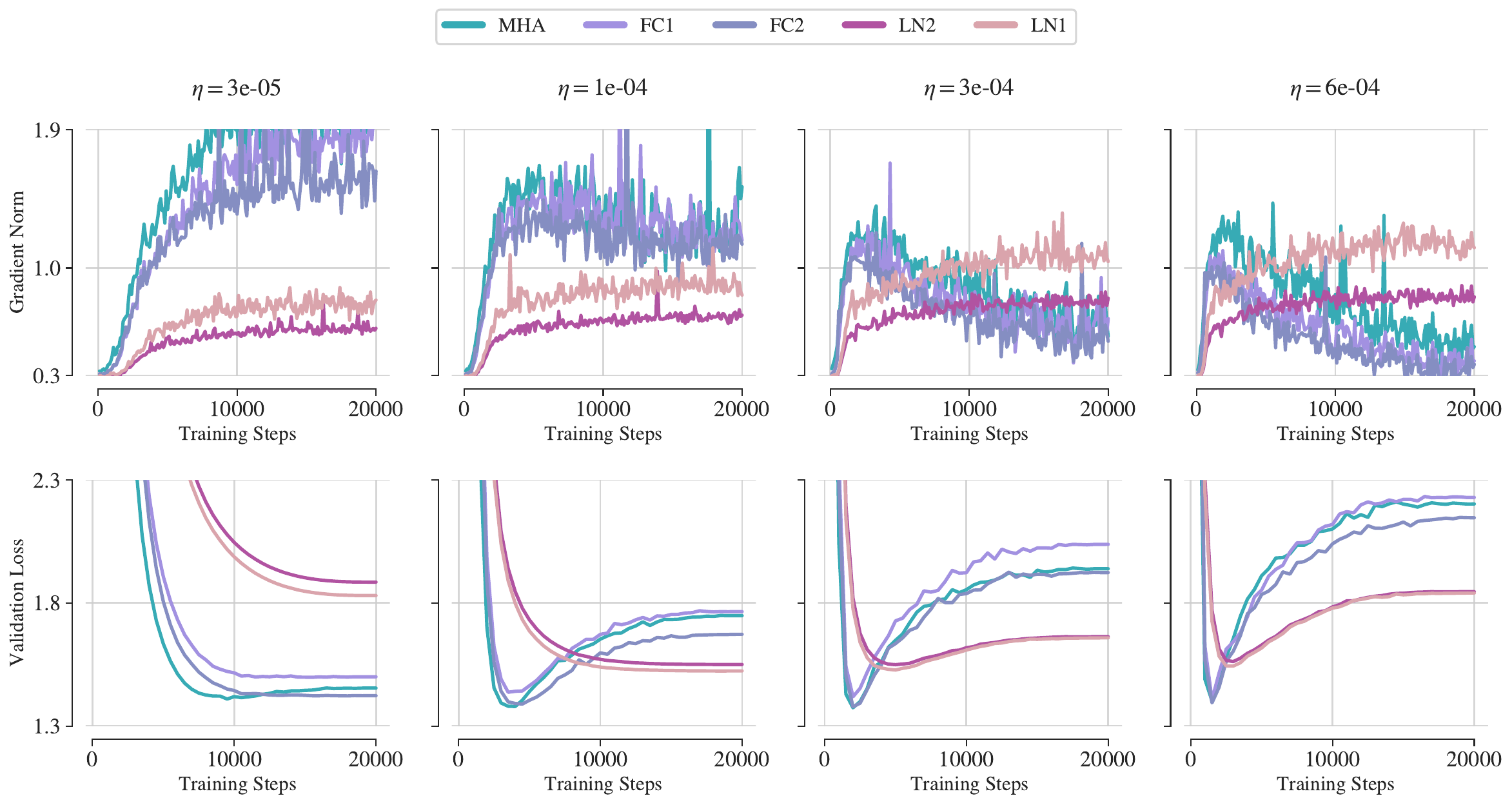}
    \caption{\textbf{Training dynamics on Sketch with Adam and seed $0$.} We display the evolution during training of the gradient norms (\textbf{top}) and the validation loss (\textbf{bottom}) of each finetuning configuration of~\cref{tab:model_configurations}, with increasing learning rate $\eta$ from \textbf{left to right}. Components are ordered in terms of decreasing plasticity in the legend. Plastic components have higher gradient norms, which leads to a steeper descent in the validation loss and better downstream performance. The benefits of plasticity are even more salient with low learning rates. Overall, higher plasticity leads to better optimization and generalization.
    }
    \label{fig:adam_training_evolution_sketch_seed_0}
\end{figure}
\end{document}